%% file: main_arxiv.tex
\documentclass{article}

\usepackage[preprint]{corl_2022} 

\usepackage{bm}
\usepackage{enumitem}
\usepackage{multirow}
\usepackage{graphicx}
\usepackage{caption}
\usepackage{subcaption}
\usepackage[ruled,linesnumbered]{algorithm2e}
\usepackage{rotating}
\usepackage{optidef}
\usepackage[table]{xcolor}
\usepackage{color,soul}
\usepackage{calc}
\usepackage{tabularx}
\usepackage{booktabs}
\usepackage{ragged2e}
\usepackage{xcolor}
\usepackage{multirow}
\usepackage{wrapfig}
\usepackage{amssymb}
\renewcommand{\paragraph}[1]{\vspace{.1em}\noindent\textbf{#1}.}
\setlist[itemize]{leftmargin=6mm}
\usepackage[export]{adjustbox}

\title{ROAD: Learning an Implicit Recursive Octree Auto-Decoder to Efficiently Encode 3D Shapes}

\author{
Sergey Zakharov, 
Rareş Ambruş,
Katherine Liu, 
Adrien Gaidon
\And
\textnormal{Toyota Research Institute}
}

\begin{document}
\maketitle

\newcommand{\method}{ROAD}
\newcommand{\methodspace}{ROAD }
\newcommand{\latentSet}{\boldsymbol{\mathcal{Z}}}
\newcommand{\latent}{\boldsymbol{z}}
\newcommand{\latentDim}{D}
\newcommand{\normal}{\boldsymbol{n}}
\newcommand{\LoD}{m}
\newcommand{\maxLoD}{M}
\newcommand{\normalPrediction}{\boldsymbol{n}}
\newcommand{\normalGT}{\normalPrediction^{GT}}
\newcommand{\surfacePointGT}{\surfacePoint^{GT}}
\newcommand{\occupancyPrediction}{o}
\newcommand{\sdfPrediction}{s}
\newcommand{\surfacePoint}{\boldsymbol{p}}
\newcommand{\estimatedSurfacePoint}{\surfacePoint}
\newcommand{\voxelCoordinate}{\boldsymbol{x}}
\newcommand{\objectIndex}{k}
\newcommand{\numObjects}{K}
\newcommand{\sdfScaling}{\alpha}
\newcommand{\probabilityThreshold}{\theta}


\input{0-abstract}
\input{1-intro}
\input{2-related}
\input{3-method}
\input{4-experiments}
\input{5-conclusion}
\newpage
\bibliography{egbib}
\appendix
\newpage

\begin{center}
\textbf{\Large Supplementary Material}
\end{center}
\captionsetup{belowskip=0pt}
\input{6-supplementary}

\end{document}

%% file: 0-abstract.tex

\begin{abstract}
 
Compact and accurate representations of 3D shapes are central to many perception and robotics tasks. State-of-the-art learning-based methods can reconstruct single objects but scale poorly to large datasets. We present a novel recursive implicit representation to efficiently and accurately encode large datasets of complex 3D shapes by recursively traversing an implicit octree in latent space.
Our implicit Recursive Octree Auto-Decoder (\method) learns a hierarchically structured latent space enabling state-of-the-art reconstruction results at a compression ratio above 99\%. We also propose an efficient curriculum learning scheme that naturally exploits the coarse-to-fine properties of the underlying octree spatial representation. We explore the scaling law relating latent space dimension, dataset size, and reconstruction accuracy, showing that increasing the latent space dimension is enough to scale to large shape datasets. Finally, we show that our learned latent space encodes a coarse-to-fine hierarchical structure yielding reusable latents across different levels of details, and we provide qualitative evidence of generalization to novel shapes outside the training set. Project website: \url{https://zakharos.github.io/projects/road/}
\end{abstract}

\keywords{Implicit shape representations, Reconstruction, Data compression} 

%% file: 1-intro.tex
\section{Introduction}

Accurately and efficiently representing 3D geometry is a cornerstone capability in computer vision and computer graphics, with many practical applications in robotics and artificial intelligence. Decades of research in this area have produced a myriad of approaches, from traditional explicit methods~\cite{rusu20113d,qi2017pointnet,curless1996volumetric,hornung2013octomap,peng2020convolutional,whelan2015elasticfusion,hanocka2019meshcnn} to learning-based implicit representations that encode shapes in the weights of neural networks and use various learning cues such as Signed Distance Fields~\cite{park2019deepsdf,sitzmann2020metasdf,zakharov2020autolabeling}, occupancy~\cite{mescheder2019occupancy,peng2020convolutional}, or radiance~\cite{mildenhall2020nerf,yu2020pixelnerf,sitzmann2021light,barron2021mip}. Choosing one representation over another typically involves various tradeoffs between accuracy, scalability and generalization~\cite{tangelder2008survey,ahmed2018survey,xie2021neural}. Related methods have either focused on modeling single shapes with increasing levels of accuracy at higher costs in terms of memory or time~\cite{mildenhall2020nerf,yifan2021geometry,takikawa2021neural,muller2022instant} or on modeling classes of shapes, typically with single MLPs, which results in the ability to generalize to novel shapes as well as adapt to test-time data via differentiability but at the expense of high-frequency details~\cite{park2019deepsdf,tang2021octfield,williams2021neural,zakharov2021single}. 

In this paper we address these key challenges through a novel neural network capable of simultaneously encoding a large number of shapes to a higher level of accuracy than previously possible. We build on recent advances which use neural fields (i.e. neural network parameterizations of continuous functions defined on Euclidean space) to represent any topology with arbitrary precision~\cite{xie2021neural} combined with continuous, generative latent spaces for 3D shape generation~\cite{park2019deepsdf,chen2019learning}. We aim to recover a compact function which learns to map complex topologies to an implicit space of shapes, without sacrificing reconstruction accuracy or differentiability. Our key insight is based on the fact that the neural field paradigm maps spatial coordinates to an encoding of the surface and thus requires thousands of evaluations to extract the underlying surface; moreover, the underlying implicit function needs to accurately model space outside the target geometry~\cite{lombardi2019neural,liu2020dist,niemeyer2020differentiable}. Alternatively, some methods explicitly define the relationship between an underlying explicit data structure and the latent space~\cite{tang2021octfield,takikawa2021neural}, thus partitioning the implicit function.

To achieve high compression while still being able to reconstruct high-frequency details, we propose an implicit Recursive Octree Auto-Decoder (\method) formulation that operates entirely in the latent space and is guided by an octree partitioning of the space. The octree data structure provides a simple yet elegant solution for increasing surface detail while traversing down the tree~\cite{yu2021plenoctrees,takikawa2021neural,tang2021octfield}, as well as an intuitive setup for a curriculum learning schedule where learning progresses according to a coarse-to-fine approach. Our method uses a single neural network to map a latent vector to eight other latent vectors corresponding to its eight children in 3-dimensional space; in turn, each of the predicted latent vectors can be fed back to the network for further subdivision. To extract a surface, we simply traverse down the tree starting from a single root latent vector, expanding nodes as needed based on predicted occupancy, until the desired level of resolution is reached. The output of a forward pass of our network is the actual surface, and it can be obtained in milliseconds, unlike seconds or minutes for related methods requiring complex operations. 

Our formulation leads to a latent space that captures shape similarity and hierarchy in a manner that is conducive to high compression as well generalization to shapes outside of the training set. Our efficient implementation leads to a reduction of up to 99\% in terms of storage space compared to the original mesh models; this includes network weights as well as any other learned features needed for reconstructing the meshes. Finally, we explore the relationship between 3D modeling power (as measured by the Chamfer distance), latent space dimension and dataset size, and show how to tune our model's capacity with a single hyperparameter --- the dimension of the latent space. 


\begin{figure*}[t]
    \centering
    \includegraphics[width=1\linewidth]{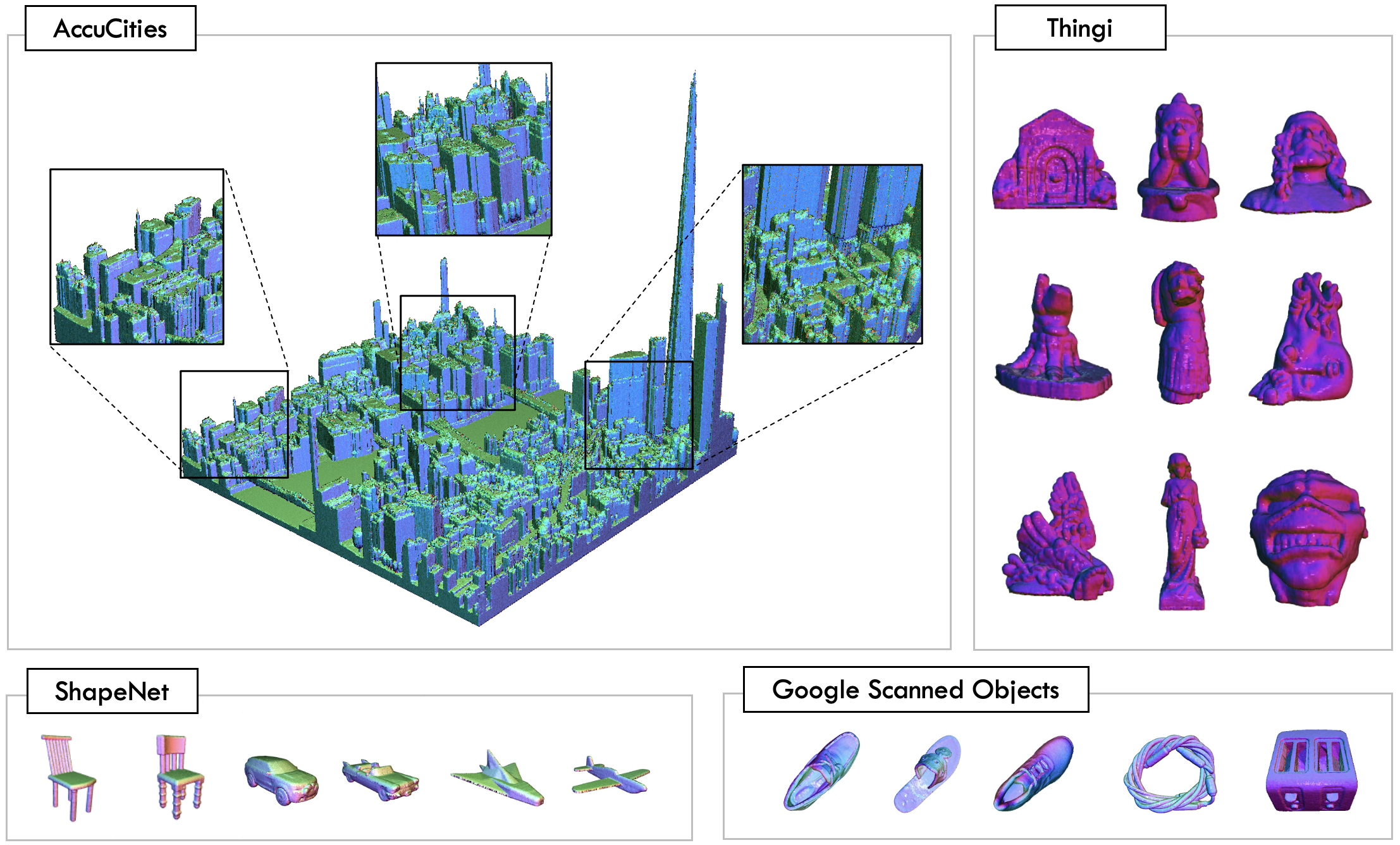}
    \caption{Our implicit Recursive Octree Auto-Decoder, \textbf{\method}, can represent many diverse shapes accurately and precisely with a small memory footprint thanks to recursive decoding of a hierarchically structured learned implicit shape representation.}
    \label{fig:teaser}
\end{figure*}

Our contributions are summarized as follows:
\begin{itemize}
\item A novel implicit representation parameterized by a recursive function that encodes an arbitrary number of 3D shapes in a shared latent space, while retaining high reconstruction fidelity and requiring up to 99\% less storage space compared to the input mesh models;
\item A curriculum learning method that naturally exploits the octree spatial data structure through a coarse-to-fine optimization scheme;
\item An analysis of the scaling law that correlates latent space dimension, dataset size and 3D reconstruction accuracy and a qualitative analysis of the learned latent space indicating a coarse-to-fine hierarchical structure resulting in reusable latents.
\end{itemize}

%% file: 2-related.tex
\section{Related Work}

\noindent{\textbf{Neural Fields}} are continuous coordinate-based neural networks that encode an underlying property of a scene. The popularity of these representations has increased dramatically as recent results have shown that with enough modeling power coordinate-based networks can be used to encode underlying physical quantities with arbitrary levels of precision~\cite{xie2021neural,park2019deepsdf,mildenhall2020nerf}. Applications of these techniques include modeling 3D shape~\cite{mescheder2019occupancy,peng2020convolutional,park2019deepsdf,sitzmann2020metasdf}, appearance / radiance~\cite{mildenhall2020nerf,yu2020pixelnerf,sitzmann2021light,barron2021mip}, geometry~\cite{yang2021geometry}, semantic information~\cite{kundu2022panoptic}, material properties~\cite{boss2021nerd,hadadan2021neural}, human shape and appearance~\cite{jiang2022neuman}, and robotics~\cite{ortiz2022isdf,adamkiewicz2022vision,yen2022nerf,yen2021inerf,chitta2021neat, simeonov2022neural}. Neural Fields have been used in robotics to represent 3D geometry and appearance with applications in grasping~\cite{breyer2020volumetric,ichnowski2021dex}, trajectory planning~\cite{adamkiewicz2022vision}, object pose estimation and refinement~\cite{zakharov2020autolabeling,irshad2022shapo, zakharov2021single, huang2022ncf}, object and surface reconstruction from sparse and noisy data~\cite{peng2020convolutional,williams2022neural}, multi-modal perception~\cite{gao2021objectfolder}, localization~\cite{moreau2022lens} and SLAM~\cite{sucar2021imap,zhu2022nice}. For an overview of recent methods and applications please consult~\cite{xie2021neural}.

\paragraph{Neural Fields for Shape Representation} These methods represent shapes in the weights of the neural networks, and vary depending on the underlying signal used to encode the 3D space, e.g. occupancy~\cite{mescheder2019occupancy,peng2020convolutional}, Signed Distance Functions~\cite{park2019deepsdf,sitzmann2020metasdf,zakharov2020autolabeling}, density and radiance~\cite{mildenhall2020nerf,barron2021mip}. A second distinction comes from the target domain, with some methods overfitting to a single scene/object~\cite{mildenhall2020nerf,takikawa2021neural,muller2022instant}, and other methods learning a generalizable prior over entire categories of shapes, e.g. as a generalizable latent space of Signed Distance Functions ~\cite{park2019deepsdf,sitzmann2020metasdf,zakharov2020autolabeling}, as a convolutional prior over grid cells~\cite{mescheder2019occupancy,peng2020convolutional} or image pixels~\cite{yu2020pixelnerf}, as weights of a kernel learned from data~\cite{williams2021neural_splines,williams2021neural,liu2022learning}, as an object centric shape~\cite{zakharov2021single} and/or appearance prior~\cite{jang2021codenerf}. The modeling power of these methods can be further improved by modulating the input coordinates with period functions~\cite{sitzmann2020implicit,tancik2020fourier}, while rendering speed, training time and networks size can be improved by factorizing the scene tensor into multiple low-rank components~\cite{chen2022tensorf}, by utilizing multiple small size MLPs~\cite{reiser2021kilonerf}, by training on Sparse Voxel Fields~\cite{liu2020neural} or via multiresolution hash input encoding~\cite{muller2022instant}. A number of methods employ an octree datastructure to guide learning towards occupied areas of space~\cite{yu2021plenoctrees,takikawa2021neural,tang2021octfield}. Closest to our method, NGLOD~\cite{takikawa2021neural} also uses an octree to adaptively fit shapes to multiple Levels-of-Detail (LoDs), however unlike~\cite{takikawa2021neural} our method can represent multiple shapes. Additionally, thanks to our recursive scheme, we only need to store root level latents and not the entire grid as is the case for~\cite{takikawa2021neural}. Recursive parameterizations have also been employed in the context of radiance fields~\cite{yang2021recursive} or for 3D shape representation~\cite{tang2021octfield}. 
Unlike~\cite{tang2021octfield}, we use a lightweight decoder-only recursive architecture to represent high quality shapes as dense oriented point clouds through which we can extract the encoded surface in real-time. Its capacity can be easily extended by simply modifying the size of the input latent vector and it is capable of storing additional attributes (such as material information) at minimal cost.

\noindent{\textbf{Differentiable Rendering}} refers to the ability to render an image and back-propagate training signal from the image back to the underlying representation; for an overview of latest methods please refer to~\cite{kato2020differentiable,tewari2021advances}. This allows 3D representations of scenes to be learned using only 2D supervision~\cite{sitzmann2019scene}, generative models of objects~\cite{mustikovela2021self}, compositionality~\cite{niemeyer2021giraffe,ost2021neural,yang2021learning}, learning from data in the wild~\cite{henzler2021unsupervised,muller2022autorf,reizenstein2021common} or test-time adaptation~\cite{zakharov2020autolabeling}. However, in the context of 3D shape representation, extracting the underlying surface from the implicit field typically involves expensive operations such as volume rendering~\cite{lombardi2019neural}, sphere tracing~\cite{liu2020dist} or ray marching~\cite{niemeyer2020differentiable}. Our method maintains differentiabilty with respect to the input and can thus be optimized given partial 3D data as well as 2D images and additionally we output the underlying surface by design, foregoing the need for expensive operations for surface extraction. 

    

    


%% file: 3-method.tex
\section{Methodology}
\label{sec:method}

\paragraph{Preliminaries} Our approach takes as input a set of shapes $S=\{S_1, \dots, S_\numObjects\}$ and learns a space of implicit surfaces that represents the input. Each shape $S_\numObjects$ is represented by an oriented point cloud consisting of points $\{\surfacePointGT_i \in \mathbb{R}^3\}_{i=1}^{|{S}_\numObjects|}$ and associated normals $\{\normalGT_i \in \mathbb{R}^3\}_{i=1}^{|{S}_\numObjects|}$, where $|S_\numObjects|$ denotes the number of points in $S_\numObjects$.

\begin{figure*}[t]
	\centering
	\includegraphics[width=1\linewidth]{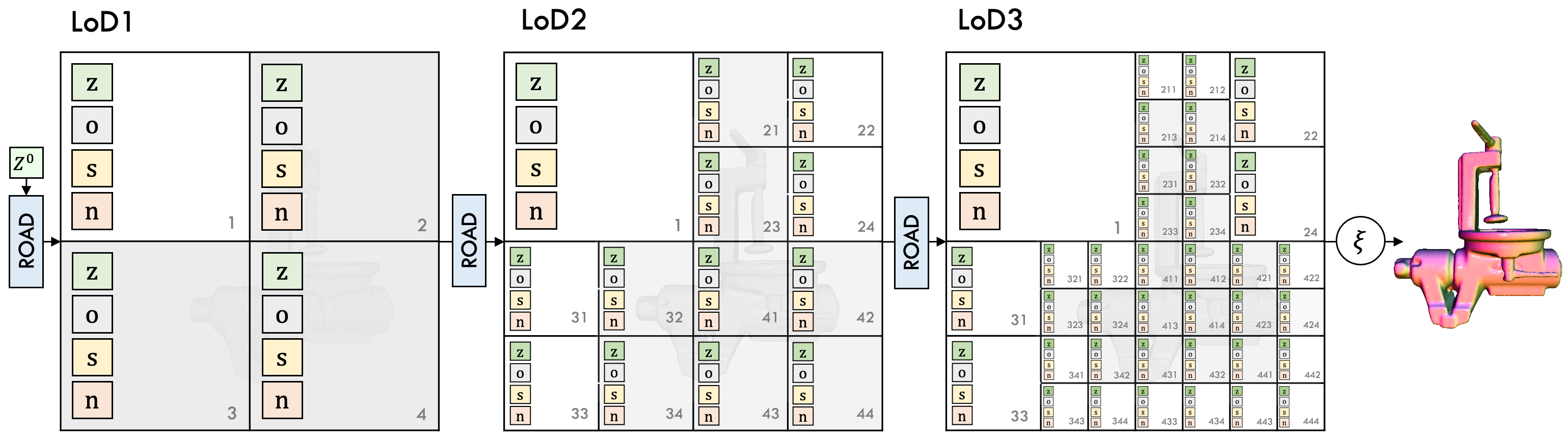}
	\caption{Our method extracts object surfaces by performing octree traversal. Starting from a latent vector of the parent cell \methodspace extracts latent vectors for all children cells together with their occupancy, local SDF and surface normal estimates. It efficiently extends to large datasets while retaining high surface reconstruction quality.}
	\label{fig:method}
\end{figure*}


\paragraph{Formulation} Our method represents each shape with a $\latentDim$-dimensional latent vector denoting the root node of an octree which is traversed recursively, up to a predefined Level-of-Detail (LoD) $\maxLoD$.
We seek to regress three functions to recursively reconstruct 3D geometry, parameterized by neural networks: $\phi:\mathbb{R}^{\latentDim} \to \mathbb{R}^{8\latentDim}$ for latent subdivision; $\psi:\mathbb{R}^\latentDim\to\mathbb{R}^5$ for mapping the latent space to surface geometry (occupancy, signed distance to surface, surface normal); and $\xi:\mathbb{R}^7\to\mathbb{R}^3$ for zero iso-surface projection. The hyperparameter $\maxLoD$, the dimension of the latent space, is linked to the capacity of our representation as a function of the dataset size. Given a latent vector $\latent^\LoD\in\mathbb{R}^\latentDim$, where $0 \leq \LoD\leq \maxLoD$ denotes the LoD of the latent vector, we define:
\begin{equation}
    \phi(\latent^\LoD) \to \{\latent^{\LoD+1}_i\}_{i=1}^{8}
\label{eq:phi}
\end{equation} 
as the function that performs a traversal of the latent space given input latent $\latent^\LoD$ and outputs a latent vector $\latent^{\LoD+1}_i=\lfloor\phi(\latent^\LoD)\rfloor_i$ for each of the $8$ possible children, where $\lfloor\phi\rfloor_i$ denotes the ith output of $\phi$. 

By definition, $\phi$ projects back to the latent space of dimension $\latentDim$, i.e. $\latent^{\LoD+1}_i\in\mathbb{R}^{\latentDim}$, forcing the resulting latent to encode both high and low level information. This formulation allows \methodspace to simply pass back the resulting latent to our recursive function, i.e. through $\phi(\latent^{\LoD + 1})$, while propagating coarser, higher-level information along the latent space. A latent $\latent^0$ at the lowest LoD therefore encodes the geometry of an entire object. Unlike other octree-based methods~\cite{takikawa2021neural} that store all the octree latents, our formulation allows us to record only the root level latents for each input shape, i.e. $\latentSet^0=\{\latent_\objectIndex^0\}_{\objectIndex=0}^{\numObjects}$.
We provide ablations for other formulations of $\phi$ in Section \ref{sec:experiments}. 

We employ $\psi$ to map any latent vector to underlying surface geometry as follows: 
\begin{equation}
    \psi(\latent^\LoD)=\{\occupancyPrediction^\LoD,\sdfPrediction^\LoD,\normalPrediction^\LoD\}
\end{equation}

The output of $\psi$ consists of $\occupancyPrediction^\LoD \in (0, 1)$ --- the occupancy estimate denoting whether to continue expanding this cell further; $\sdfPrediction^\LoD \in (-1, 1)$ --- the constrained signed distance value from the center of the cell to the surface of the object; and $\normalPrediction^\LoD \in \mathbb{R}^3$ --- the surface normal vector. To extract the surface information at a particular LoD $\LoD$ starting from a root latent $\latent^0$ we perform a tree traversal as follows:

\begin{equation}
\label{eq:recursion}
    \psi(\underbrace{\phi (\dots (\phi}_{\text{$\LoD$ times}}(\latent^0)))=\{\occupancyPrediction^\LoD,\sdfPrediction^\LoD,\normalPrediction^\LoD\}
\end{equation}

\begin{wrapfigure}{r}{0.4\textwidth}
	\centering
	\includegraphics[width=1\linewidth]{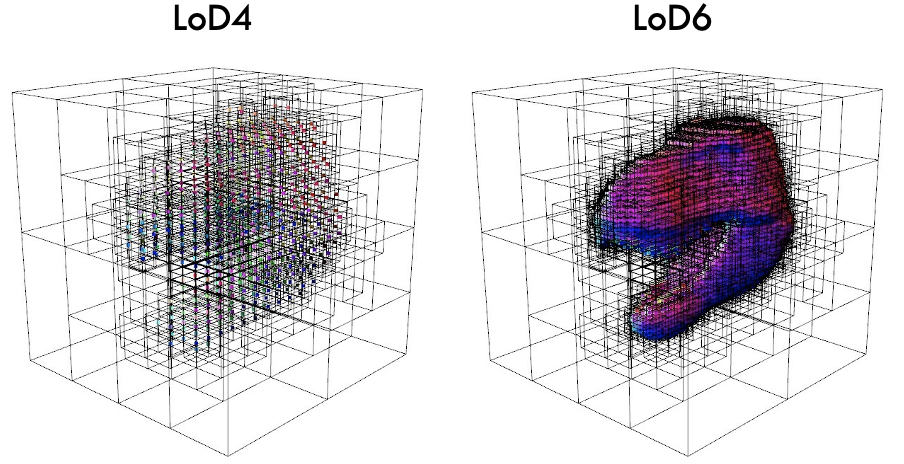}
	\caption{\methodspace enables surface points to be directly extracted at different levels of detail.}
	\label{fig:lods}
\end{wrapfigure}

Note that for clarity we omitted the notation $\lfloor\cdot\rfloor$ in Eq.\ref{eq:recursion}, however after each subdivision the appropriate child latent is selected, according to the desired branch of the octree to be expanded. We highlight that unlike other SDF based methods that implicitly encode surfaces~\cite{park2019deepsdf}, our method does not take as input Euclidean coordinates, and the mapping between different LoD levels is achieved entirely via the learned latents $\latent^\LoD \to \latent^{\LoD+1}$ which are forced to encode both local structure as well as global shape information. Moreover, unlike grid-based representations~\cite{mescheder2019occupancy,peng2020convolutional} we fully exploit the sparse nature of the underlying octree representation, using the occupancy $o_i$ to recursively expand only occupied cells. 

\begin{wrapfigure}{r}{0.4\textwidth}
    \centering
    \includegraphics[width=0.4\textwidth]{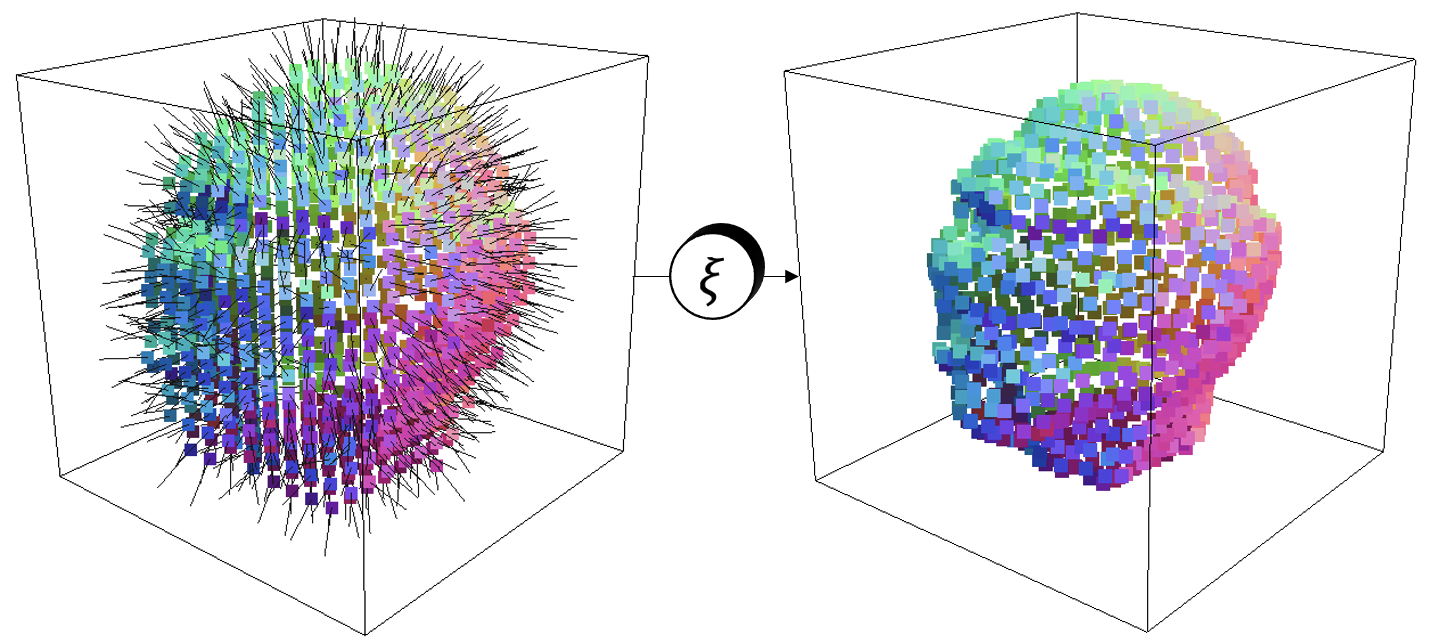}
	\caption{Zero-isosurface projection.}
	\label{fig:iso_projection}  
\end{wrapfigure}

\paragraph{Shape Recovery}
To recover the final shape, we perform the zero-isosurface projection (see Fig.~\ref{fig:iso_projection}), using distance $\sdfPrediction_i$, surface normal estimates $\normalPrediction_i$, and voxel cell center coordinates $\voxelCoordinate_i$ at the desired LoD.
The voxel centers are determined and tracked whenever we subdivide a cell. 
To extract the object surface at LoD $\LoD$, we use the following equation: $\estimatedSurfacePoint_i^\LoD$ = $\xi(\voxelCoordinate_i^\LoD, \sdfPrediction_i^\LoD, \normalPrediction_i^\LoD) = \voxelCoordinate_i^\LoD - \sdfScaling^\LoD \normalPrediction_i^\LoD \sdfPrediction_i^\LoD$, where $\sdfScaling^\LoD$ is a value that scales $\sdfPrediction^\LoD$. In our experiments we use a scaling factor equal to the voxel size at LoD $\LoD$, i.e. $\sdfScaling^\LoD = 2/(2^\LoD)$.
Once the projection is performed for all the query points at a particular LoD, we get a dense surface point cloud that is differentiable back to the input latent vector $\latent^0$. This property allows us to perform optimization to complete partial shapes based on the prior encapsulated in the network. An overview of our complete pipeline is provided in Algorithm~\ref{alg:octree}.




\input{tables/extraction2}

\begin{figure*}[t]
	\centering
	\includegraphics[width=1\linewidth]{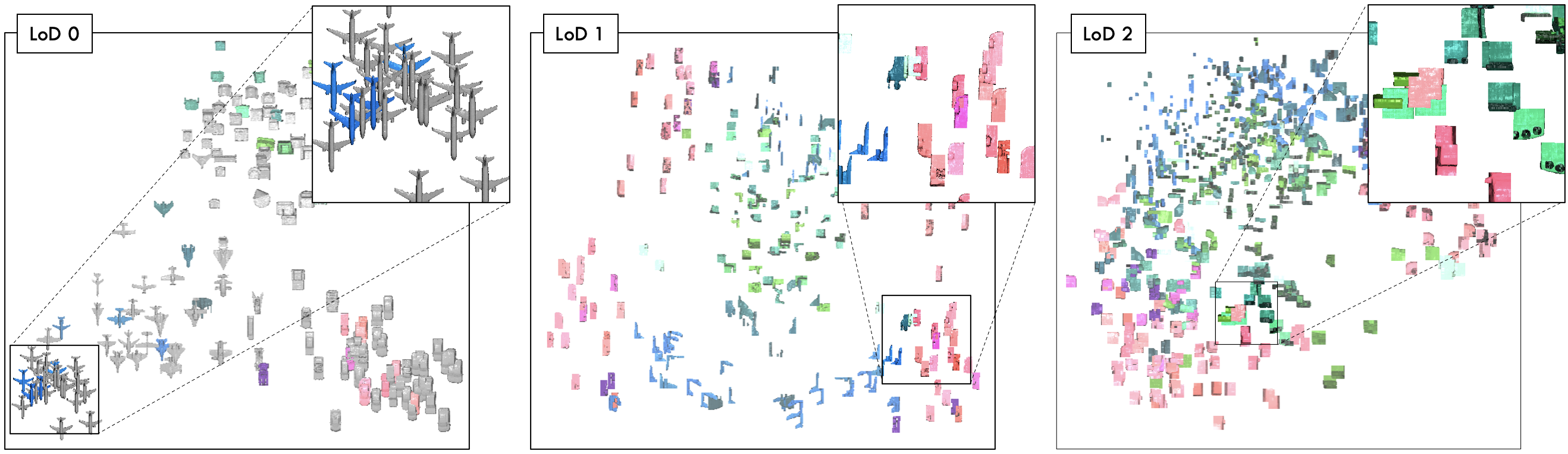}
	\caption{\textbf{Latent space structure.} We use principle component analysis to visualize the encoded geometries of the ShapeNet150 dataset in two dimensions. Object color is related to object instance, with objects of the same class having similar colors, and is carried through each LoD. For visualization purposes, grey objects in LoD 0 are not propagated to the higher LoDs. Similar latent vectors encode similar geometries resulting in a clear class separation at LoD 0. Similar areas of the projected latent space become increasingly shared by the different classes at higher LoDs, suggesting that our approach efficiently encodes object geometry by learning common geometric primitives.}
	\label{fig:latent_pca}
\end{figure*}

\paragraph{Architecture and Training}
We parameterize the functions $\phi, \psi$ with a single MLP with parameters $\theta$, choosing a SIREN-based~\cite{sitzmann2019siren} network as periodic activation functions have been shown to be more capable at representing fine details.
Our MLP uses a single layer encoder and multiple 2-layer decoder heads to output occupancy $\occupancyPrediction$, SDF $\sdfPrediction$, surface normals $\normalPrediction$. All hidden layers are 512-dimensional.
We train our method by optimizing both the latent vectors $\latentSet$ and the parameters $\theta$ of the MLP using the Adam solver with a learning rate of $6\times10^{-5}$.

To supervise training, we define losses for each of the decoder levels at every LoD. Occupancy loss $\mathcal{L}_o$ as a binary cross entropy, whereas SDF $\mathcal{L}_s$ and surface normals $\mathcal{L}_n$ losses minimize the $l_2$ distance between respective predictions and ground truth values. The final loss is formulated as:
\begin{equation}
    \mathcal{L} = \sum_{\LoD \in \maxLoD} w_{o} \mathcal{L}_o^\LoD + w_s^\LoD \mathcal{L}_s^\LoD + w_n \mathcal{L}_n^\LoD,
\end{equation}
where $w_o = 1$, $w_s^d$ is a function returning an inverse voxel radius for level $\LoD$, and $w_v = 0.1$. 


\paragraph{Curriculum Learning}
When it comes to training on large datasets or on datasets with high resolution models requiring high LoDs, the vanilla training procedure requires much more time to converge as opposed to training on simpler datasets. To alleviate this problem we introduce a curriculum training procedure. Instead of initiating training from the desired final LoD, we change the first LoD to be lower (we use LoD 3 in our experiments) and keep track of the mean of predicted occupancy confidences
at a given level $\LoD$. Lower LoDs are faster to train on due to a smaller number of latents to optimize. Moreover, it makes it faster to sample random training points at each iteration, thus further accelerating the training procedure. To compute our confidence score, we first take a softmax over two occupancy values for all estimated voxels at a given LoD, and then take a max value between all the pairs. If our average confidence $\probabilityThreshold$ is high enough in either occupancy or non occupancy, we jump to the next LoD and repeat this procedure until the final LoD is reached. In our experiments, we show that this simple technique helps to accelerate the training process especially when training on large and high resolution datasets. We set $\probabilityThreshold=0.95$ in our experiments.



%% file: tables/extraction2.tex
\begin{algorithm}[b]
	\KwIn{ 
			$\maxLoD$ maximum recursion depth,
			$\latentSet^0=\{\latent^0\}$ object latent vector
		}
		\KwOut{ 
			$\textbf{p} \in \mathbb{R}^3$ surface points,
		}
		
		\tcc{Recursively subdivide voxels until desired LoD is reached}
		\For{$ \LoD \in \{1, \ldots, \maxLoD\} $}{
		    $\latentSet^{\LoD} \leftarrow \{\}$ 
		    
		    \For {$\latent^{\LoD-1}$ in $\latentSet^{\LoD-1}$}{
		        $\{\latent_i^\LoD,o_i^\LoD,s_i^\LoD,\normal_i^\LoD\}_{i=1}^8 \leftarrow \text{\method}(\latent^{\LoD-1})$ \tcp*{recursive subdivision}
		        $\latentSet^{\LoD}  \leftarrow \latentSet^{\LoD} \cup \{\latent_i^\LoD |\occupancyPrediction_i^\LoD \geq \probabilityThreshold
		        \}$ \tcp*{add occupied latents}
		    }
		    
			
		}
		\tcc{Extract object shape}
		
		\For {$\latent^{\LoD}$ in $\latentSet^{\LoD}$}{
		    $\textbf{p} \leftarrow \textbf{p} \cup \xi(\text{\method}(\latent^\LoD))$ 
		}
		\KwRet{$\textbf{p}$}
		
		\caption{Octree-based recursive surface extraction}
		\label{alg:octree}
	\end{algorithm}

%% file: 4-experiments.tex
\section{Experiments}
\label{sec:experiments}

To demonstrate the 3D reconstruction and compression capabilities of our approach we run a number of experiments on the ShapeNet~\cite{chang2015shapenet}, Thingi10K~\cite{zhou2016thingi10k}, Google Scanned Objects~\cite{downs2022google}, and the AccuCities~\cite{AccuCities} datasets. We report reconstruction metrics as measured by the Chamfer distance (multiplied by $10^3$) as well as interesection over union over points uniformly sampled in the bounding volume of the ground truth shape. 

\input{tables/reconstruction}

\begin{wrapfigure}{r}{0.45\textwidth}
    \centering
    \includegraphics[width=0.45\textwidth]{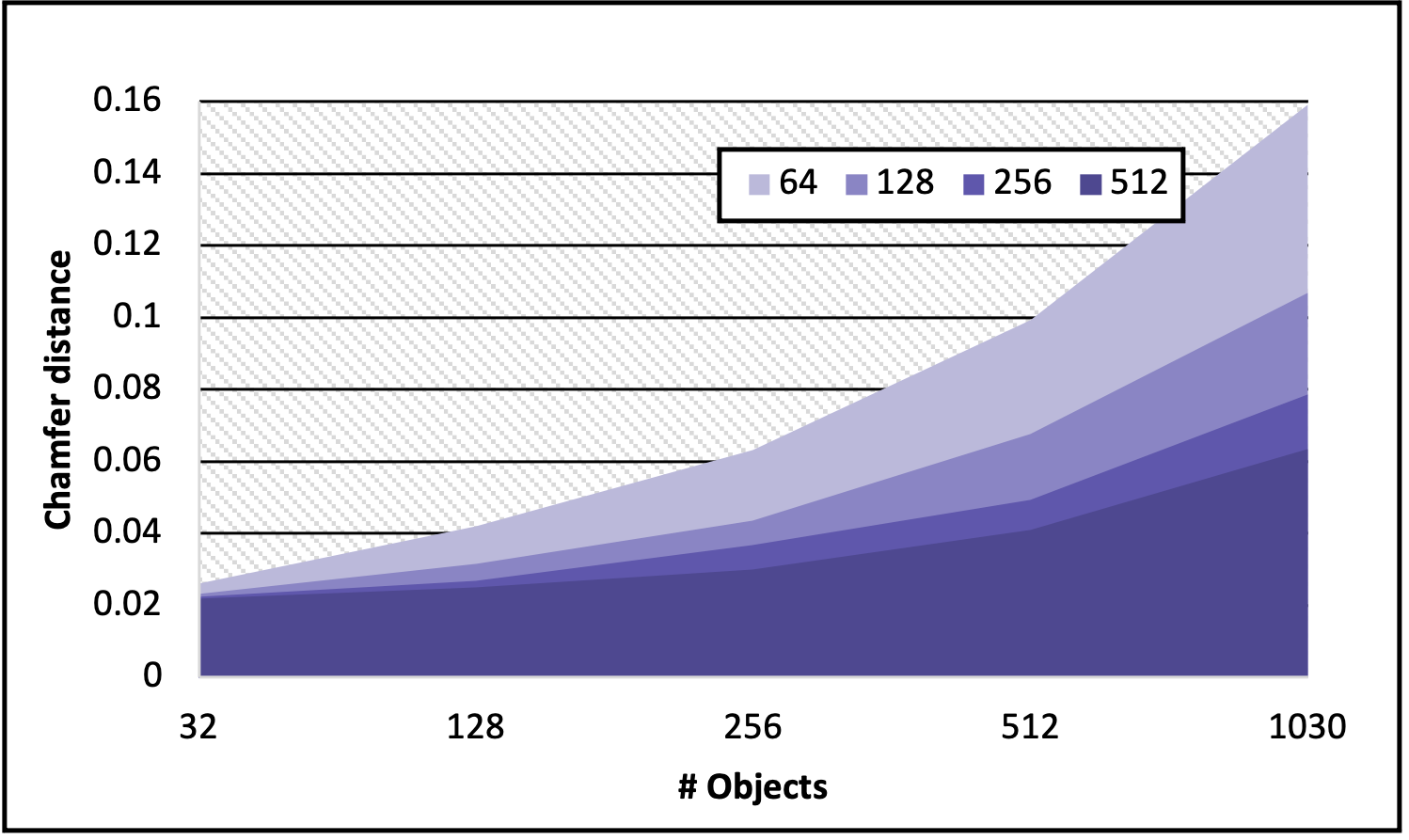}
	\caption{Chamfer distance vs data quantity for different latent vector sizes.}
	\label{fig:scaling} 
\end{wrapfigure}

\paragraph{Reconstruction}
We follow the protocol of~\cite{takikawa2021neural} and train on a subset of 150 objects from ShapeNet~\cite{chang2015shapenet} and a subset of 32 objects from Thingi10K~\cite{zhou2016thingi10k}. We report results in Table~\ref{tab:reconstructionz}. We note that the baselines we compare against train one model for each shape in the dataset (i.e.~\cite{takikawa2021neural} trains 32 networks for each object in Thingi32). Owing to the efficient recursive decoding scheme implemented by our method, we can train a single model for each dataset and still be competitive in terms of network size and reconstruction accuracy. We set the latent size $\latentDim$ to 64 for Thingi32 and to 96 for ShapeNet150 respectively. Indeed, our network achieves state-of-the-art reconstruction results of ${0.036}$ Chamfer distance and an IoU of more than ${94\%}$ on ShapeNet150 with a network of size 3.8MB: this amounts to a compression of more than $99\%$, as the original mesh dataset measures 630MB. We report a similar compression ratio on Thingi32 (473MB) while achieving state-of-the-art reconstruction accuracy of ${0.017}$ Chamfer distance and a competitive IoU of $98.7\%$.

We further explore the reconstruction capabilities of our method by encoding a model from the AccuCities~\cite{AccuCities} dataset: a neighborhood from London consisting of 1.9 million triangles and requiring 252MB of disk space. We set the latent size $\latentDim$ to 512. Qualitative results are shown in Fig.\ref{fig:teaser} with additional images in the supplementary. Quantitatively we achieve a Chamfer distance of 0.04 when comparing against the ground truth model, while using a network of size 11 MB. 



\paragraph{Scaling to larger datasets}
For this experiment we use the entire Google Scanned Objects~\cite{downs2022google} dataset consisting of a total of 1030 object models. We introduce training splits of different sizes (32, 128, 256, 512, 1030) to study the relation between the dataset complexity and the latent vector size, a novel hyperparameter specific to our formulation. We restrict our analysis into the scaling properties of our network to the dimension of the latent space, and mention that other methods that are generally applicable to machine learning models (i.e. number of layers, training schedule, etc.) can be used to further tweak the performance of our method. Our results are recorded in Fig.~\ref{fig:scaling}: for each split we train networks with increasing latent vector sizes (64, 128, 256, 512) and record the resulting Chamfer distance by comparing the reconstructed models with the ground truth models of that specific split. We note a strong correlation between dataset complexity and latent size. Specifically, our results indicate that our method can efficiently scale to an increasing number of shapes by only modifying the latent vector size while keeping the network parameterization intact.

\begin{figure}[t]
	\centering
	\includegraphics[width=1\linewidth]{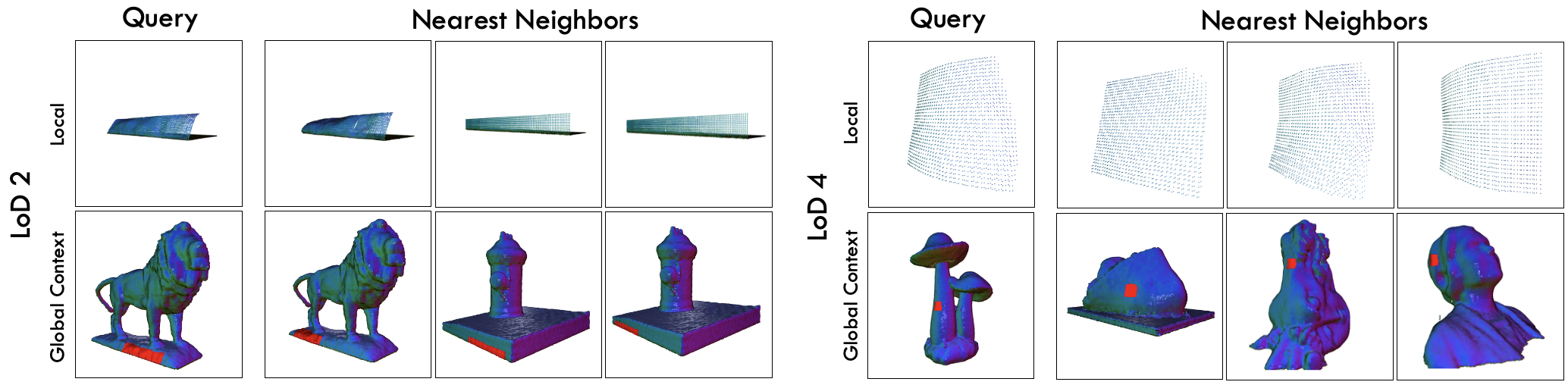}
	\caption{\textbf{Nearest neighbors.} We visualize the geometry encoded by example latents, as well as the geometry encoded by the nearest neighbors within the same LoD across all objects in the Thingi32 dataset, determined by the Euclidean distance metric on the latent vector space (top row). We color the points in the octree cell corresponding to the latent, showing that our approach enables latents with similar shapes to be used at different global positions (bottom row).}
	\label{fig:nearest_neighbor}  
	\vspace{-4mm}
\end{figure}

\paragraph{Latent Space Analysis}
To qualitatively analyze the properties of the learned latent space, we project the latent space at specific LoDs into two dimensions via principle component analysis to visualize the encoded geometries of the ShapeNet dataset, as shown in Fig. \ref{fig:latent_pca}.
\begin{wrapfigure}[18]{r}{0.45\textwidth}
    \centering
    \includegraphics[width=0.45\textwidth]{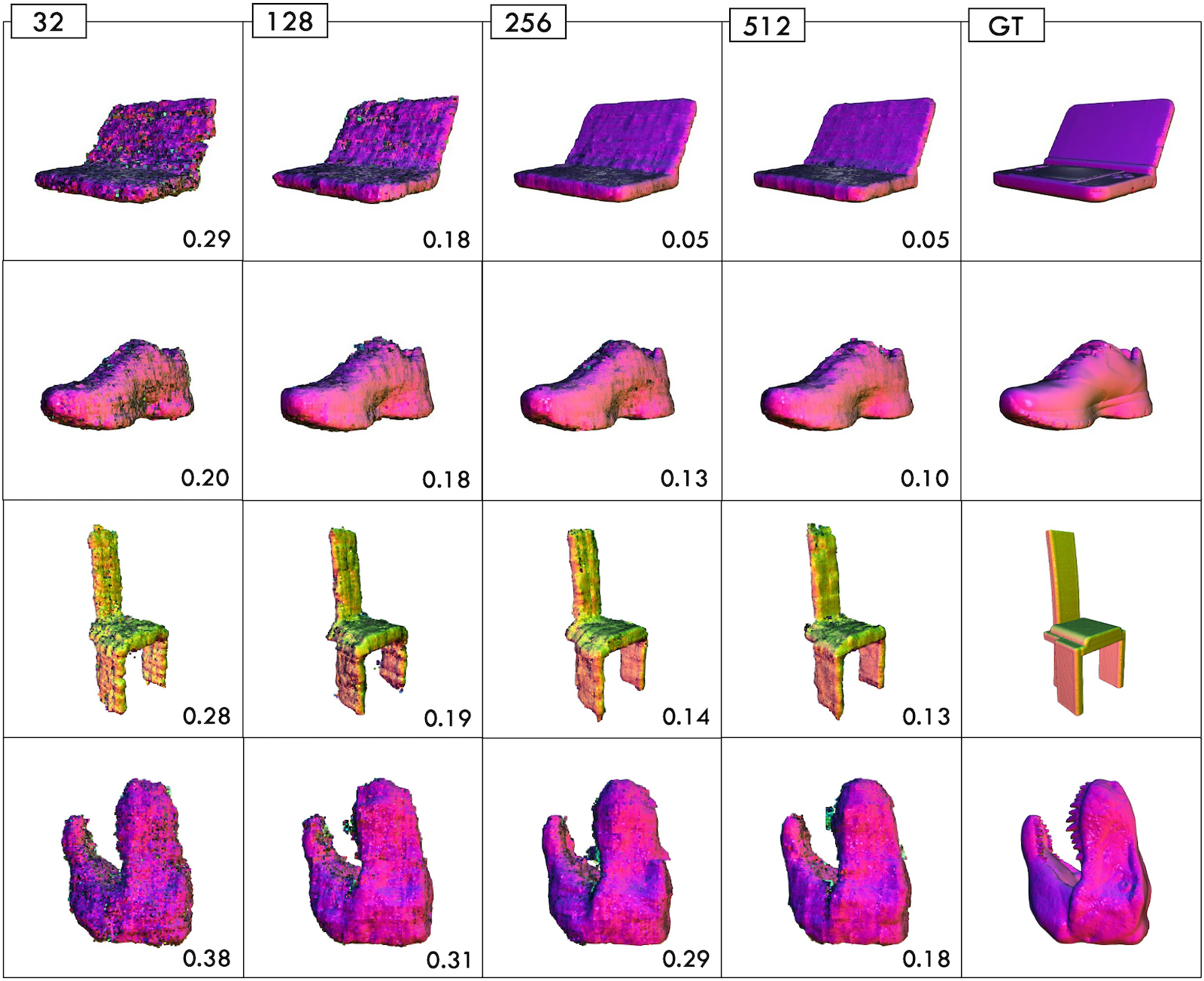}
	\caption{\textbf{Generalization.} We demonstrate our method's increasing generalization ability when trained on many objects.}
	\label{fig:generalization} 
\end{wrapfigure}
We observe that objects of the same class are spatially close in the projected space at the top level LoD, demonstrating that similar latent vectors encode similar geometries. Furthermore, at higher LoDs, similar areas of the projected latent space are increasingly shared by the different classes, suggesting that our approach efficiently encodes object geometry by learning geometric primitives common in the dataset.

We also visualize the nearest neighbors of specific latent vectors from the network trained on Thingi32 to LoD 9 in Fig.~\ref{fig:nearest_neighbor}. We show that similar latent vectors can represent the local geometry at different object coordinates, without explicit positional encoding. Increasing the LoD also intuitively reduces the geometric complexity represented by the latent, as seen by the 3D edge feature from LoD 2 and the oriented patch feature from LoD 4. 



\paragraph{Generalization}
In this experiment we demonstrate the generalization capabilities of our method. We take four networks from the data compression experiment each trained on a different split of the Google Scanned Objects dataset (32, 128, 256, and 512 objects) and optimize latent vectors to fit unseen models from the same as well as other datasets (Thingi32, ShapeNet150) while keeping network weights frozen. The results are shown in Fig.~\ref{fig:generalization}. Reconstruction quality plotted in terms of Chamfer distance shows increasing generalization capability for networks trained on more models.

\paragraph{Surface Extraction}
By design, our method differentiably extracts object surfaces in real time with minimal memory overheads. In comparison, NGLOD~\cite{takikawa2021neural} or other SDF-based~\cite{park2019deepsdf} methods require either an expensive sphere ray tracing or non-differentiable marching cubes to extract surface. Table~\ref{tab:surface_extraction} compares inference times when extracting the object surface at different levels of density. For our method we perform inference up to LoDs 6, 7 and 8 respectively, which corresponds to approximately 20000, 80000 and 300000 surface points, respectively for the Thingi32 models. We compare against a marching cubes baseline from~\cite{park2019deepsdf} and a sphere tracing baseline from~\cite{takikawa2021neural}, and we iterate until the desired number of surface points is sampled. For a fair comparison with our method we extract 20000, 80000 and 300000 using both sphere tracing and marching cubes. We note that our method extracts the object surface up to 3 orders of magnitude faster that the sphere tracing baseline; our method requires \textit{a total of 1-2 seconds} to extract object surfaces for the \textit{entire} ShapeNet150 dataset. The marching cubes baseline, while faster than sphere tracing, is still not real-time capable and not differentiable. Finally, we observe that the results for this experiment were obtained on a single A6000 GPU, without any optimization.


\begin{table}[!t]
    \begin{minipage}{.48\textwidth}
    \centering
        \setlength{\tabcolsep}{8pt}
        {\scriptsize
            \begin{tabular}{c|ccc}
            \toprule
            Superposition & Storage (MB) & gIoU  & Chamfer \\
            \midrule
            Direct &   3.2    &    98.7   &  0.017\\
            Addition &   3.2    &   96.9    & 0.039 \\
            Concatenation & 17.9  &  99.4     &  0.013\\
            \bottomrule
            \end{tabular}%
            }
             \caption{\textbf{Latent Fusion.} Here we compare different ways to propagate latent vectors to the next LoDs.}
          \label{tab:fusion}%
    \end{minipage}%
    \hfill
    \begin{minipage}{.48\textwidth}
    \centering
        \setlength{\tabcolsep}{8pt}
        {\scriptsize
            \begin{tabular}{c|c|c|c}
            \toprule
            Surface density &  Low   &  Medium  & High  \\
            \midrule
            Sphere tracing & 5 min & 6 min & 10 min \\
            Marching cubes & 0.1 s & 0.9 s & 6 s \\
            Ours  & \textbf{11 ms} & \textbf{13 ms} & \textbf{17 ms} \\
            \bottomrule
            \end{tabular}%
            }
             \caption{\textbf{Inference time.} Our method extracts object surfaces in real time, significantly outperforming the state of the art.}
          \label{tab:surface_extraction}%
    \end{minipage} 
    \vspace{-8mm}
\end{table}



\paragraph{Latent Vector Fusion}
As described in Section~\ref{sec:method} and in Fig.~\ref{fig:method}, our method propagates information through the latent space via the recursive function $\phi$ (see Eq.\ref{eq:phi}). Here we explore different forms of latent subdivision: addition and concatenation, with the complete definitions provided in the supplementary. 
While addition does not change the dimension of the latent vector $\latentDim$, it explicitly defines how information is propagated from parent to child latent (i.e. via addition). Conversely, concatenation makes $\latentDim$ increase with each recursion level, and information is directly copied as we traverse the latent space. This introduces significant modifications to the underlying neural network architecture, requiring specialized networks at each LoD. The results of this ablative analysis are summarized in Table~\ref{tab:fusion}. As expected, contcatenation serves as an upper bound for performance and it achieves the highest performance but requires $5\times$ more storage space, and a more complicated formulation. Although similar in formuation to direct regression, the addition version of our method achieves poor results, which we attribute to the artificial constraint imposed on how information is propagated through the latent space. 

%% file: tables/reconstruction.tex
\setlength{\tabcolsep}{6pt}
\begin{table}[t]
  \centering
    \resizebox{1\textwidth}{!}{
    \begin{tabular}{c|ccc|ccc}
    \toprule
    \multirow{2}[4]{*}{Method} & \multicolumn{3}{c|}{ShapeNet150} & \multicolumn{3}{c}{Thingi32} \\
\cmidrule{2-7}          & Storage (MB) ($\downarrow$) & gIoU ($\uparrow$)  & Chamfer ($\downarrow$) & Storage (MB) ($\downarrow$) & gIoU ($\uparrow$)  & Chamfer ($\downarrow$) \\
    \midrule
    DeepSDF~\cite{park2019deepsdf} & 1052.6 & 86.9  & 0.316 & 224.6 & 96.8  & 0.053 \\
    FFN~\cite{tancik2020fourier}   & 301.6 & 88.5  & 0.077 & 64.3  & 97.7  & 0.033 \\
    SIREN~\cite{sitzmann2019siren} & 151.3 & 78.4  & 0.381 & 32.3  & 95.1  & 0.077 \\
    Neural Implicits~\cite{davies2020overfit} & 4.4 & 82.2  & 0.500 & \textbf{0.9} & 96.0  & 0.092 \\
    NGLOD~\cite{takikawa2021neural} & 185.4 & 91.7  & 0.062 & 39.6  & \textbf{99.4} & 0.027 \\
    \midrule
    Ours / LoD6 & \multirow{4}[1]{*}{\textbf{3.8}} &   86.3    & 0.175 & \multirow{4}[1]{*}{3.2} & 96.4  & 0.138 \\
    Ours / LoD7 &  &   94.2    & 0.067 &  & 98.4  & 0.045 \\
    Ours / LoD8 &       & 94.9 & 0.041 &       & 98.7  & 0.022 \\
    Ours / LoD9 &       & \textbf{94.9} & \textbf{0.036} &       & 98.7  & \textbf{0.017} \\
    \bottomrule
    \end{tabular}%
    }
    \caption{\textbf{Shape Reconstruction}. This table shows reconstruction and compression comparisons against two datasets. As opposed to the baselines, our method trains a single model for the entire dataset, while still outperforming them in terms of reconstruction quality.}
    \label{tab:reconstructionz}%
    \vspace{-4mm}
\end{table}%

%% file: 5-conclusion.tex
\section{Discussion}

\paragraph{Limitations and Future Work}
Our representation currently only supports 3D geometry. For future work, we would like to explore its extension to other modalities (object color and material properties) as well as representations (images and radiance fields~\cite{mildenhall2020nerf}). Another interesting direction that could be explored is a combination of our pipeline with different downstream tasks (object detection and pose estimation). Our representaton is fully differentiable and  thus allows the propagation of useful 3D gradients for shape optimization given partial information. 


\paragraph{Conclusion}
We presented a novel recursive implicit representation to effectively represent and compress 3D geometry by framing it as the traversal of an implicit octree in a learned latent space. It extracts geometry in real-time and scales to large datasets while retaining high reconstruction quality. As a result, we outperform state-of-the-art reconstruction results on the ShapeNet150 and Thingi32 datasets, even when compared to methods training a single network per model. Our analysis of the representation explores the structure of the latent space and presents a scaling law defining a relationship between latent space dimension, dataset size and reconstruction accuracy.

%% file: 6-supplementary.tex
\section{Latent Vector Fusion}
\label{latent_vector}

In this section we provide additional details to complement the \textit{Latent Vector Fusion} experiments presented in the main paper, with results summarized in Table 2 in the main paper. 

Recall the latent subdivision function, $\phi:\mathbb{R}^{\latentDim} \to \mathbb{R}^{8\latentDim}$, defined as $\phi(\latent^\LoD)= \{\latent^{\LoD+1}_i\}_{i=1}^{8}$ (cf. Equation 1 in the main paper). Thus, $\phi$ directly regresses the next level latents.  

\paragraph{Addition} We define $\phi_{add}:\mathbb{R}^{\latentDim} \to \mathbb{R}^{8\latentDim}$ as: $\phi_{add}(\latent^\LoD)=\{\latent^\LoD + \latent^{\LoD+1}_i\}_{i=1}^{8}$. $\phi_{add}$ maintains the dimensionality of the latent space during traversal, as does the original formulation of $\phi$, i.e. $\latent^\LoD\in\mathbb{R}^{\latentDim}$ and $\latent^{\LoD+1}_i\in\mathbb{R}^{\latentDim}$.

\paragraph{Concatenation} In this case, the signature of the octree traversal function in latent space changes from one LoD to the next. This introduces significant modifications to the underlying neural network architecture, requiring specialized networks at each LoD. Specifically, at LoD $m$, we define $\phi_{concat}:\mathbb{R}^{\LoD \times \latentDim} \to \mathbb{R}^{(\LoD + 1) \times 8\latentDim}$ as: $\phi_{concat}(\latent^\LoD) = \{\text{concatenate}(\latent^\LoD,\latent^{\LoD+1}_i)\}_{i=1}^{8}$, i.e. $\latent^{\LoD}\in\mathbb{R}^{\LoD \times \latentDim}$ and $\latent^{\LoD+1}_i\in\mathbb{R}^{\latentDim}$. 
\section{Evaluation Details}
\label{evaluation_details}

Following NGLOD's implementation~\cite{takikawa2021neural}, we uniformly sample $2
^{17} = 131 072$ points within the unit cube for gIoU computation. To get occupancy estimates, we recover an object mesh using Poisson surface reconstruction~\cite{kazhdan2006poisson} as implemented in Open3D~\cite{zhou2018open3d}. Similarly, we sample $2^{17}$ points on the ground truth mesh for Chamfer distance computation. We use Pytorch3D~\cite{ravi2020accelerating} Chamfer distance implementation. 
We also use the original NGLOD implementation as well as NGLOD’s re-implementations of presented baselines: DeepSDF, FFN, SIREN, and Neural Implicits.




\section{Extended Generalization Experiment}

\begin{figure}
\centering
\begin{subfigure}{.525\textwidth}
  \centering
  \includegraphics[width=1\linewidth]{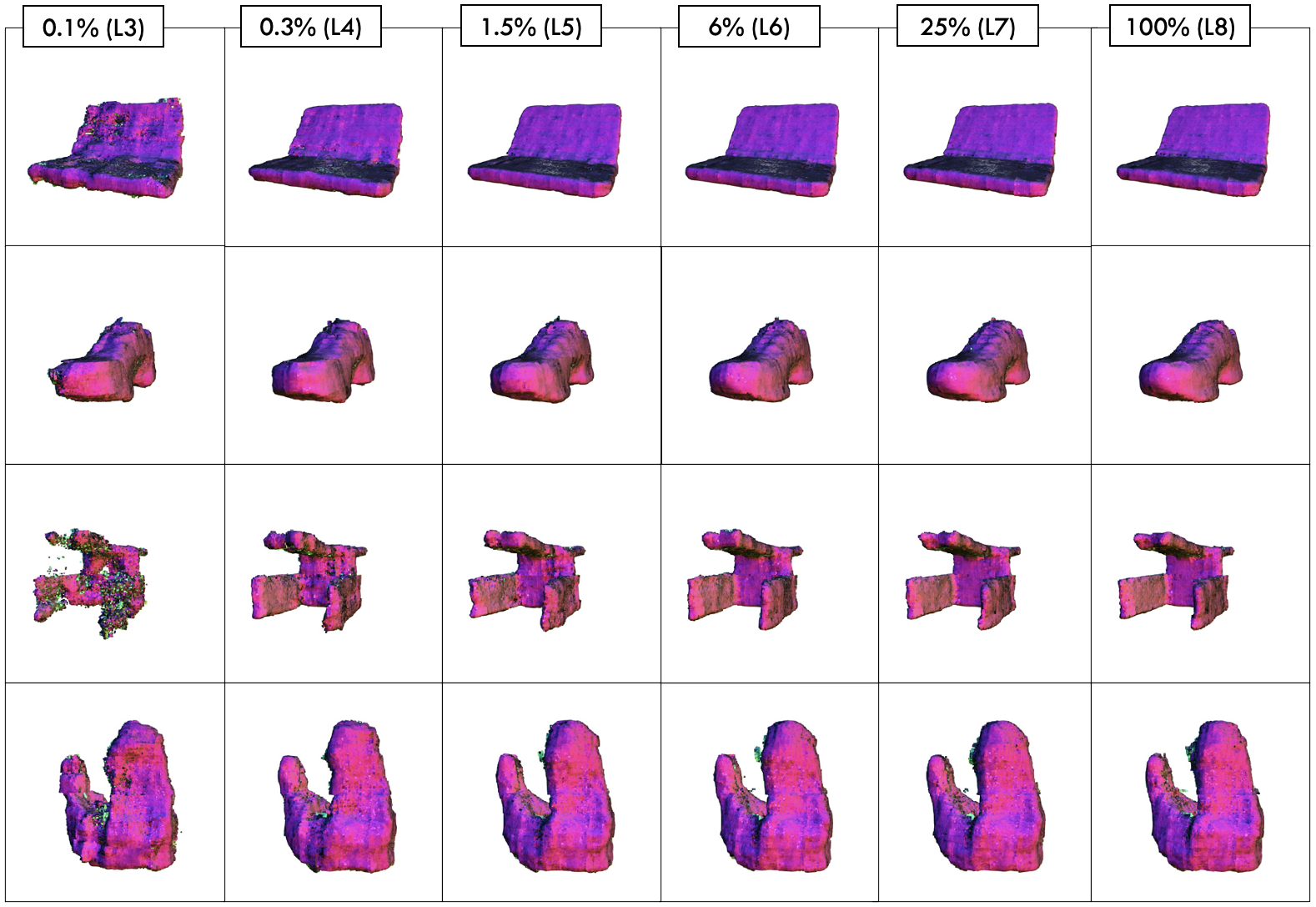}
  \caption{Sparsity ablation}
  \label{fig:sparsity}
\end{subfigure}%
\hfill
\begin{subfigure}{.44\textwidth}
  \centering
  \includegraphics[width=1\linewidth]{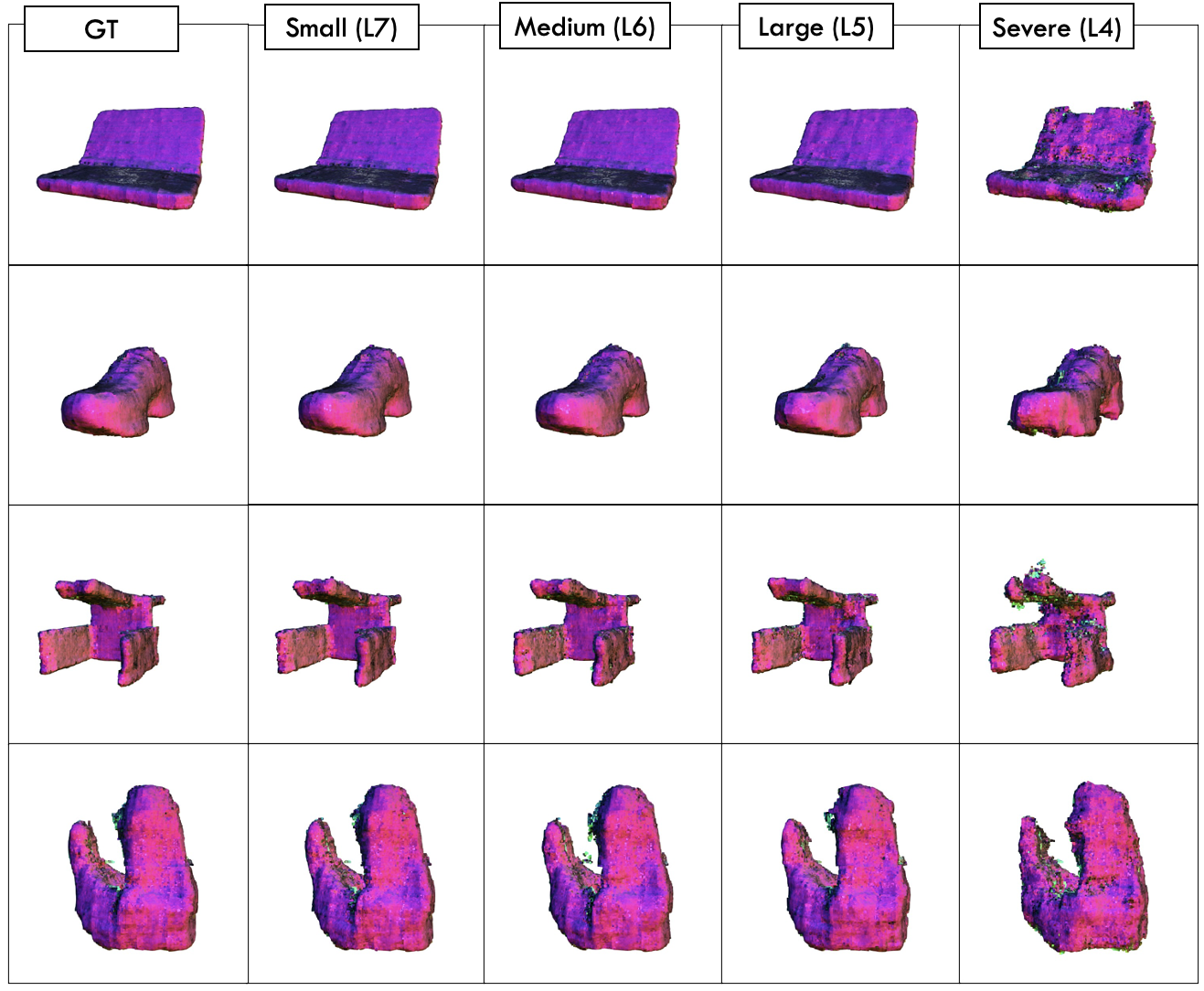}
  \caption{Noise ablation}
  \label{fig:noise}
\end{subfigure}
\caption{Extended Generalization Experiment}
\label{fig:test}
\end{figure}

We extend the generalization experiment to include multiple grades of sparsity and noise. We use a network trained on 512 dense objects of the Google Scanned Objects dataset~\cite{downs2022google}. We then optimize latent vectors to fit unseen objects from the generalization experiment. Given a ground truth dense unseen object point cloud we apply sparse supervision computed from the dense GT point cloud to estimate the surface geometry. We optimize the pre-trained ROAD to a lower LoD, i.e. provide a coarser supervisory signal than the network was trained to. We then extract the surface to the highest LoD. LoDs 3 through 7 represent approximately 0.1\%, 0.3\%, 0.15\%, 6\%, and 25\% of the supervision at LoD8. We observe that even in the case of optimizing only to LoD3, our method is still able to converge to reasonable shapes (see Fig.~\ref{fig:sparsity}).

We additionally demonstrate how noise affects the generalization performance. Similarly to the  sparsity experiments above, we optimize the pre-trained ROAD network to a lower LoD and additionally we randomly perturb SDF annotations at the final LoD of interest with a uniform noise distribution scaled by the voxel size. This procedure corresponds to adding different levels of metric noise at LoD 7 (small), 6 (medium), 5 (large), and 4 (severe). We then use a network trained on unperturbed data (from the same set of 512 Google Scanned Objects as in the experiment above) to fit to the occupancy and noisy surface annotations at a particular LoD; for all LoDs below the query LoD we supervise only on occupancy. Finally, we visualize the fully extracted object (i.e. at LoD 8). Once again, we observe that our method is robust to introduced perturbations and is able to faithfully reconstruct objects even when noise is introduced (see Fig.~\ref{fig:noise}).

\section{Ground Truth Labels}
In our experiments, we extract ground truth labels from meshes and pointclouds, and generally require dense surface points to obtain accurate labels. In practice, the occupancy label of a voxel at a particular LoD is determined by querying whether a point exists within the voxel of interest, and the SDF value and normal are extracted from the nearest neighbor to the voxel center. We observe that these same quantities could also be extracted from an object represented by an SDF. Additionally, we pre-compute and store these annotations once per dataset over all LoDs.

\section{Additional Qualitative Results}
\label{qualitative}
Below, we plot additional qualitative reconstruction results for Google Scanned Objects~\cite{downs2022google} and Thingi32~\cite{zhou2016thingi10k} datasets. Please refer to the supplementary video for further qualitative results, latent space visualizations, and our method's architecture review.
\input{figures/thingi32}
\input{figures/google}


%% file: figures/thingi32.tex
\begin{figure}[t]
	\centering
	\begin{adjustbox}{minipage=\textwidth,scale=1}
	\subfloat{\includegraphics[width=0.16\textwidth]{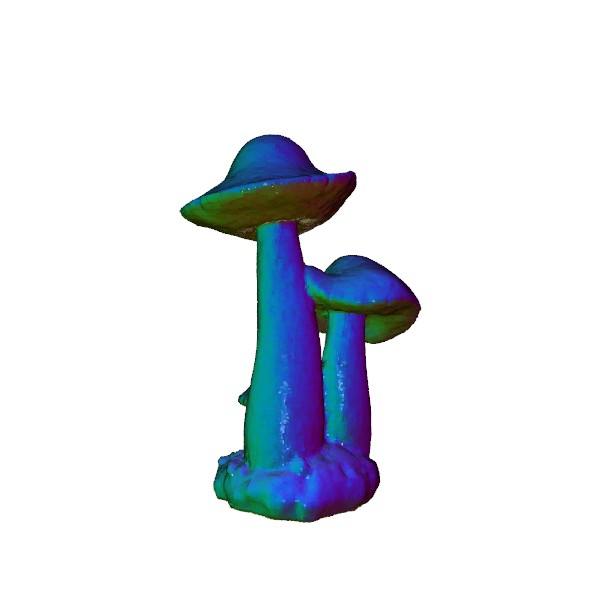}}
	\hfill
	\subfloat{\includegraphics[width=0.16\textwidth]{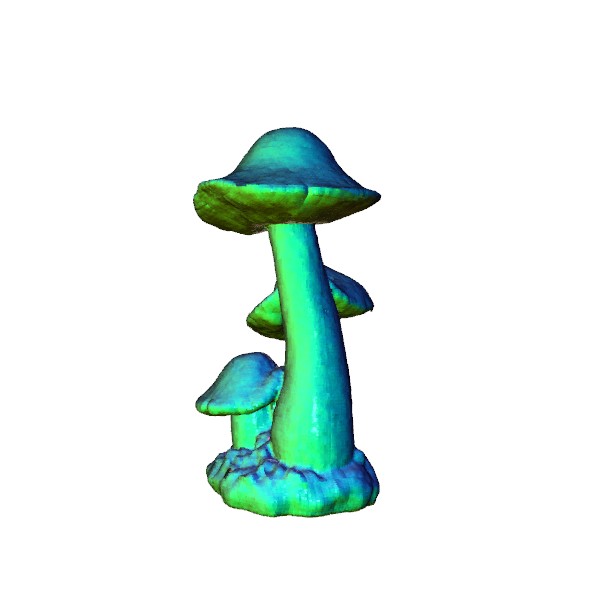}}
	\hfill
	\subfloat{\includegraphics[width=0.16\textwidth]{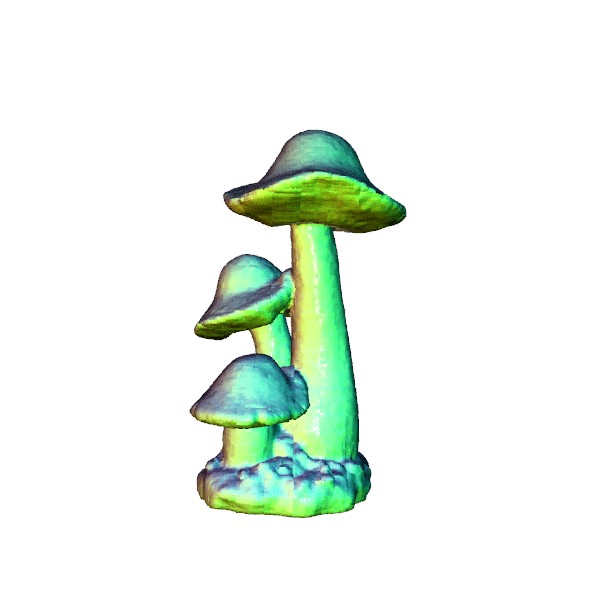}}
	\hfill
	\subfloat{\includegraphics[width=0.16\textwidth]{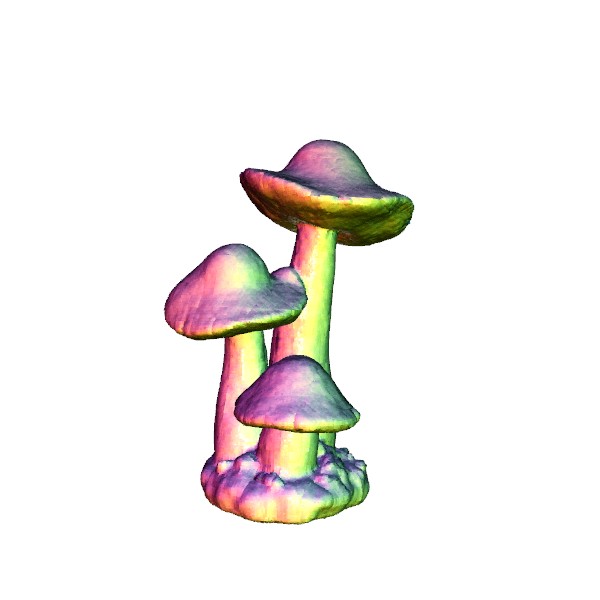}}
	\hfill
	\subfloat{\includegraphics[width=0.16\textwidth]{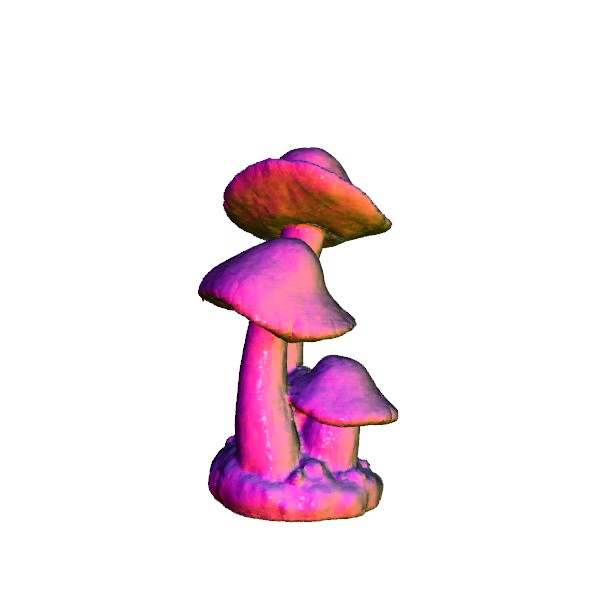}}
	\hfill
	\subfloat{\includegraphics[width=0.16\textwidth]{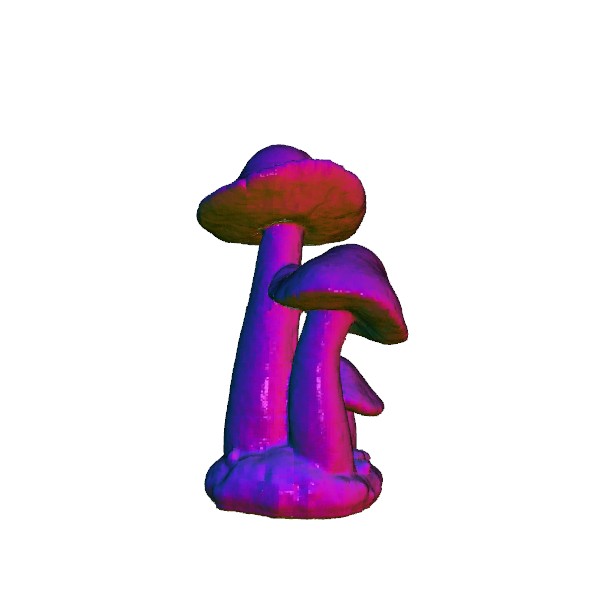}}
	\hfill
	
	\subfloat{\includegraphics[width=0.16\textwidth]{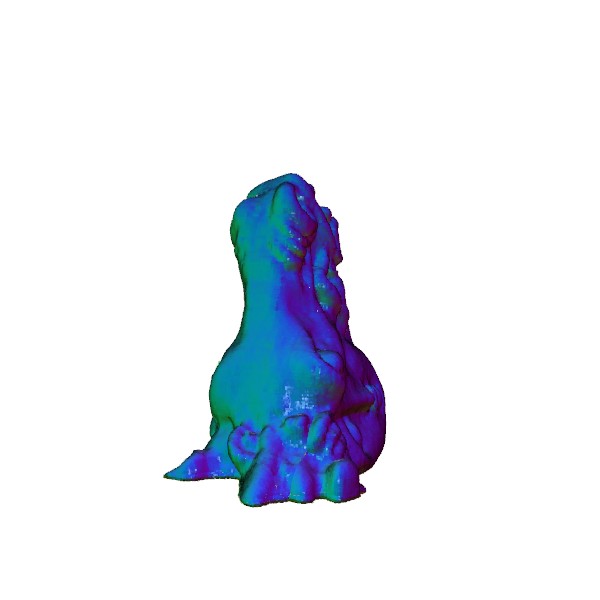}}
	\hfill
	\subfloat{\includegraphics[width=0.16\textwidth]{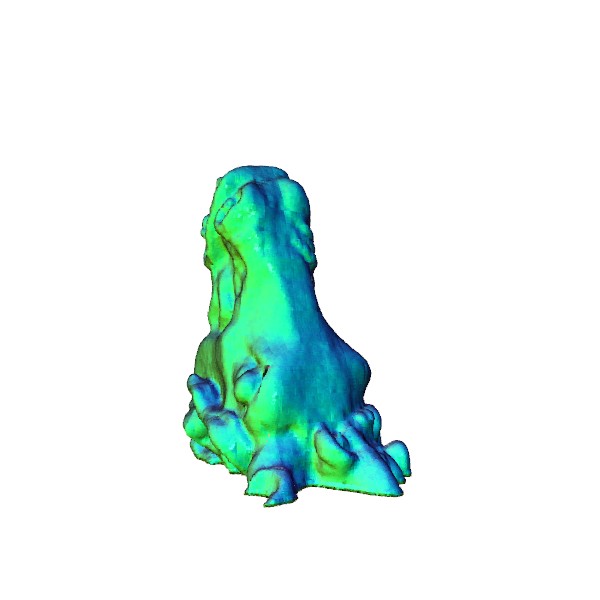}}
	\hfill
	\subfloat{\includegraphics[width=0.16\textwidth]{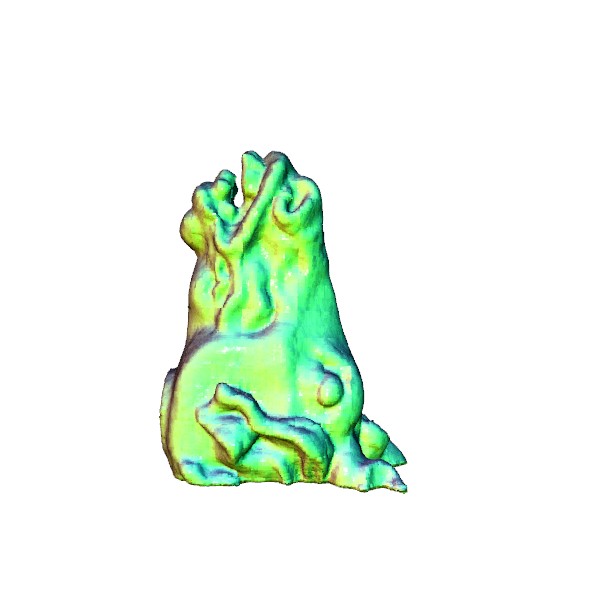}}
	\hfill
	\subfloat{\includegraphics[width=0.16\textwidth]{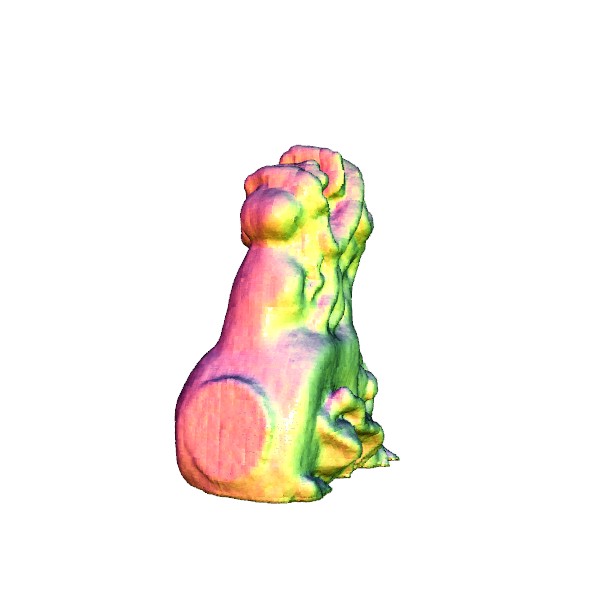}}
	\hfill
	\subfloat{\includegraphics[width=0.16\textwidth]{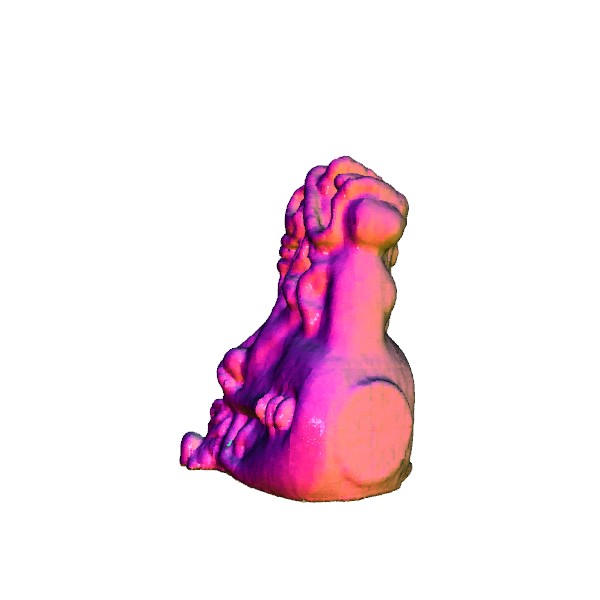}}
	\hfill
	\subfloat{\includegraphics[width=0.16\textwidth]{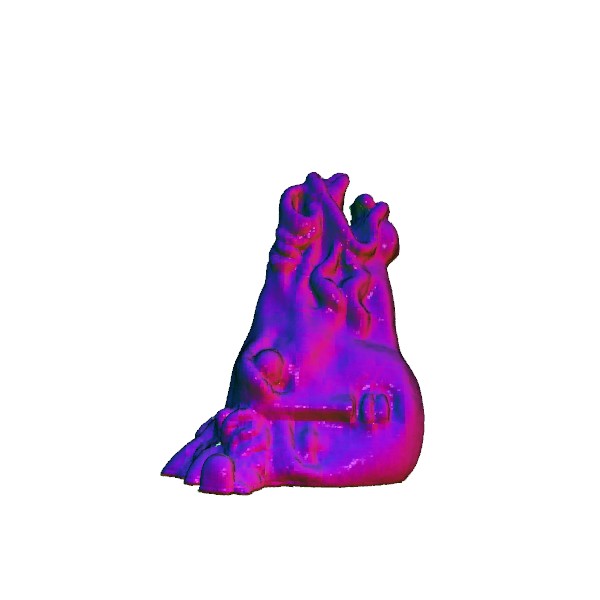}}
	\hfill

	\subfloat{\includegraphics[width=0.16\textwidth]{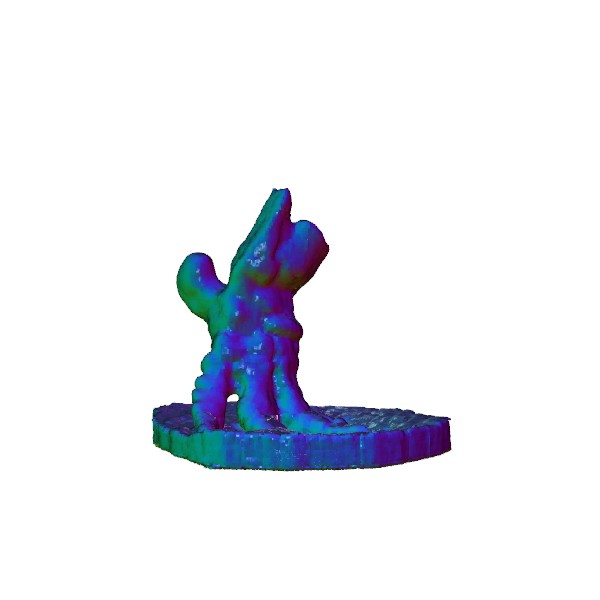}}
	\hfill
	\subfloat{\includegraphics[width=0.16\textwidth]{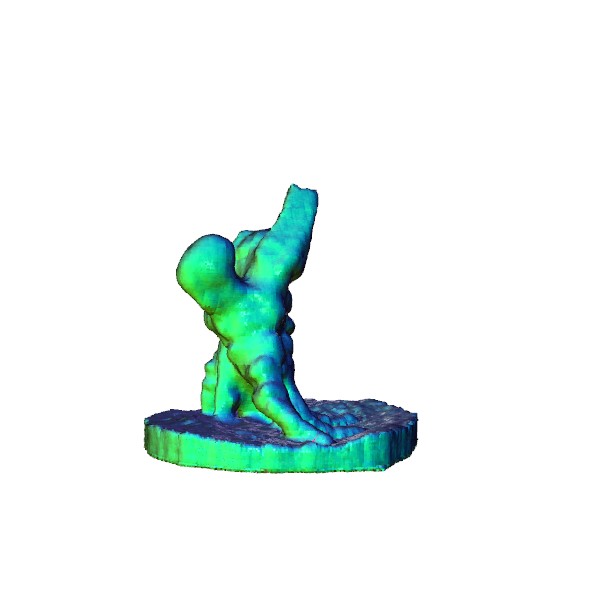}}
	\hfill
	\subfloat{\includegraphics[width=0.16\textwidth]{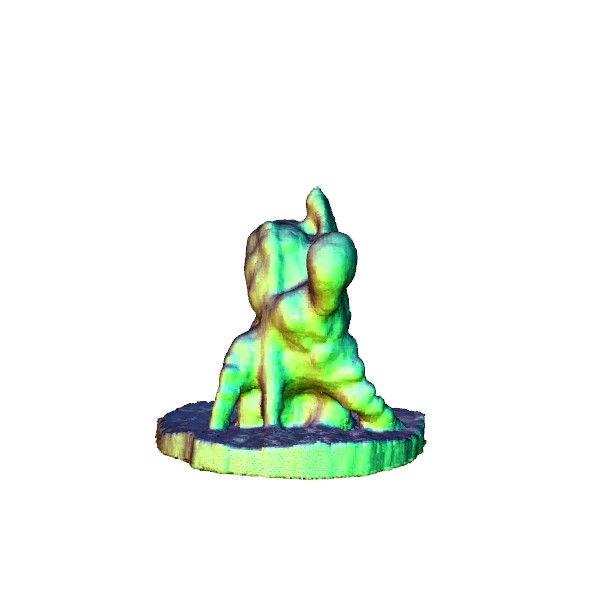}}
	\hfill
	\subfloat{\includegraphics[width=0.16\textwidth]{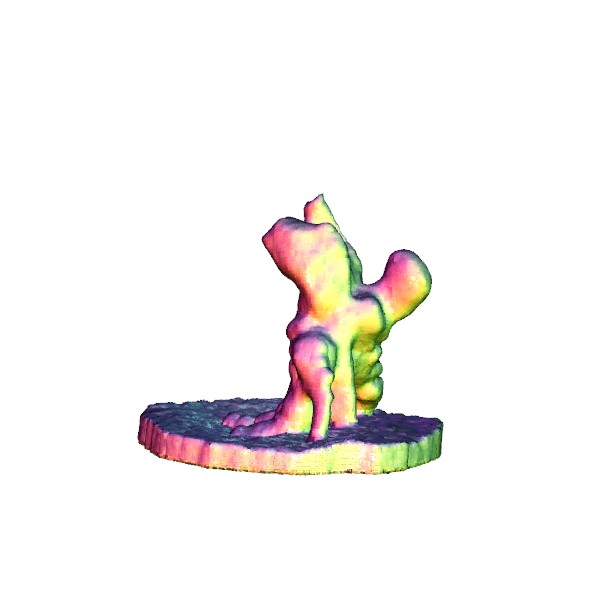}}
	\hfill
	\subfloat{\includegraphics[width=0.16\textwidth]{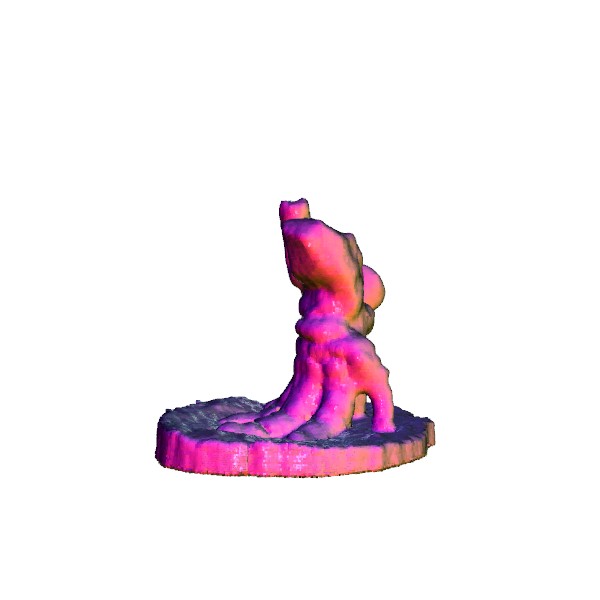}}
	\hfill
	\subfloat{\includegraphics[width=0.16\textwidth]{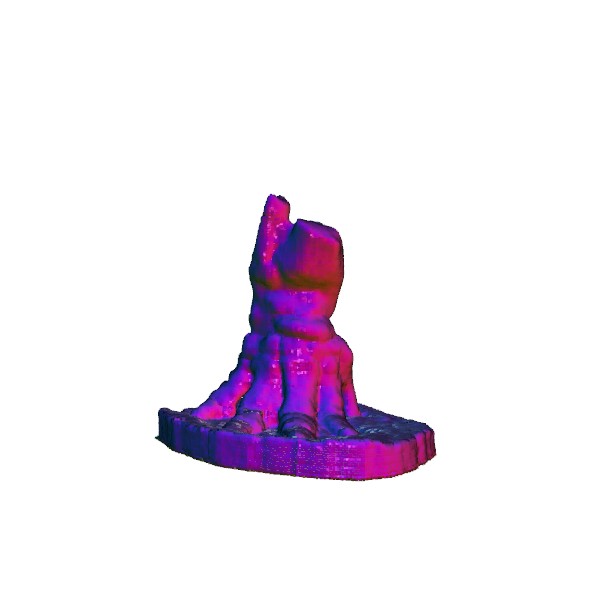}}
	\hfill
	
	\subfloat{\includegraphics[width=0.16\textwidth]{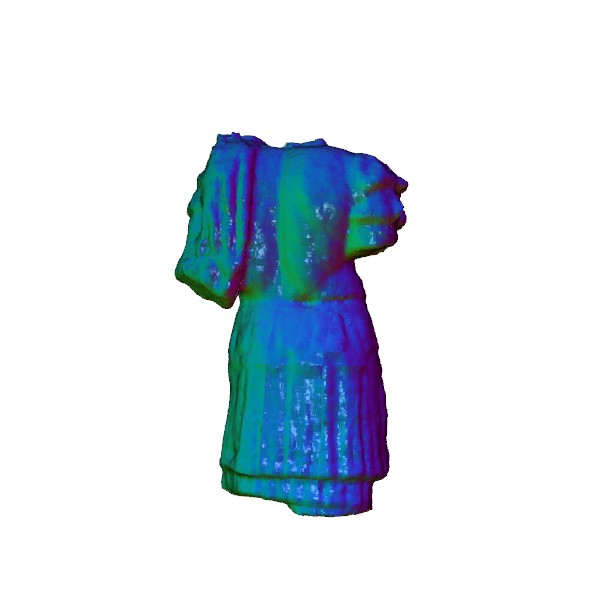}}
	\hfill
	\subfloat{\includegraphics[width=0.16\textwidth]{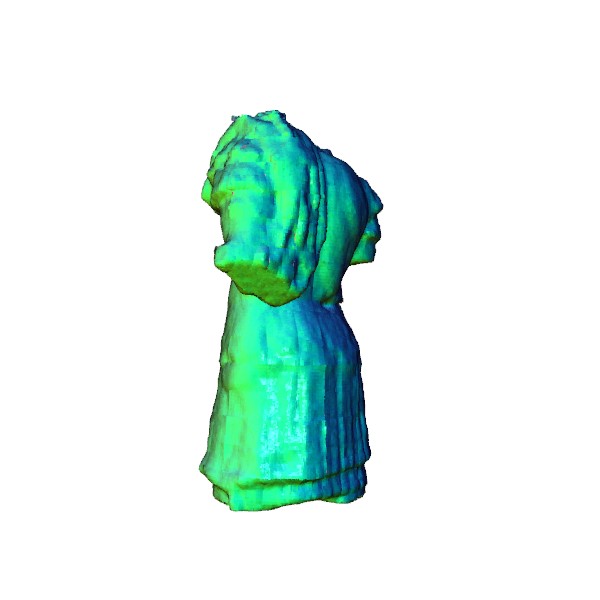}}
	\hfill
	\subfloat{\includegraphics[width=0.16\textwidth]{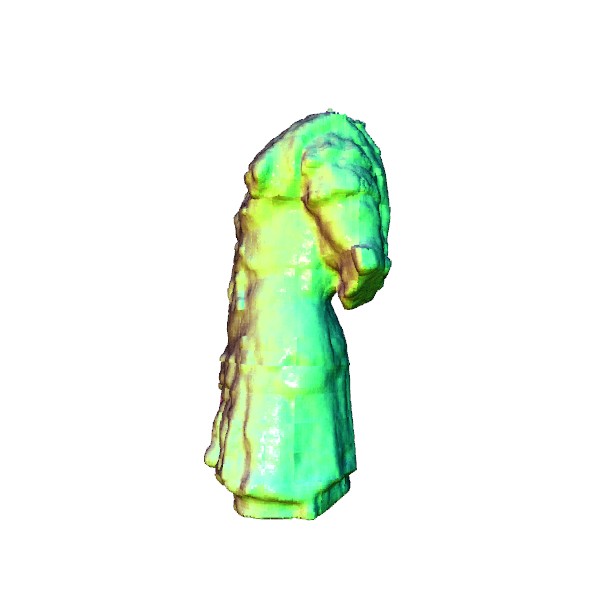}}
	\hfill
	\subfloat{\includegraphics[width=0.16\textwidth]{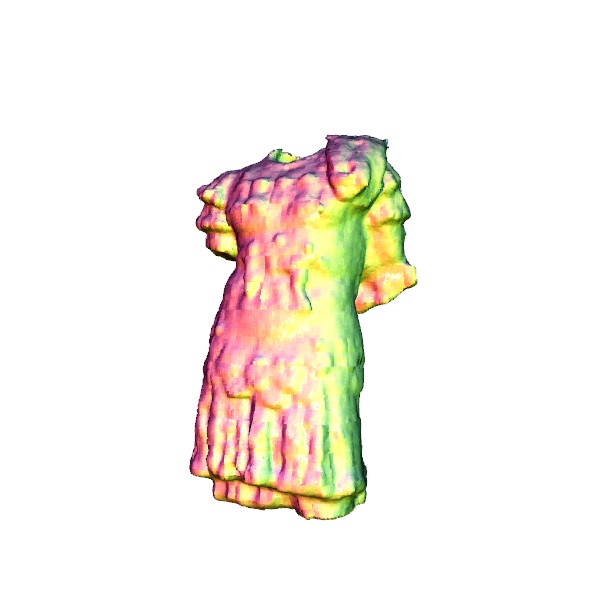}}
	\hfill
	\subfloat{\includegraphics[width=0.16\textwidth]{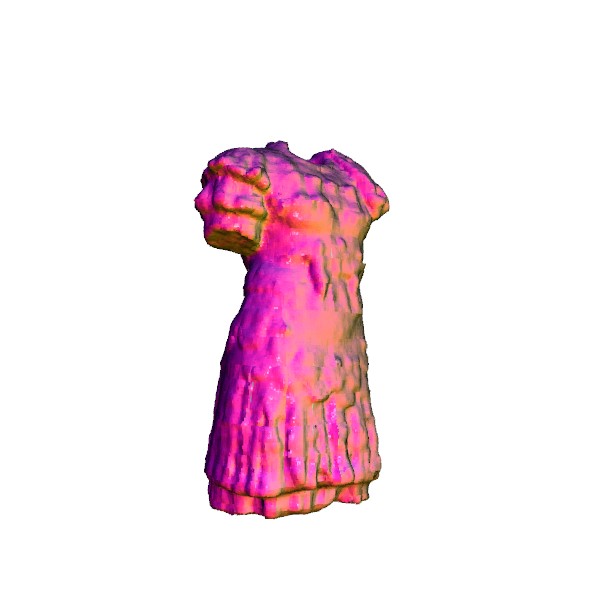}}
	\hfill
	\subfloat{\includegraphics[width=0.16\textwidth]{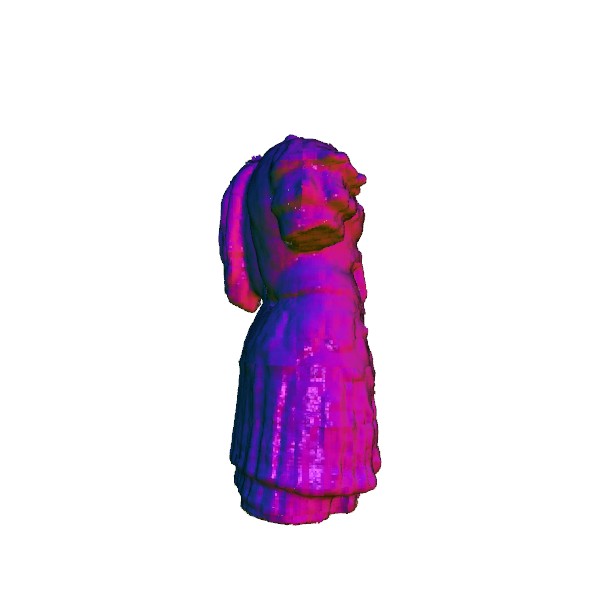}}
	\hfill

	\subfloat{\includegraphics[width=0.16\textwidth]{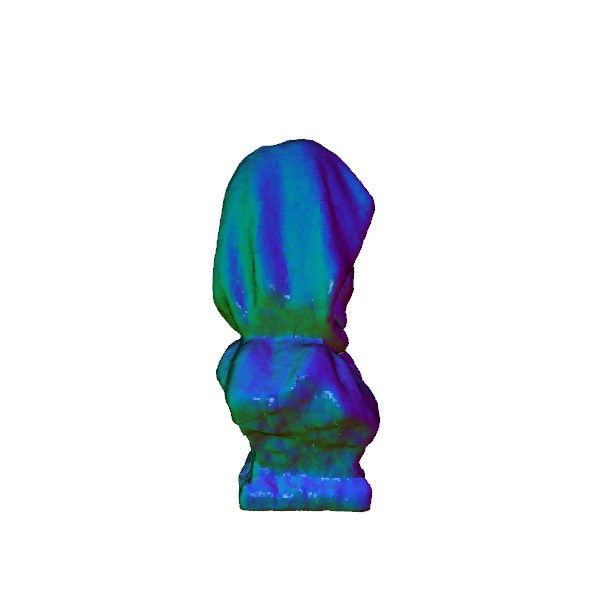}}
	\hfill
	\subfloat{\includegraphics[width=0.16\textwidth]{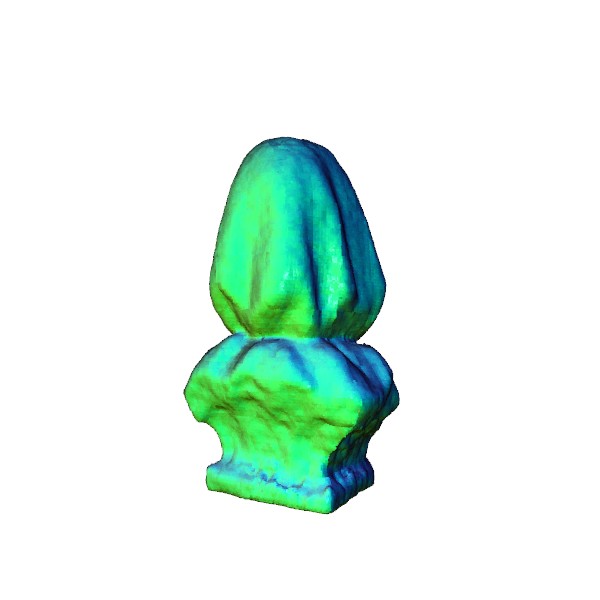}}
	\hfill
	\subfloat{\includegraphics[width=0.16\textwidth]{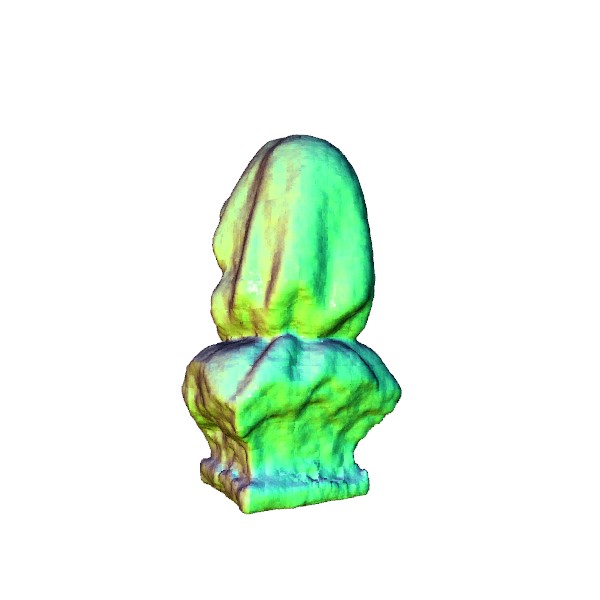}}
	\hfill
	\subfloat{\includegraphics[width=0.16\textwidth]{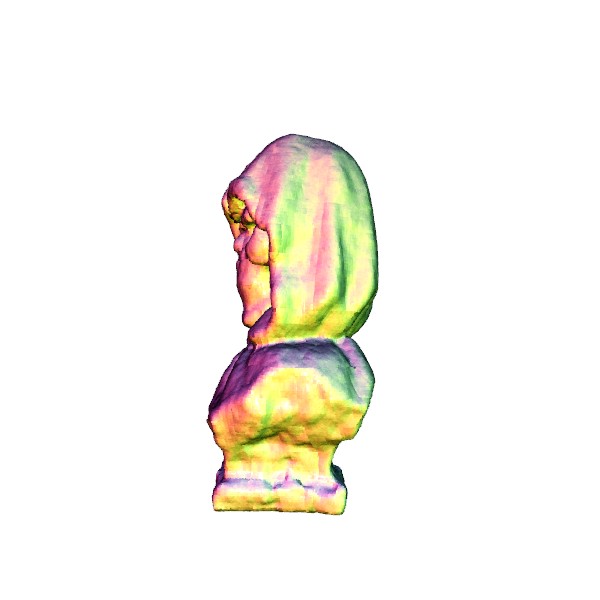}}
	\hfill
	\subfloat{\includegraphics[width=0.16\textwidth]{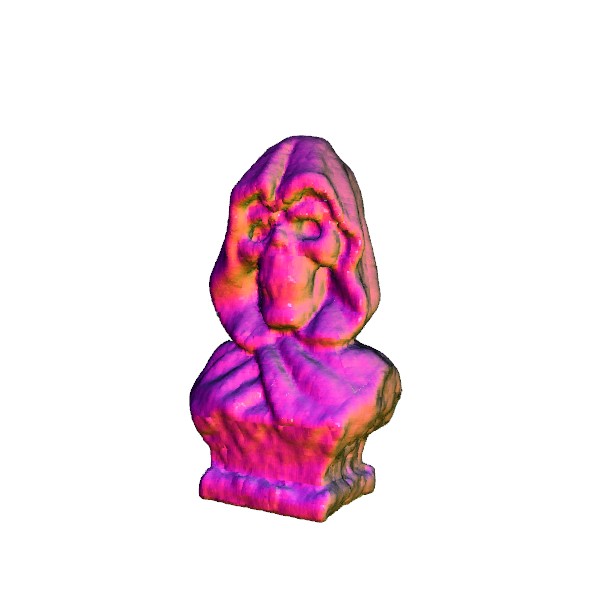}}
	\hfill
	\subfloat{\includegraphics[width=0.16\textwidth]{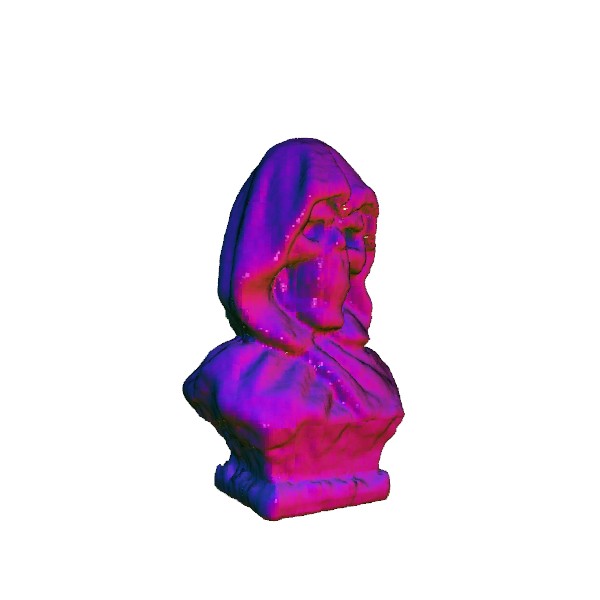}}
	\hfill
	
	\subfloat{\includegraphics[width=0.16\textwidth]{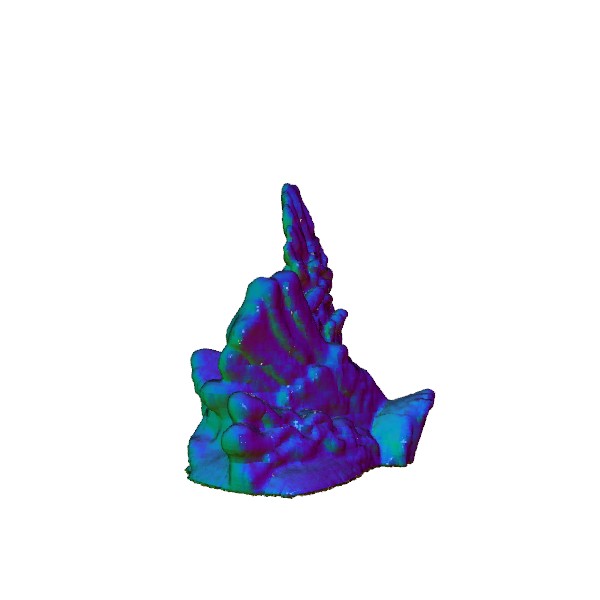}}
	\hfill
	\subfloat{\includegraphics[width=0.16\textwidth]{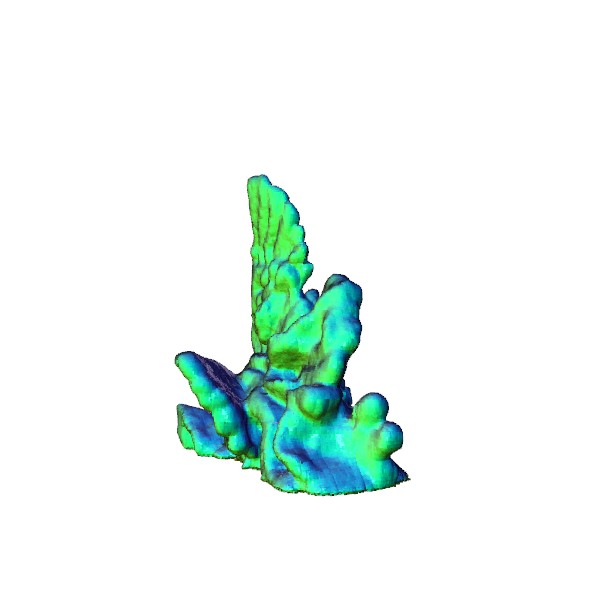}}
	\hfill
	\subfloat{\includegraphics[width=0.16\textwidth]{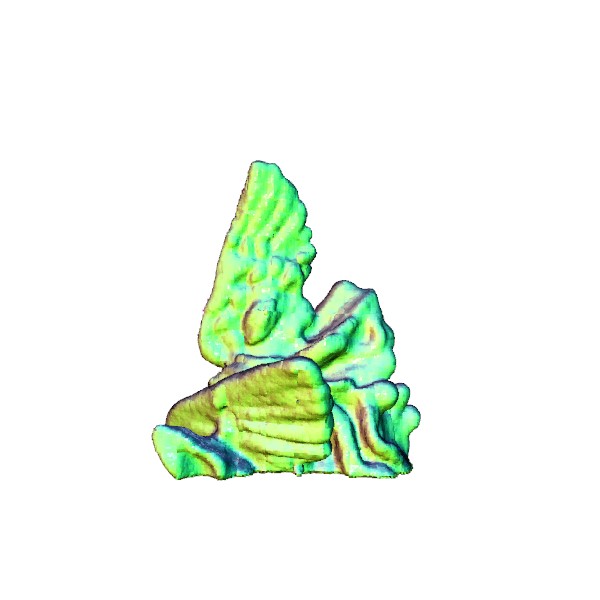}}
	\hfill
	\subfloat{\includegraphics[width=0.16\textwidth]{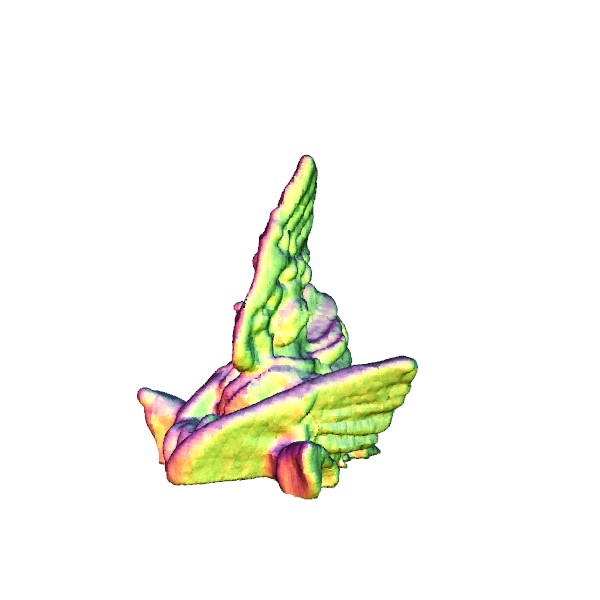}}
	\hfill
	\subfloat{\includegraphics[width=0.16\textwidth]{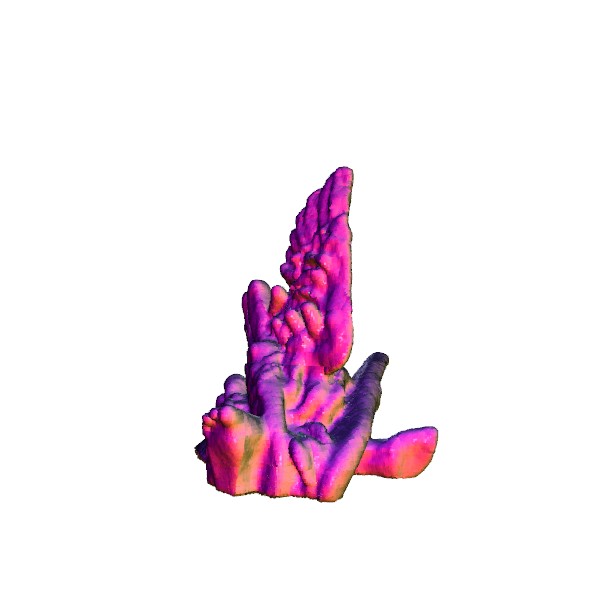}}
	\hfill
	\subfloat{\includegraphics[width=0.16\textwidth]{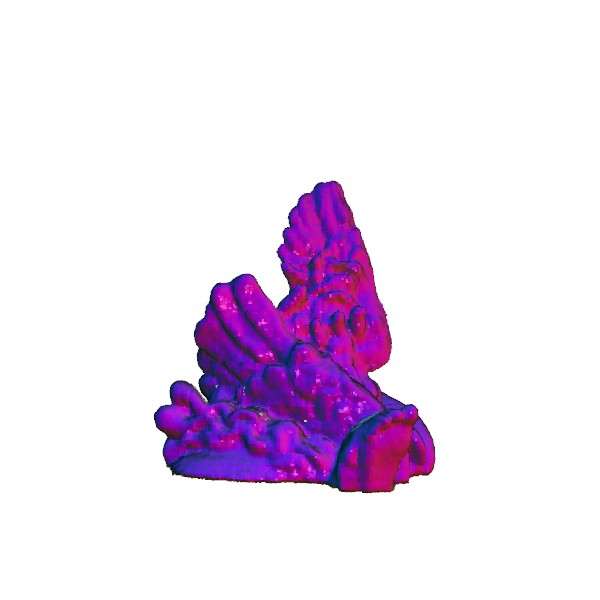}}
	\hfill
	
	\subfloat{\includegraphics[width=0.16\textwidth]{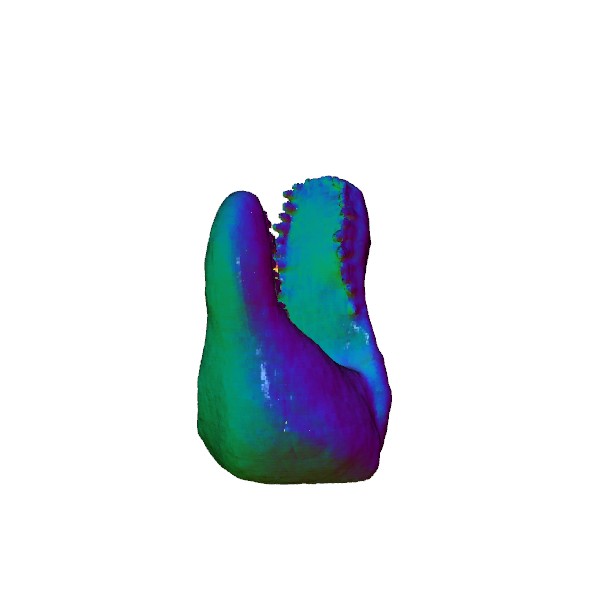}}
	\hfill
	\subfloat{\includegraphics[width=0.16\textwidth]{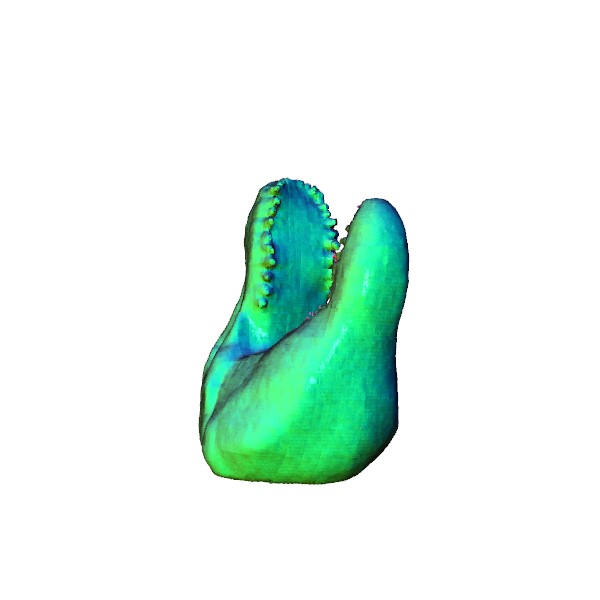}}
	\hfill
	\subfloat{\includegraphics[width=0.16\textwidth]{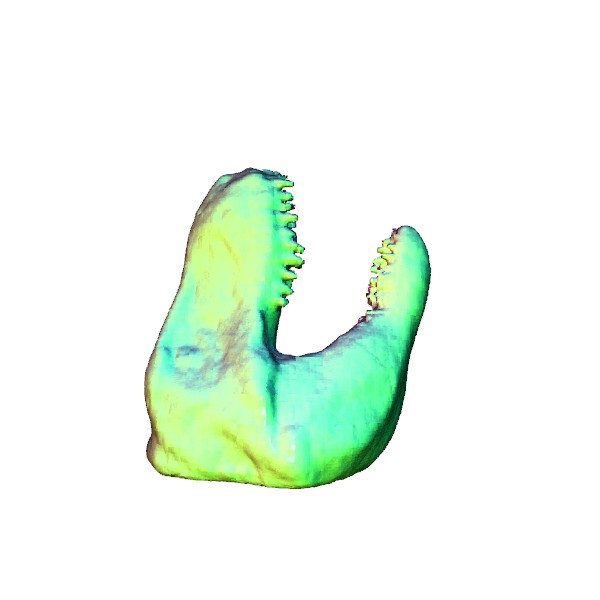}}
	\hfill
	\subfloat{\includegraphics[width=0.16\textwidth]{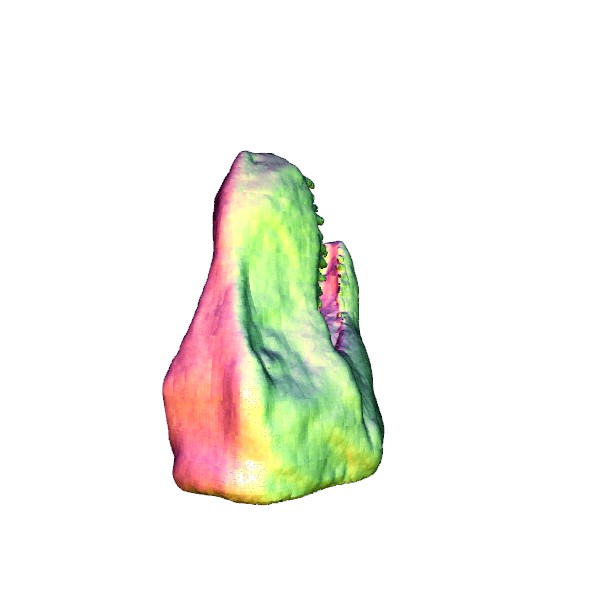}}
	\hfill
	\subfloat{\includegraphics[width=0.16\textwidth]{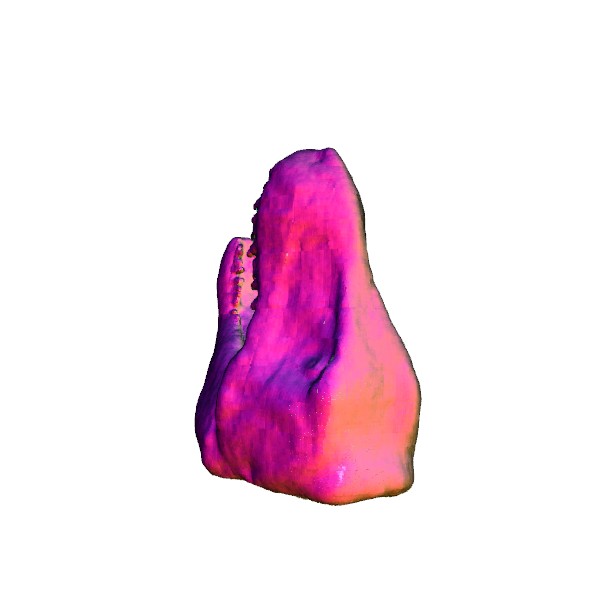}}
	\hfill
	\subfloat{\includegraphics[width=0.16\textwidth]{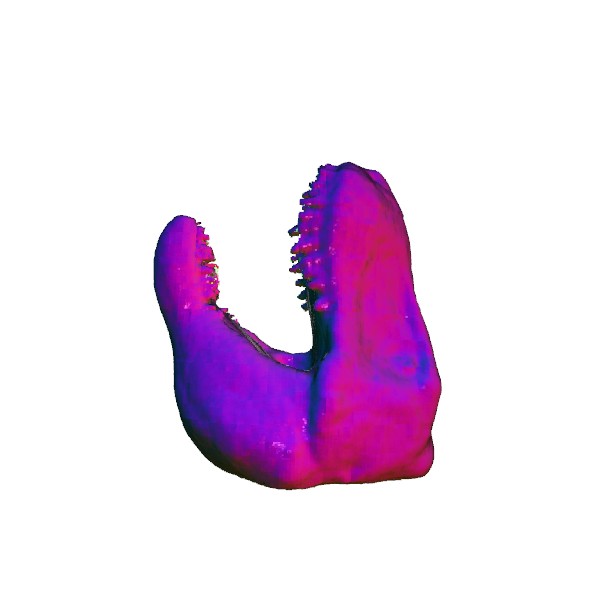}}
	\hfill
	
	\subfloat{\includegraphics[width=0.16\textwidth]{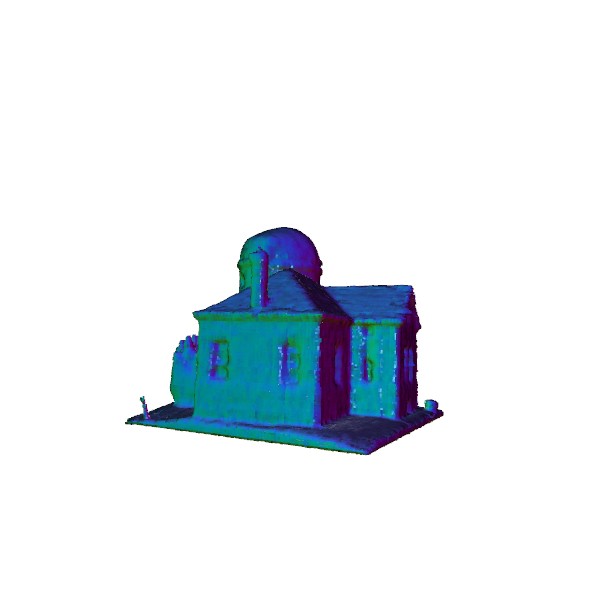}}
	\hfill
	\subfloat{\includegraphics[width=0.16\textwidth]{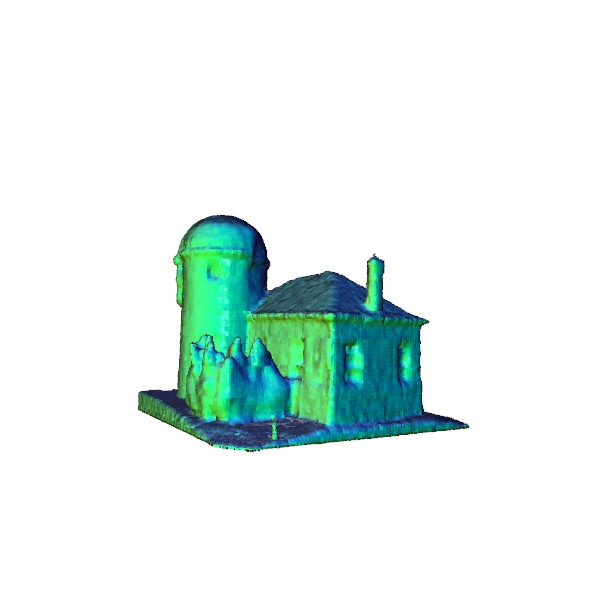}}
	\hfill
	\subfloat{\includegraphics[width=0.16\textwidth]{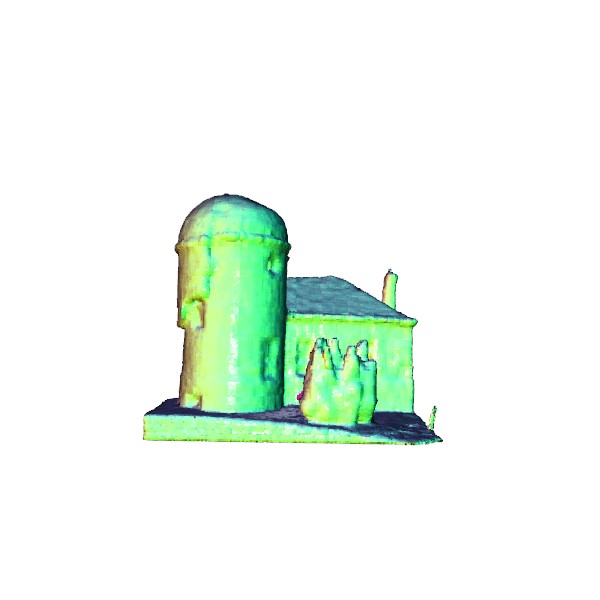}}
	\hfill
	\subfloat{\includegraphics[width=0.16\textwidth]{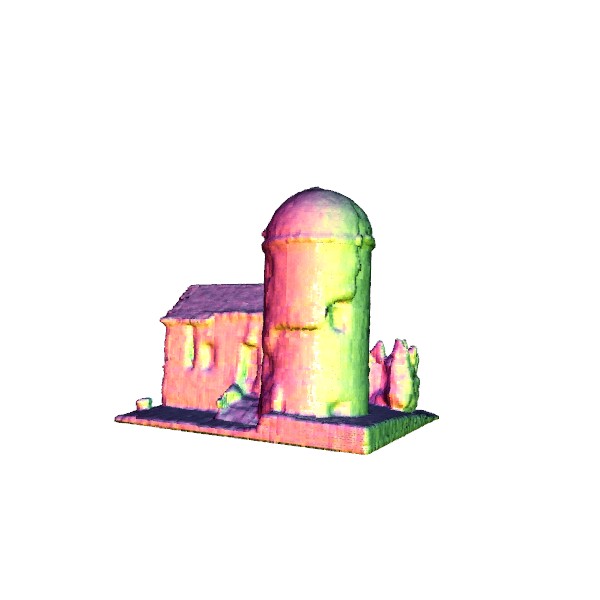}}
	\hfill
	\subfloat{\includegraphics[width=0.16\textwidth]{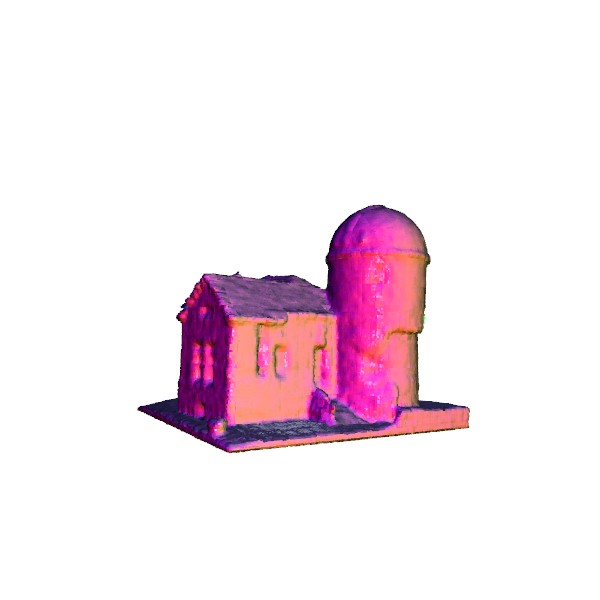}}
	\hfill
	\subfloat{\includegraphics[width=0.16\textwidth]{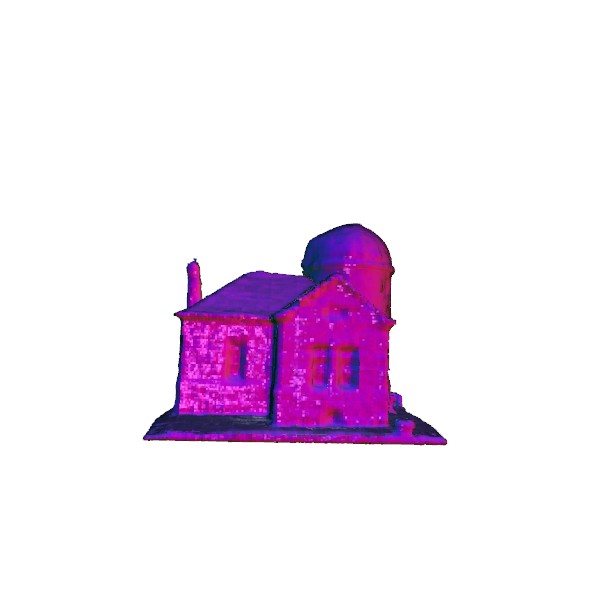}}
	\hfill
	
	\subfloat{\includegraphics[width=0.16\textwidth]{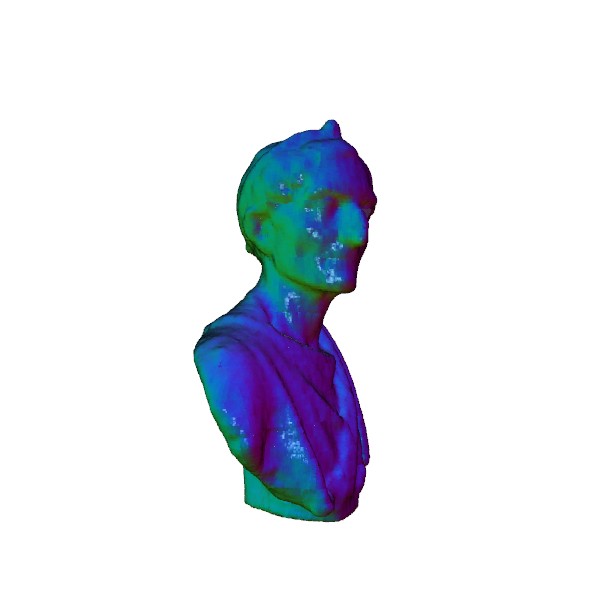}}
	\hfill
	\subfloat{\includegraphics[width=0.16\textwidth]{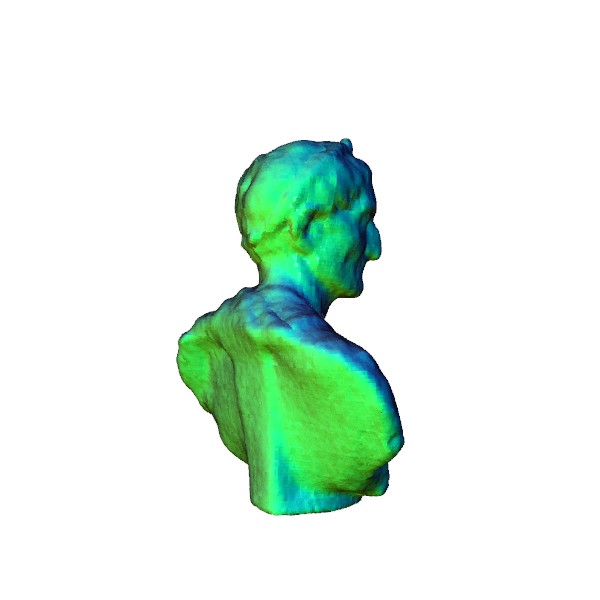}}
	\hfill
	\subfloat{\includegraphics[width=0.16\textwidth]{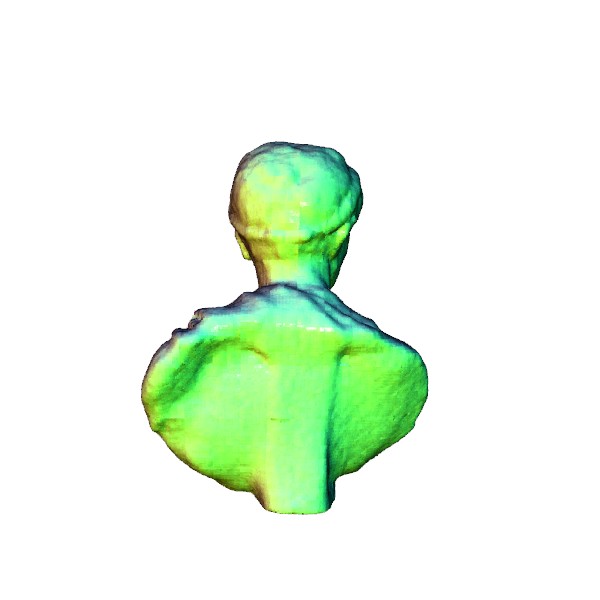}}
	\hfill
	\subfloat{\includegraphics[width=0.16\textwidth]{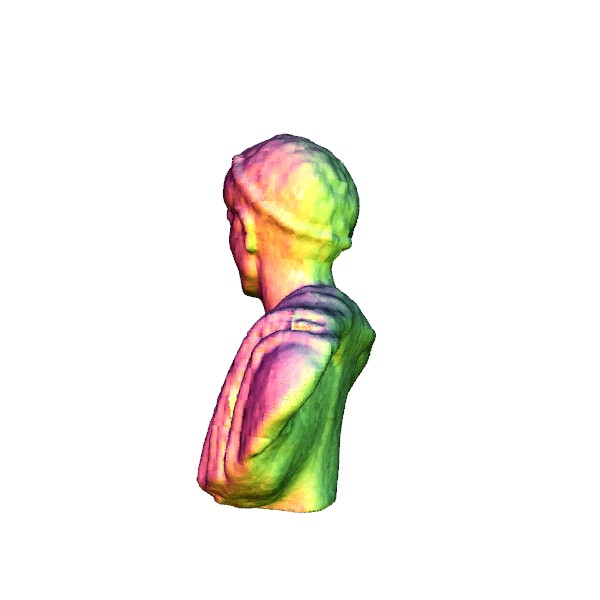}}
	\hfill
	\subfloat{\includegraphics[width=0.16\textwidth]{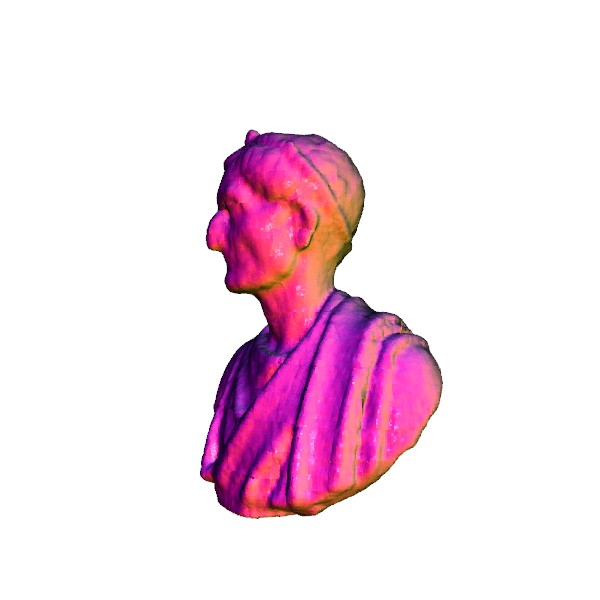}}
	\hfill
	\subfloat{\includegraphics[width=0.16\textwidth]{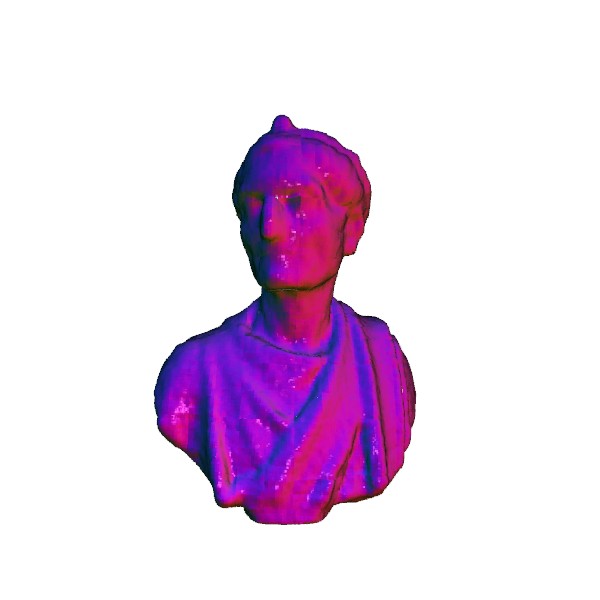}}
	\hfill
	
	\caption{Example reconstruction results on the Thingi32 dataset.}
	\end{adjustbox}
\end{figure}

%% file: figures/google.tex
\begin{figure}[t]
	\centering
	\begin{adjustbox}{minipage=\textwidth,scale=1}
	\subfloat{\includegraphics[width=0.16\textwidth]{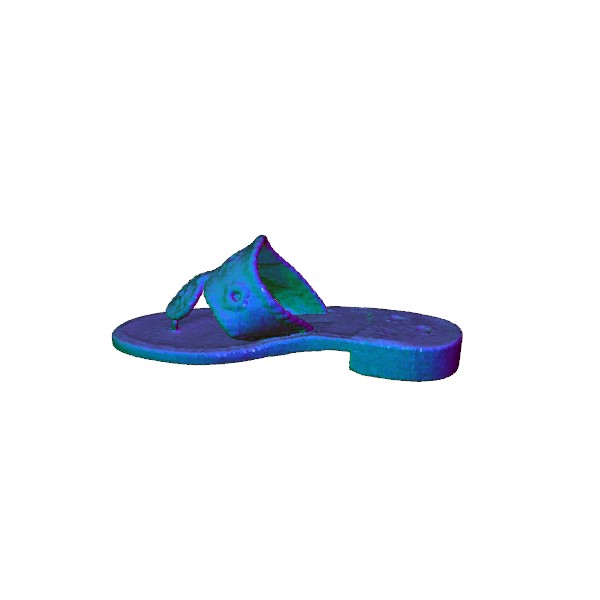}}
	\hfill
	\subfloat{\includegraphics[width=0.16\textwidth]{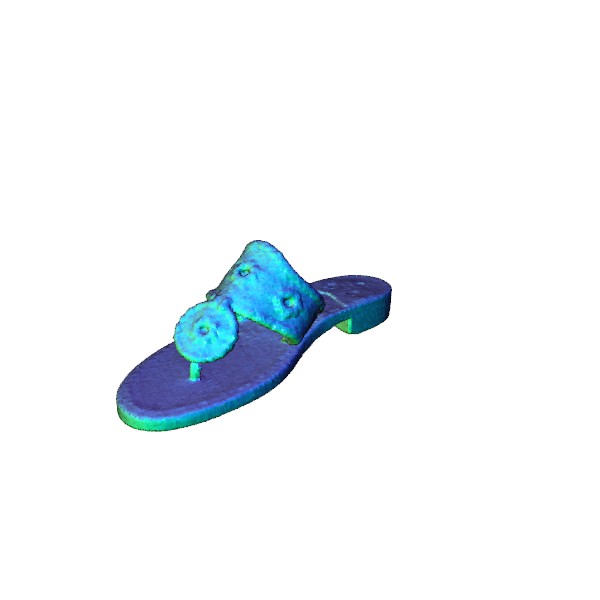}}
	\hfill
	\subfloat{\includegraphics[width=0.16\textwidth]{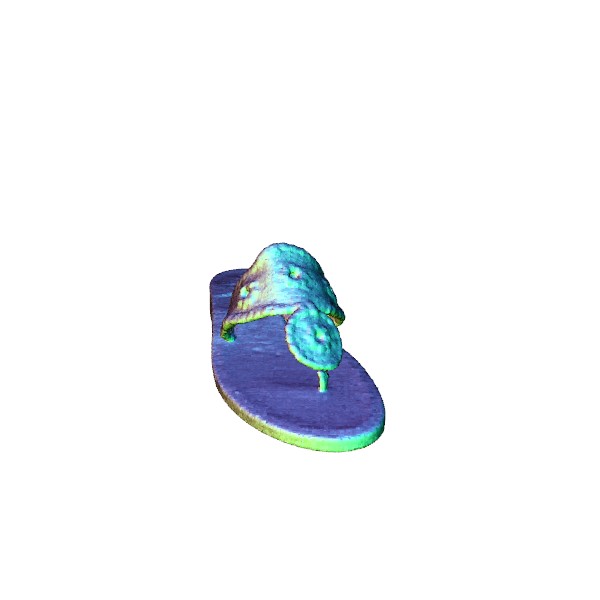}}
	\hfill
	\subfloat{\includegraphics[width=0.16\textwidth]{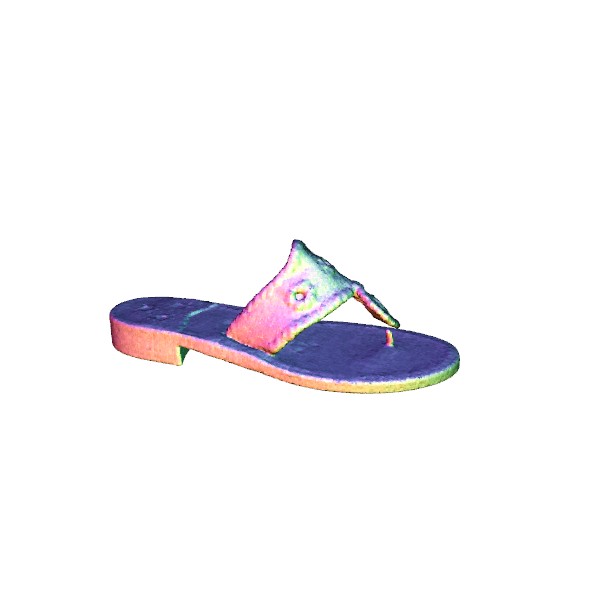}}
	\hfill
	\subfloat{\includegraphics[width=0.16\textwidth]{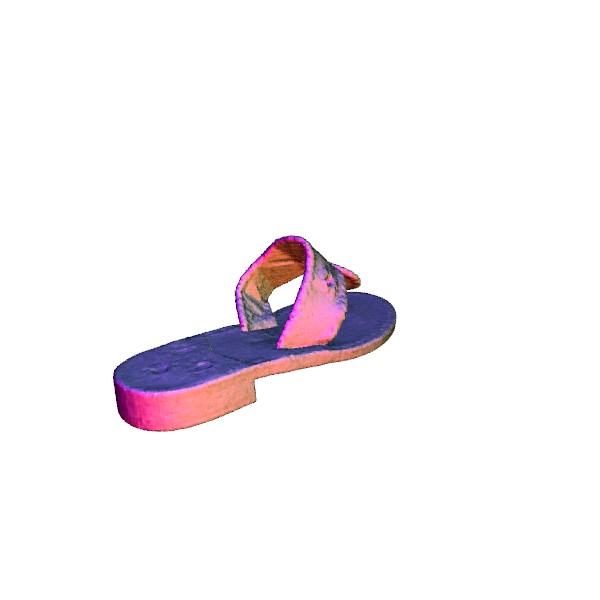}}
	\hfill
	\subfloat{\includegraphics[width=0.16\textwidth]{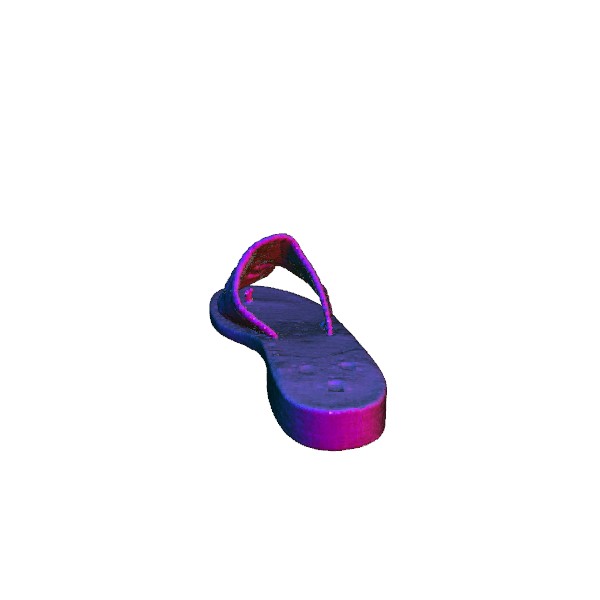}}
	
	\subfloat{\includegraphics[width=0.16\textwidth]{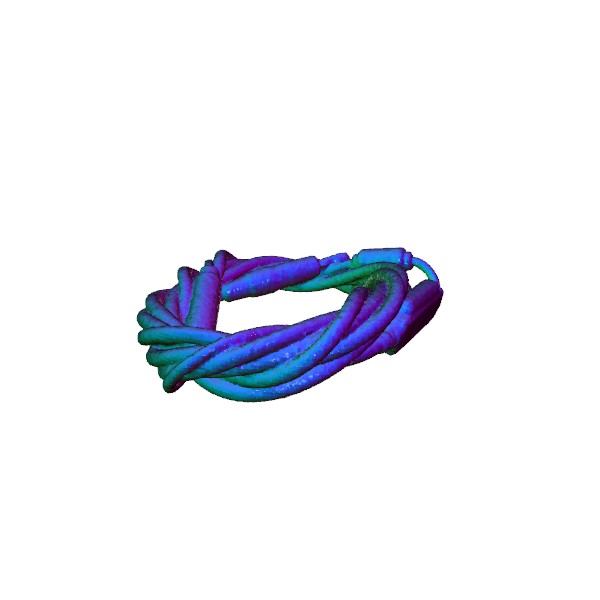}}
	\hfill
	\subfloat{\includegraphics[width=0.16\textwidth]{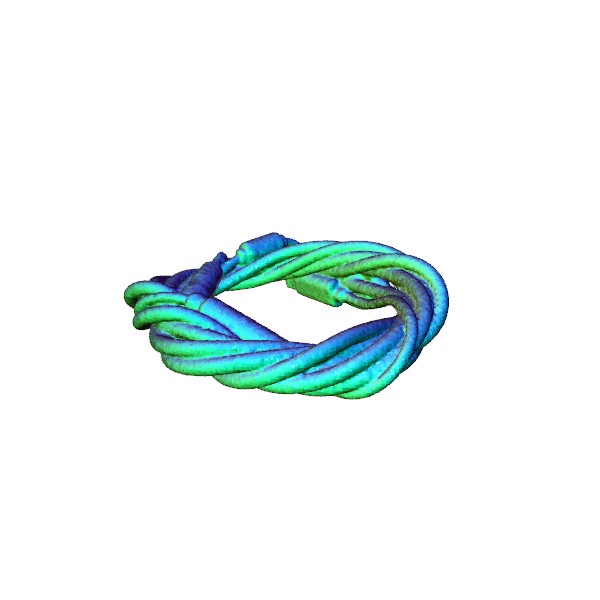}}
	\hfill
	\subfloat{\includegraphics[width=0.16\textwidth]{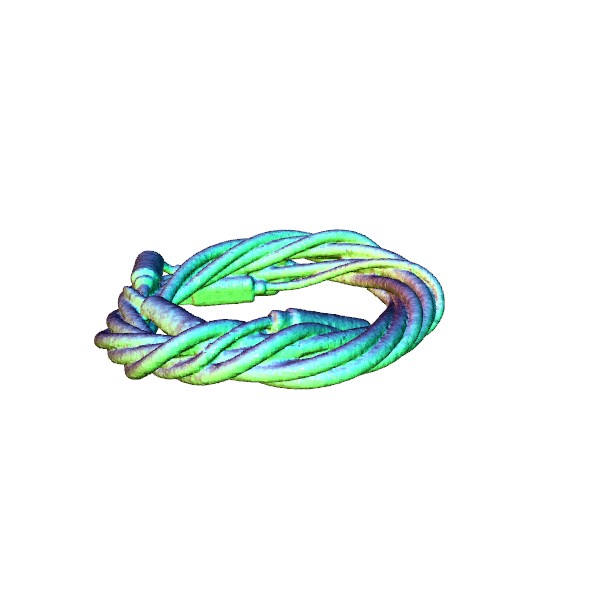}}
	\hfill
	\subfloat{\includegraphics[width=0.16\textwidth]{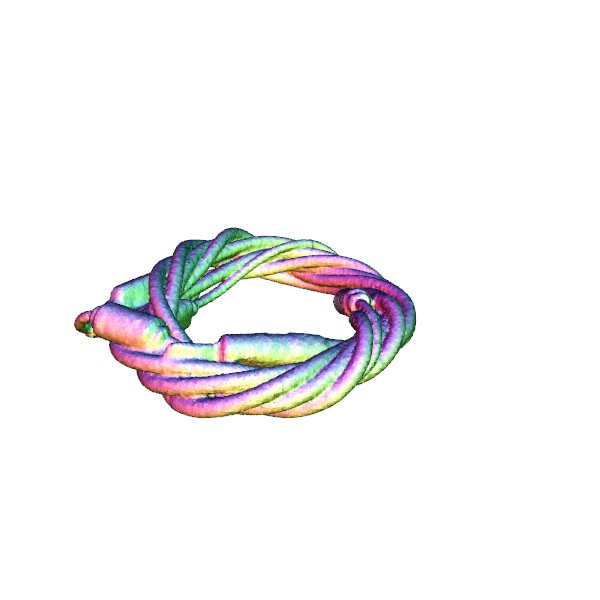}}
	\hfill
	\subfloat{\includegraphics[width=0.16\textwidth]{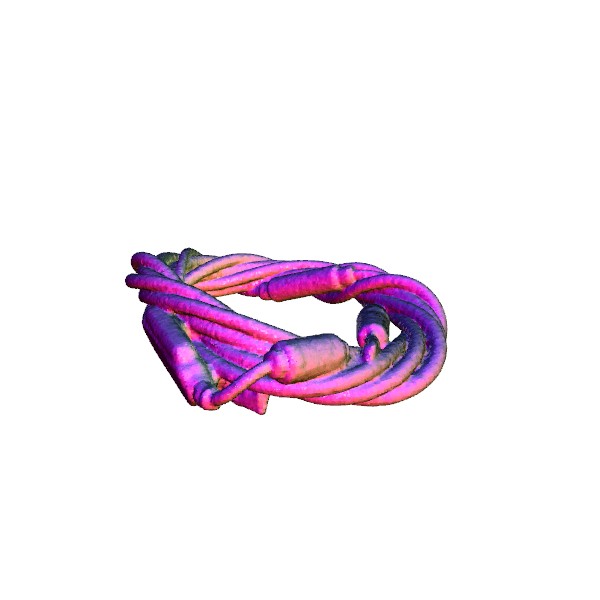}}
	\hfill
	\subfloat{\includegraphics[width=0.16\textwidth]{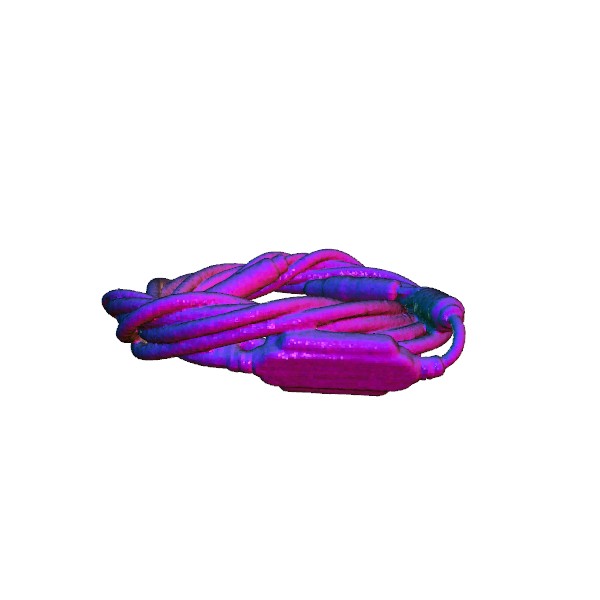}}
	\hfill

	\subfloat{\includegraphics[width=0.16\textwidth]{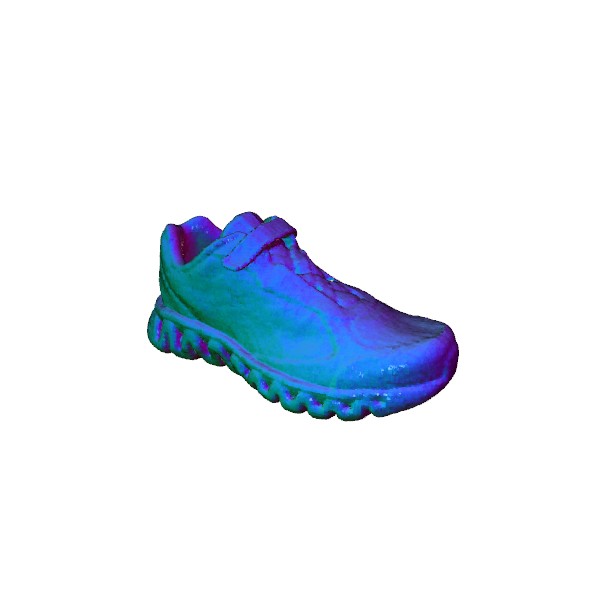}}
	\hfill
	\subfloat{\includegraphics[width=0.16\textwidth]{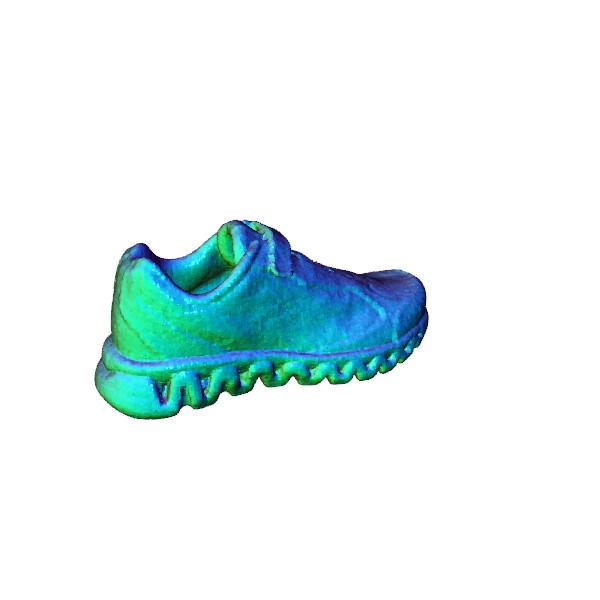}}
	\hfill
	\subfloat{\includegraphics[width=0.16\textwidth]{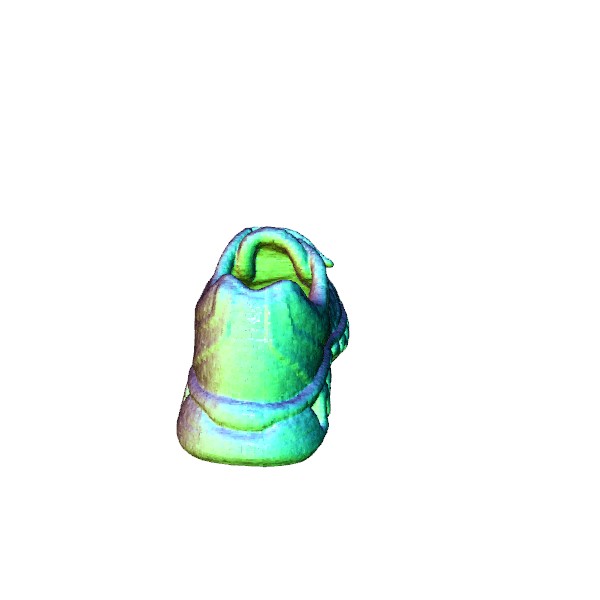}}
	\hfill
	\subfloat{\includegraphics[width=0.16\textwidth]{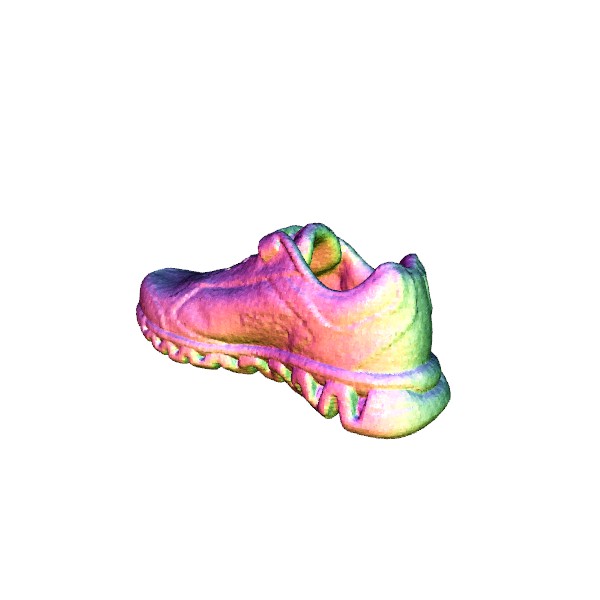}}
	\hfill
	\subfloat{\includegraphics[width=0.16\textwidth]{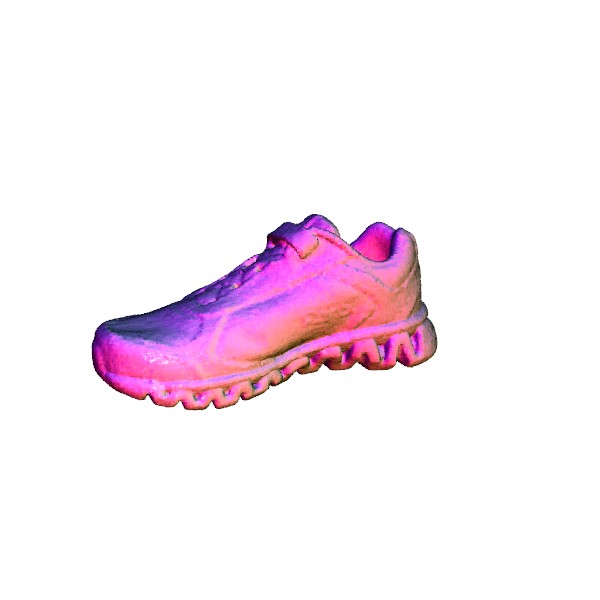}}
	\hfill
	\subfloat{\includegraphics[width=0.16\textwidth]{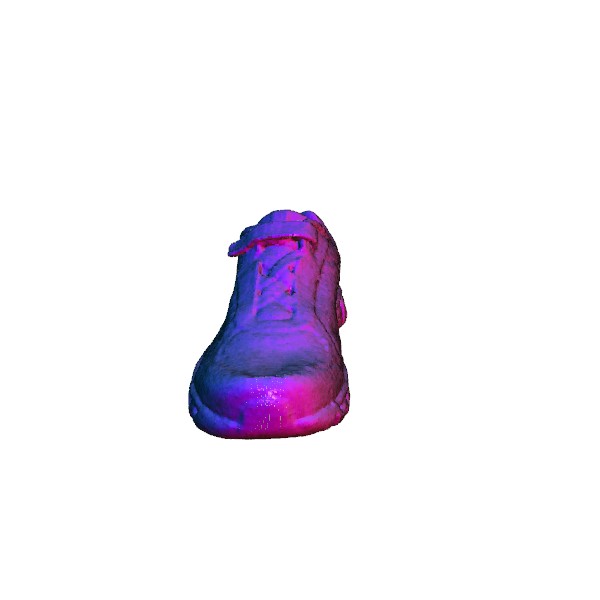}}
	\hfill
	
	\subfloat{\includegraphics[width=0.16\textwidth]{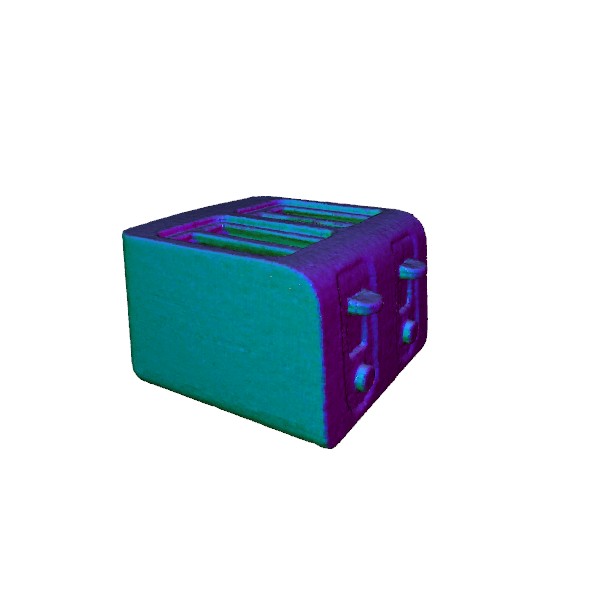}}
	\hfill
	\subfloat{\includegraphics[width=0.16\textwidth]{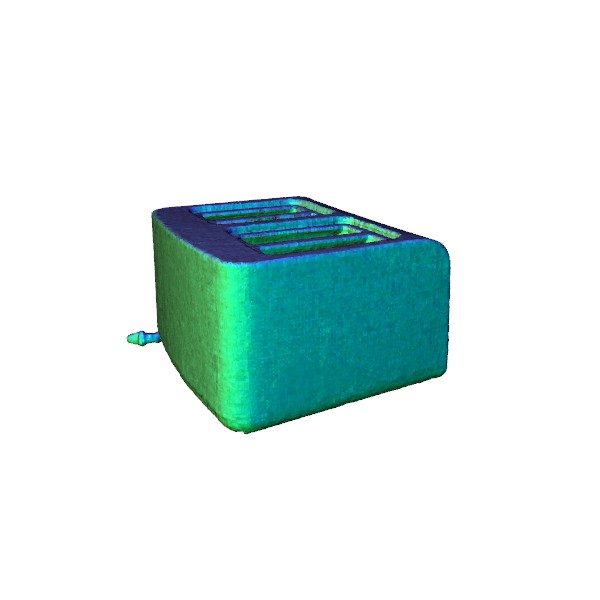}}
	\hfill
	\subfloat{\includegraphics[width=0.16\textwidth]{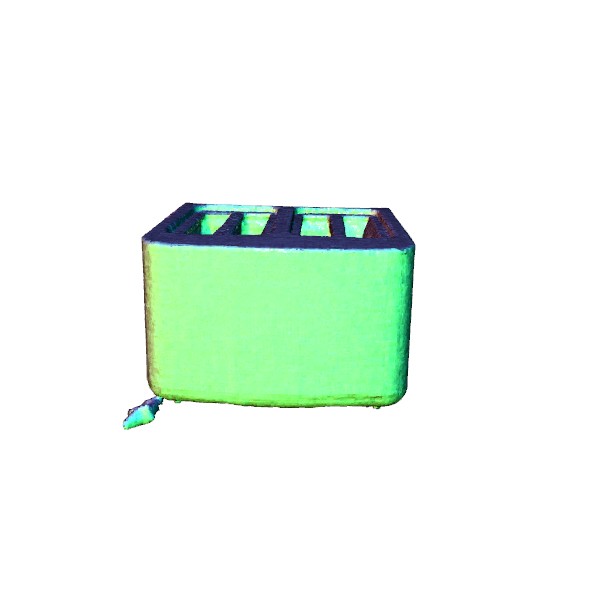}}
	\hfill
	\subfloat{\includegraphics[width=0.16\textwidth]{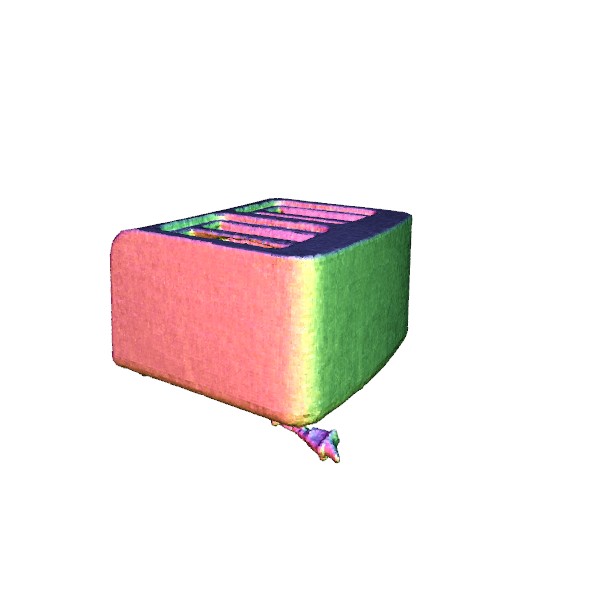}}
	\hfill
	\subfloat{\includegraphics[width=0.16\textwidth]{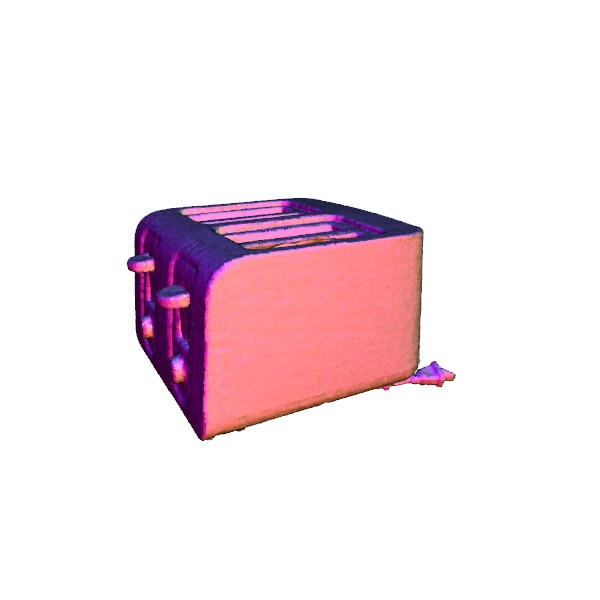}}
	\hfill
	\subfloat{\includegraphics[width=0.16\textwidth]{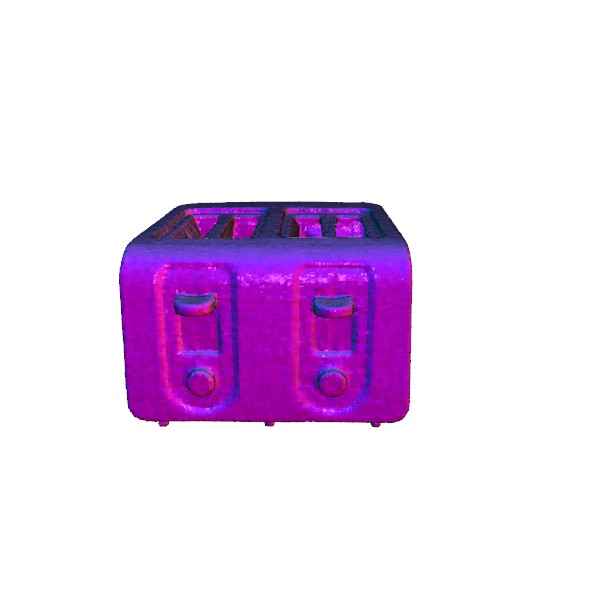}}
	\hfill

	\subfloat{\includegraphics[width=0.16\textwidth]{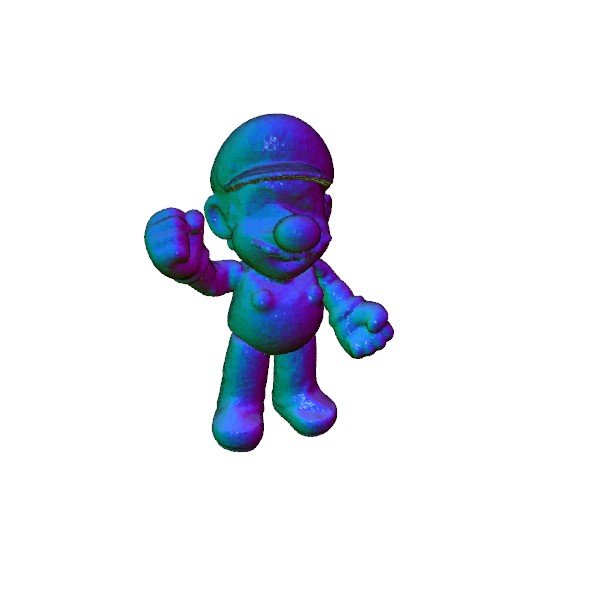}}
	\hfill
	\subfloat{\includegraphics[width=0.16\textwidth]{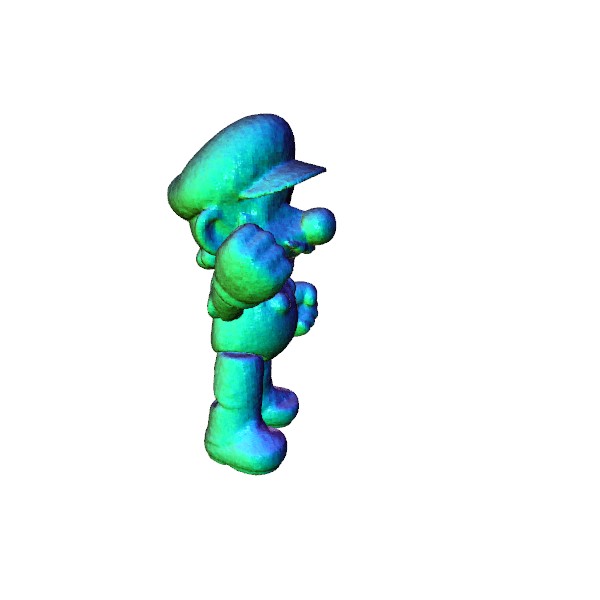}}
	\hfill
	\subfloat{\includegraphics[width=0.16\textwidth]{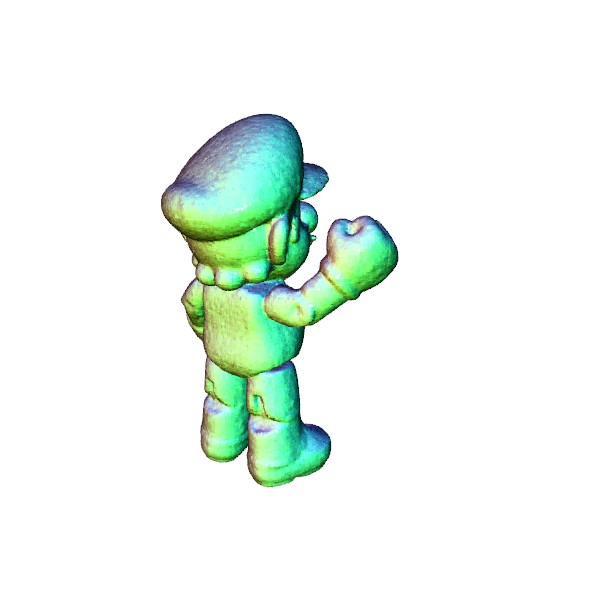}}
	\hfill
	\subfloat{\includegraphics[width=0.16\textwidth]{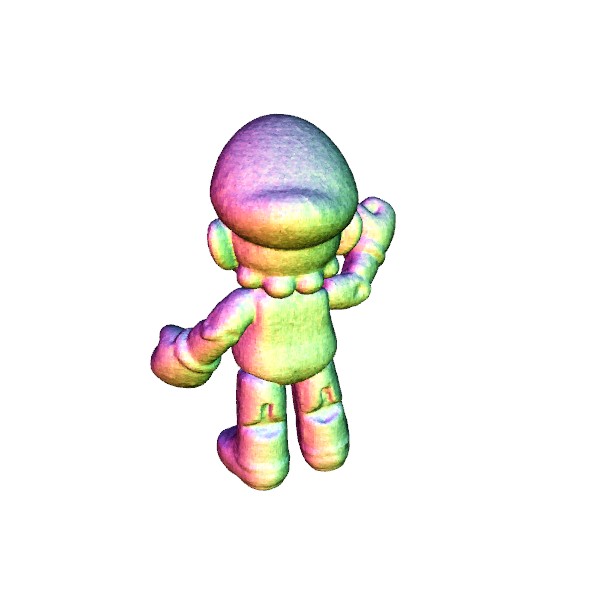}}
	\hfill
	\subfloat{\includegraphics[width=0.16\textwidth]{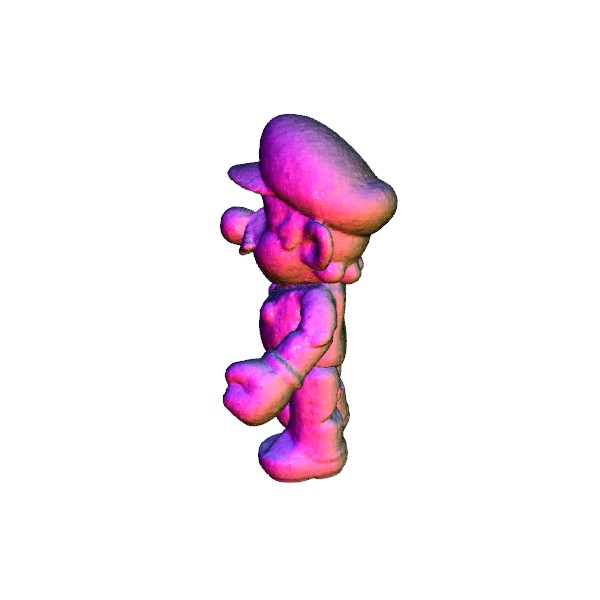}}
	\hfill
	\subfloat{\includegraphics[width=0.16\textwidth]{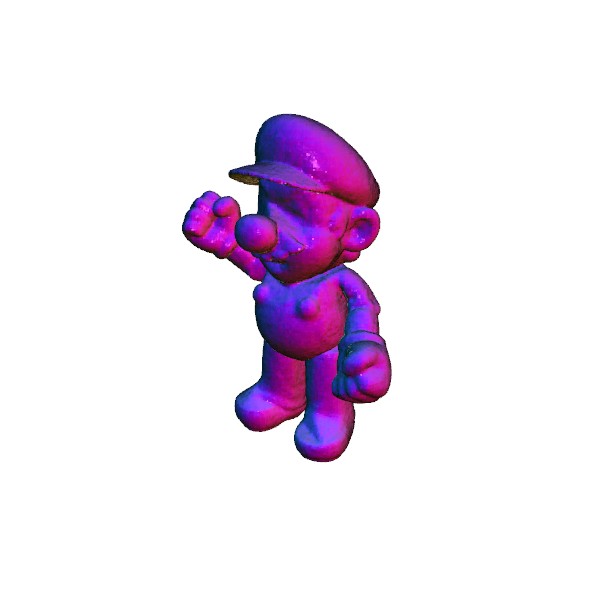}}
	\hfill
	
	\subfloat{\includegraphics[width=0.16\textwidth]{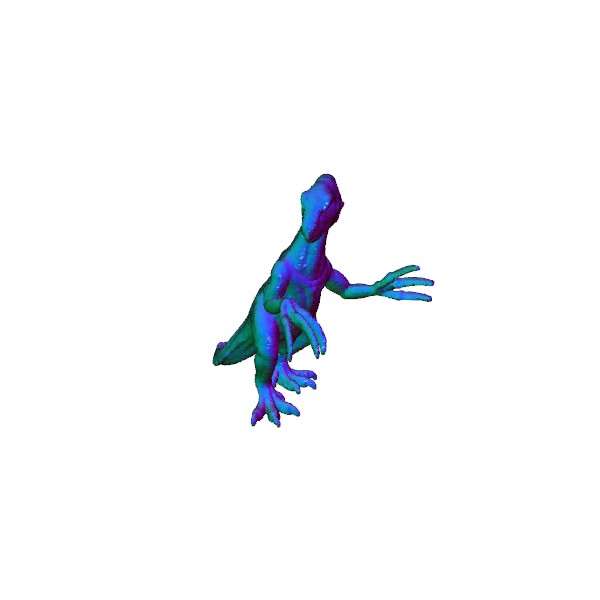}}
	\hfill
	\subfloat{\includegraphics[width=0.16\textwidth]{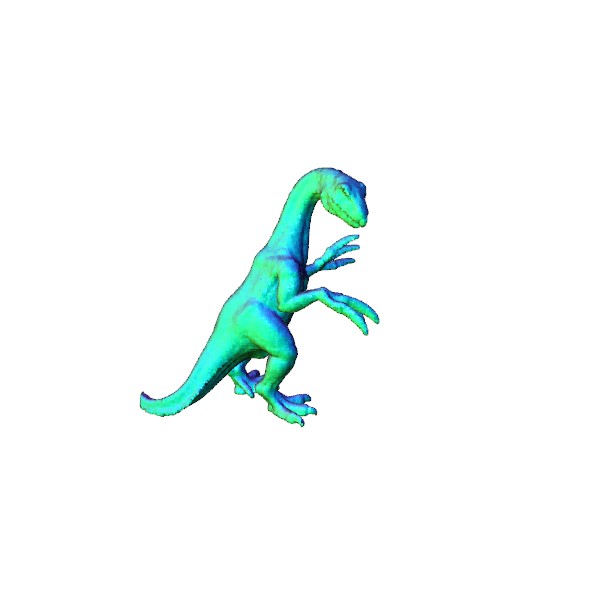}}
	\hfill
	\subfloat{\includegraphics[width=0.16\textwidth]{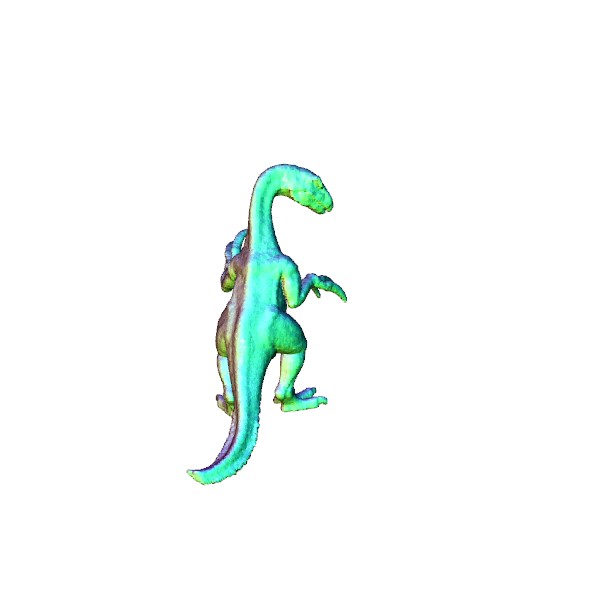}}
	\hfill
	\subfloat{\includegraphics[width=0.16\textwidth]{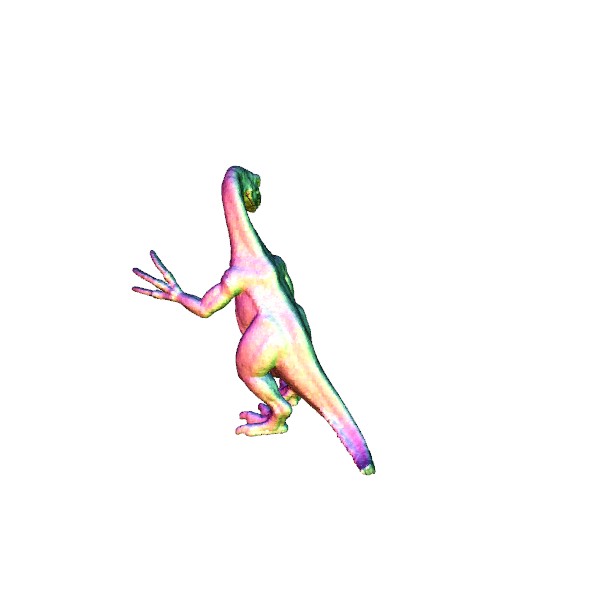}}
	\hfill
	\subfloat{\includegraphics[width=0.16\textwidth]{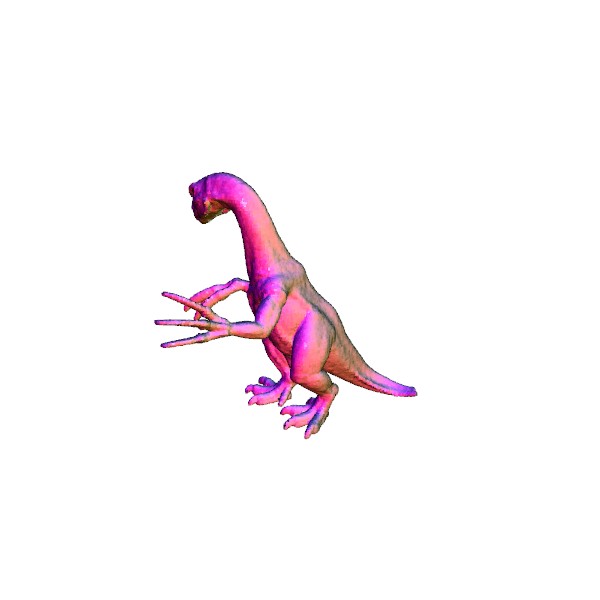}}
	\hfill
	\subfloat{\includegraphics[width=0.16\textwidth]{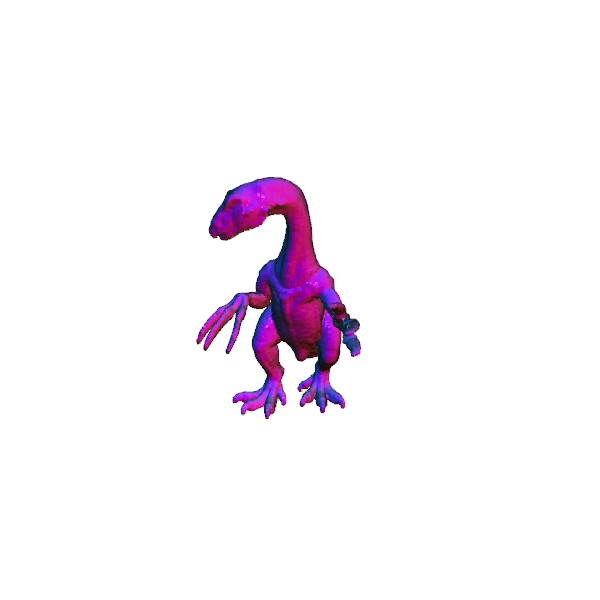}}
	\hfill
	
	\subfloat{\includegraphics[width=0.16\textwidth]{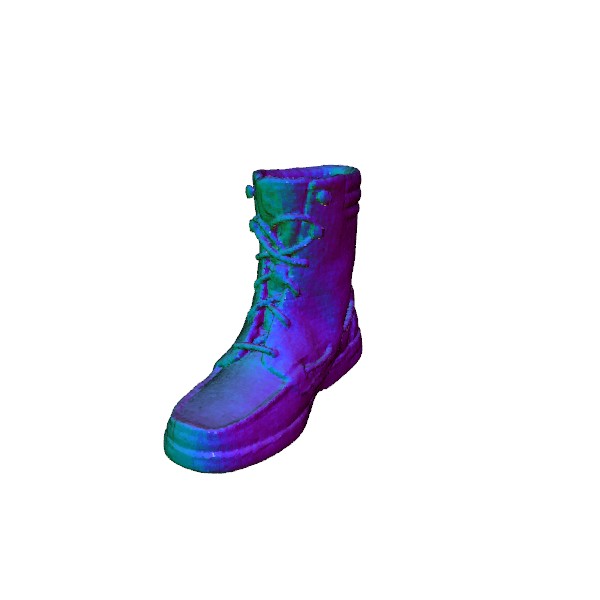}}
	\hfill
	\subfloat{\includegraphics[width=0.16\textwidth]{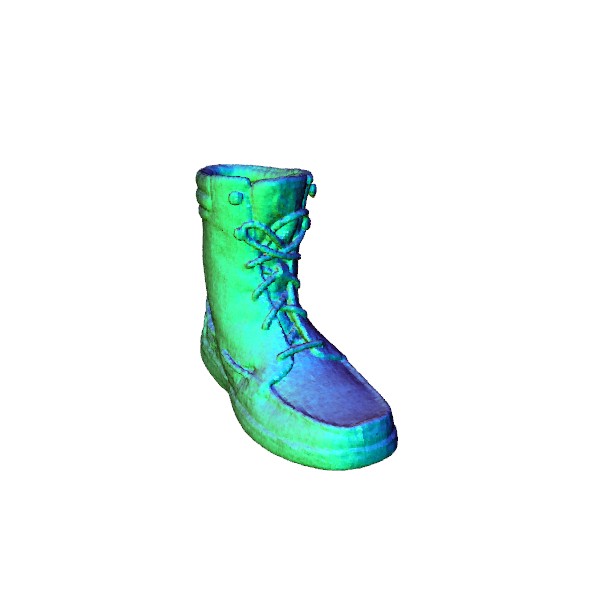}}
	\hfill
	\subfloat{\includegraphics[width=0.16\textwidth]{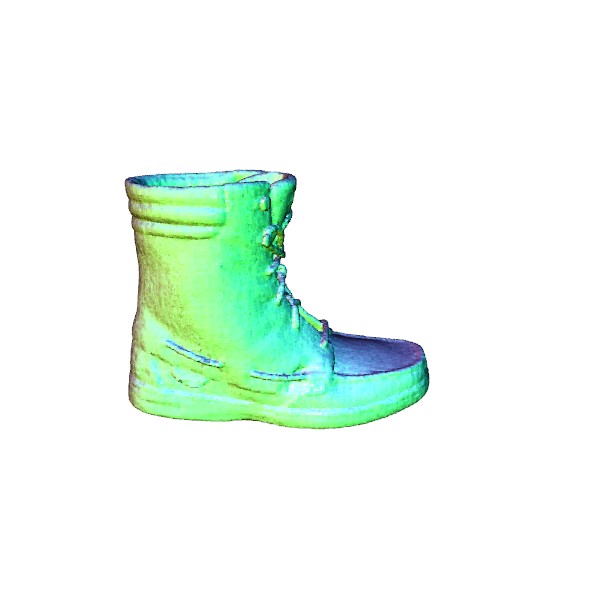}}
	\hfill
	\subfloat{\includegraphics[width=0.16\textwidth]{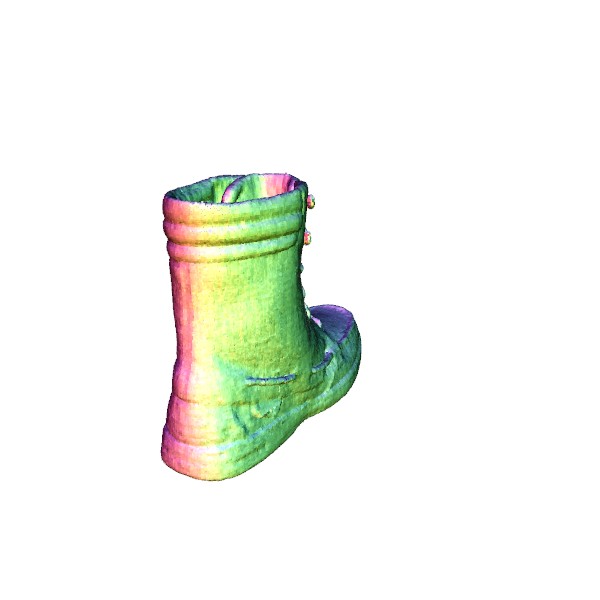}}
	\hfill
	\subfloat{\includegraphics[width=0.16\textwidth]{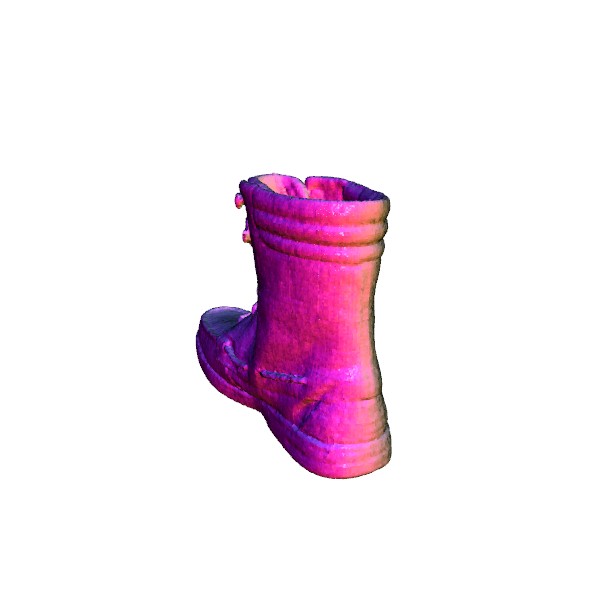}}
	\hfill
	\subfloat{\includegraphics[width=0.16\textwidth]{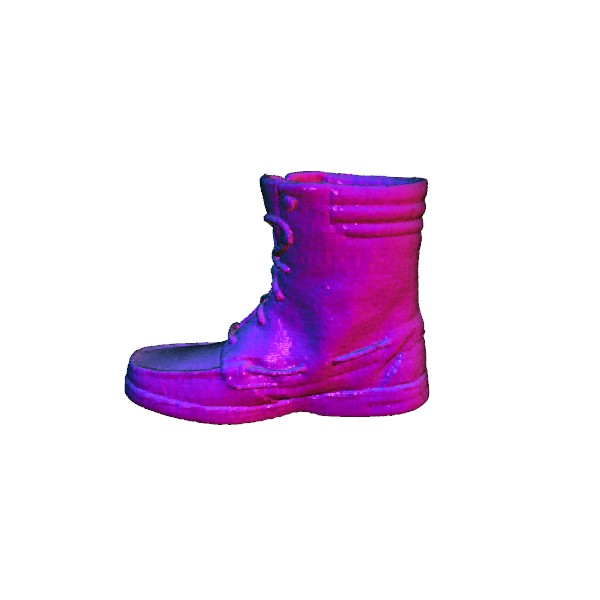}}
	\hfill
	
	\subfloat{\includegraphics[width=0.16\textwidth]{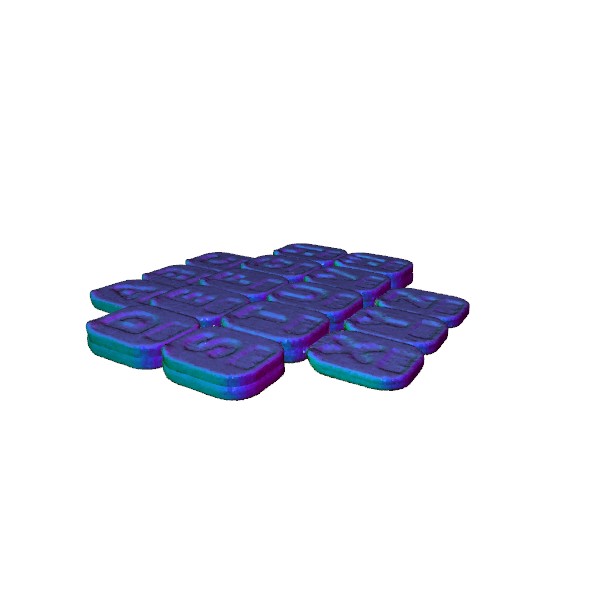}}
	\hfill
	\subfloat{\includegraphics[width=0.16\textwidth]{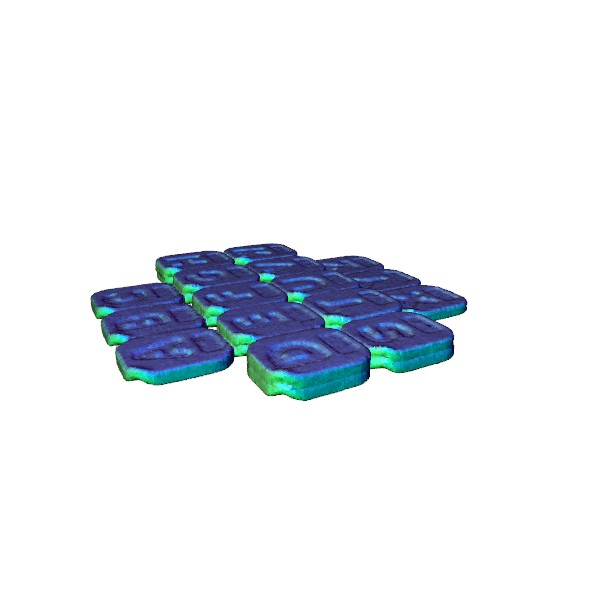}}
	\hfill
	\subfloat{\includegraphics[width=0.16\textwidth]{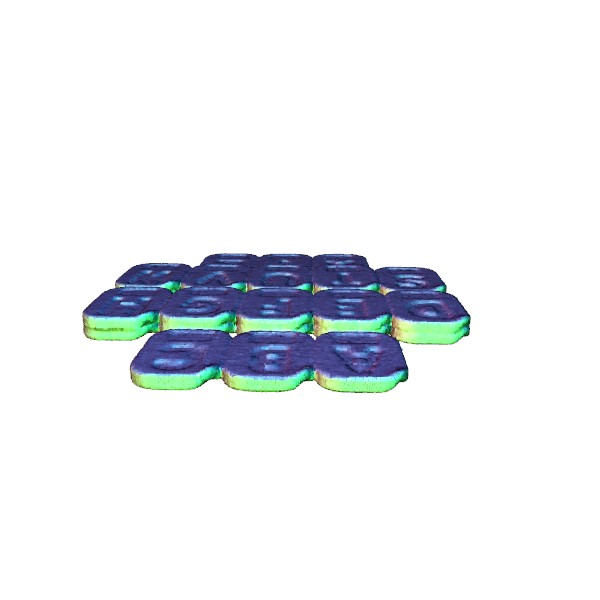}}
	\hfill
	\subfloat{\includegraphics[width=0.16\textwidth]{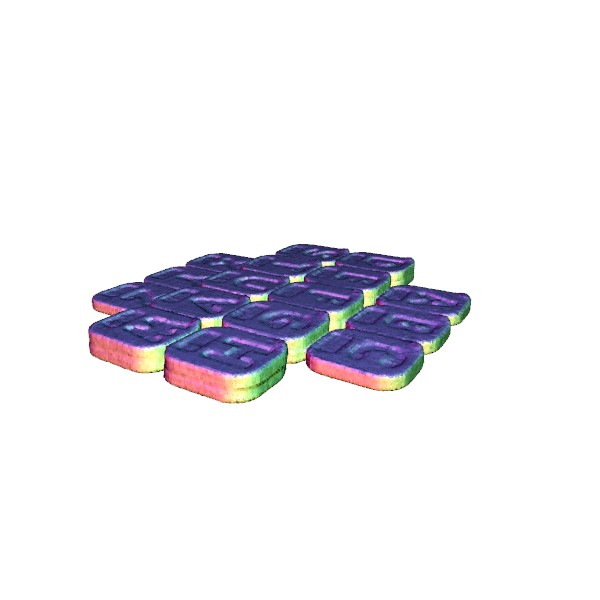}}
	\hfill
	\subfloat{\includegraphics[width=0.16\textwidth]{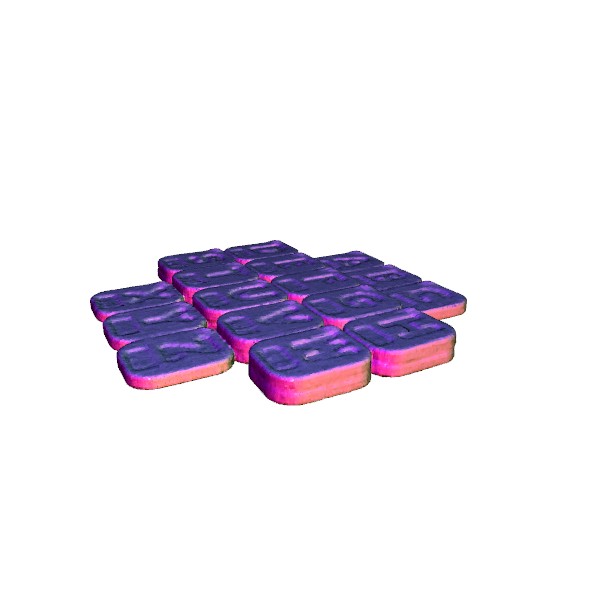}}
	\hfill
	\subfloat{\includegraphics[width=0.16\textwidth]{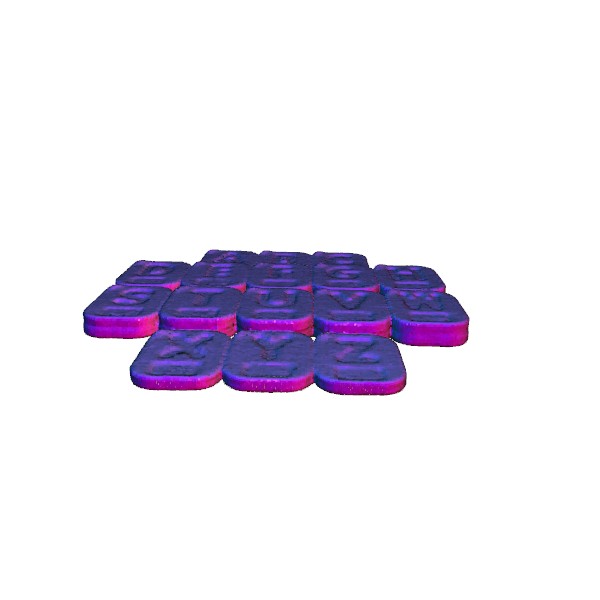}}
	\hfill
	
	\subfloat{\includegraphics[width=0.16\textwidth]{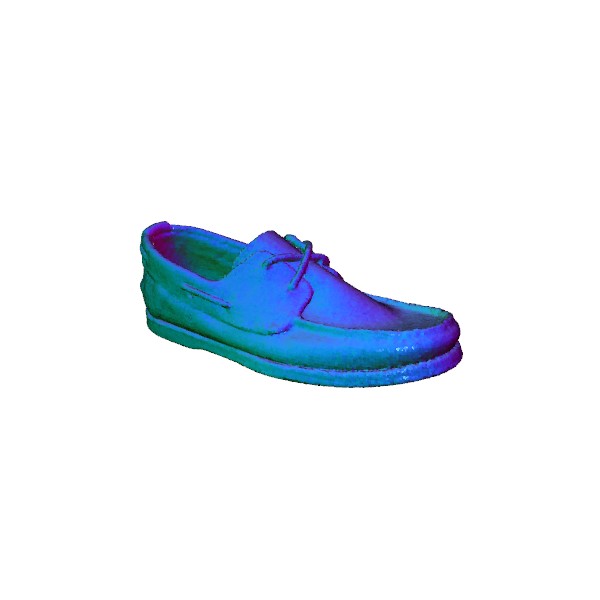}}
	\hfill
	\subfloat{\includegraphics[width=0.16\textwidth]{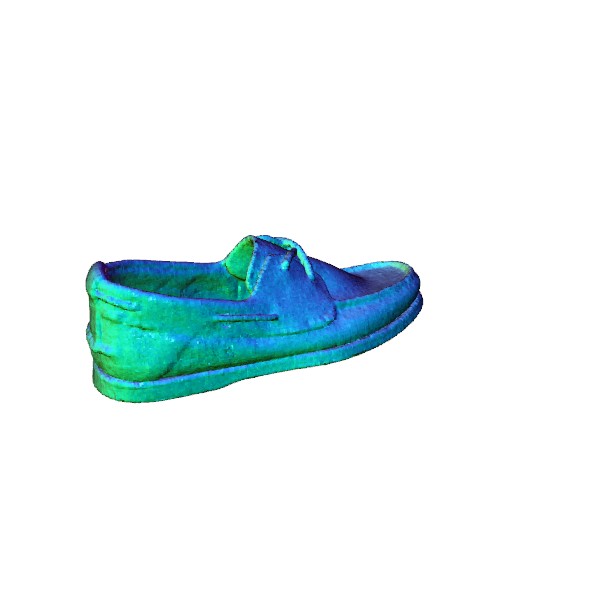}}
	\hfill
	\subfloat{\includegraphics[width=0.16\textwidth]{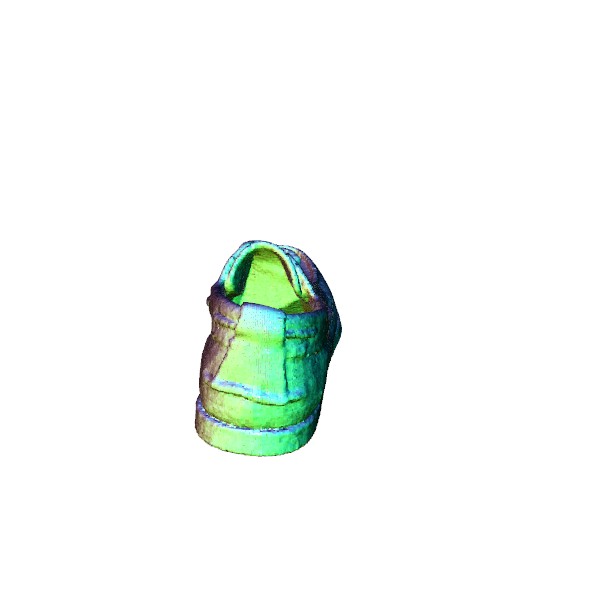}}
	\hfill
	\subfloat{\includegraphics[width=0.16\textwidth]{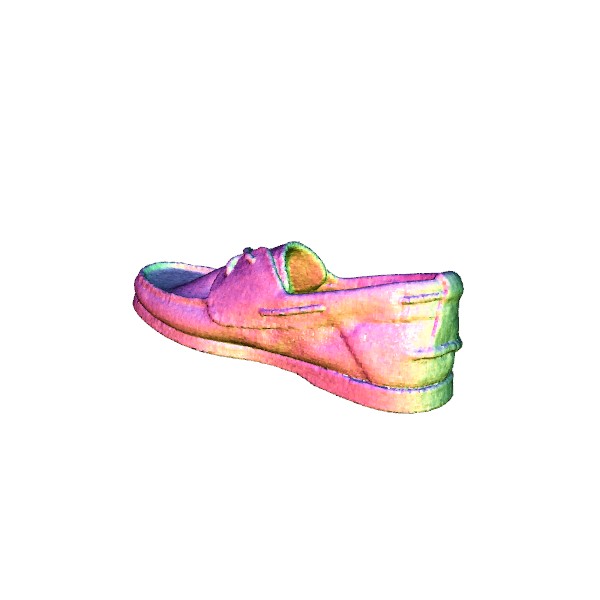}}
	\hfill
	\subfloat{\includegraphics[width=0.16\textwidth]{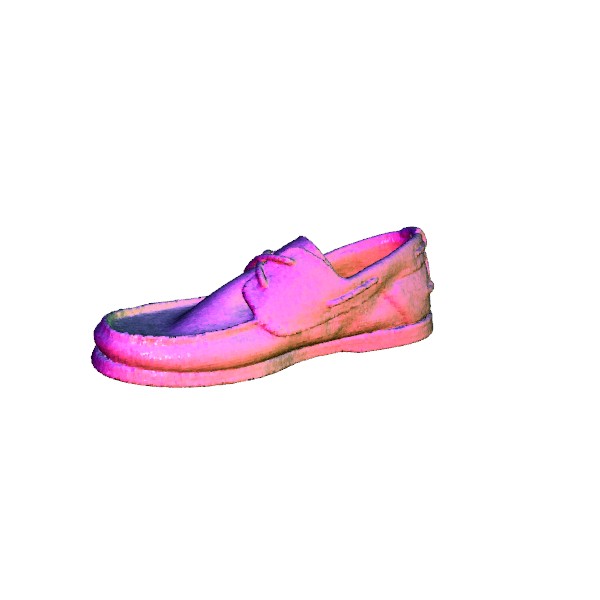}}
	\hfill
	\subfloat{\includegraphics[width=0.16\textwidth]{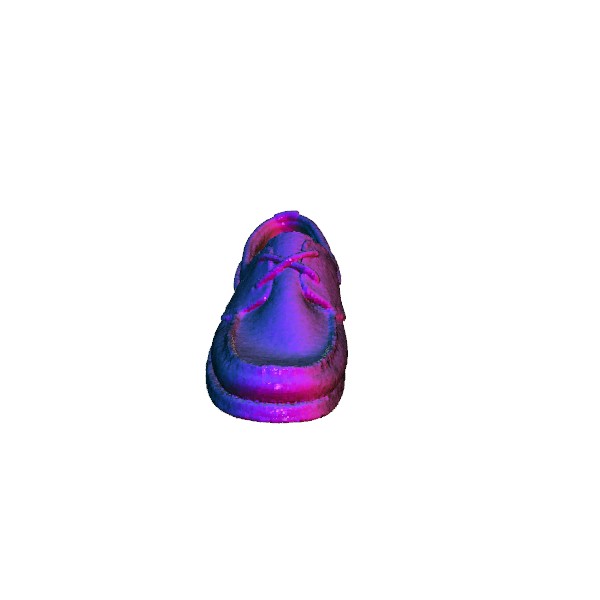}}
	\hfill
	
	\caption{Example reconstruction results on the Google Scanned Objects dataset.}
	\end{adjustbox}
\end{figure}

%% file: main_arxiv.bbl
\begin{thebibliography}{77}
\providecommand{\natexlab}[1]{#1}
\providecommand{\url}[1]{\texttt{#1}}
\expandafter\ifx\csname urlstyle\endcsname\relax
  \providecommand{\doi}[1]{doi: #1}\else
  \providecommand{\doi}{doi: \begingroup \urlstyle{rm}\Url}\fi

\bibitem[Rusu and Cousins(2011)]{rusu20113d}
R.~B. Rusu and S.~Cousins.
\newblock 3d is here: Point cloud library (pcl).
\newblock In \emph{ICRA}, 2011.

\bibitem[Qi et~al.(2017)Qi, Su, Mo, and Guibas]{qi2017pointnet}
C.~R. Qi, H.~Su, K.~Mo, and L.~J. Guibas.
\newblock Point{N}et: Deep learning on point sets for 3d classification and
  segmentation.
\newblock In \emph{CVPR}, 2017.

\bibitem[Curless and Levoy(1996)]{curless1996volumetric}
B.~Curless and M.~Levoy.
\newblock A volumetric method for building complex models from range images.
\newblock In \emph{SIGGRAPH}, 1996.

\bibitem[Hornung et~al.(2013)Hornung, Wurm, Bennewitz, Stachniss, and
  Burgard]{hornung2013octomap}
A.~Hornung, K.~M. Wurm, M.~Bennewitz, C.~Stachniss, and W.~Burgard.
\newblock Octomap: An efficient probabilistic 3d mapping framework based on
  octrees.
\newblock \emph{Autonomous robots}, 2013.

\bibitem[Peng et~al.(2020)Peng, Niemeyer, Mescheder, Pollefeys, and
  Geiger]{peng2020convolutional}
S.~Peng, M.~Niemeyer, L.~Mescheder, M.~Pollefeys, and A.~Geiger.
\newblock Convolutional occupancy networks.
\newblock In \emph{ECCV}, 2020.

\bibitem[Whelan et~al.(2015)Whelan, Leutenegger, Salas-Moreno, Glocker, and
  Davison]{whelan2015elasticfusion}
T.~Whelan, S.~Leutenegger, R.~Salas-Moreno, B.~Glocker, and A.~Davison.
\newblock Elasticfusion: Dense slam without a pose graph.
\newblock Robotics: Science and Systems, 2015.

\bibitem[Hanocka et~al.(2019)Hanocka, Hertz, Fish, Giryes, Fleishman, and
  Cohen-Or]{hanocka2019meshcnn}
R.~Hanocka, A.~Hertz, N.~Fish, R.~Giryes, S.~Fleishman, and D.~Cohen-Or.
\newblock Meshcnn: a network with an edge.
\newblock \emph{TOG}, 2019.

\bibitem[Park et~al.(2019)Park, Florence, Straub, Newcombe, and
  Lovegrove]{park2019deepsdf}
J.~J. Park, P.~Florence, J.~Straub, R.~Newcombe, and S.~Lovegrove.
\newblock Deepsdf: Learning continuous signed distance functions for shape
  representation.
\newblock In \emph{CVPR}, 2019.

\bibitem[Sitzmann et~al.(2020)Sitzmann, Chan, Tucker, Snavely, and
  Wetzstein]{sitzmann2020metasdf}
V.~Sitzmann, E.~Chan, R.~Tucker, N.~Snavely, and G.~Wetzstein.
\newblock Metasdf: Meta-learning signed distance functions.
\newblock \emph{NeurIPS}, 2020.

\bibitem[Zakharov et~al.(2020)Zakharov, Kehl, Bhargava, and
  Gaidon]{zakharov2020autolabeling}
S.~Zakharov, W.~Kehl, A.~Bhargava, and A.~Gaidon.
\newblock Autolabeling 3d objects with differentiable rendering of sdf shape
  priors.
\newblock In \emph{CVPR}, 2020.

\bibitem[Mescheder et~al.(2019)Mescheder, Oechsle, Niemeyer, Nowozin, and
  Geiger]{mescheder2019occupancy}
L.~Mescheder, M.~Oechsle, M.~Niemeyer, S.~Nowozin, and A.~Geiger.
\newblock Occupancy networks: Learning 3d reconstruction in function space.
\newblock In \emph{CVPR}, 2019.

\bibitem[Mildenhall et~al.(2020)Mildenhall, Srinivasan, Tancik, Barron,
  Ramamoorthi, and Ng]{mildenhall2020nerf}
B.~Mildenhall, P.~P. Srinivasan, M.~Tancik, J.~T. Barron, R.~Ramamoorthi, and
  R.~Ng.
\newblock Nerf: Representing scenes as neural radiance fields for view
  synthesis.
\newblock In \emph{ECCV}, 2020.

\bibitem[Yu et~al.(2021)Yu, Ye, Tancik, and Kanazawa]{yu2020pixelnerf}
A.~Yu, V.~Ye, M.~Tancik, and A.~Kanazawa.
\newblock pixelnerf: Neural radiance fields from one or few images.
\newblock In \emph{CVPR}, 2021.

\bibitem[Sitzmann et~al.(2021)Sitzmann, Rezchikov, Freeman, Tenenbaum, and
  Durand]{sitzmann2021light}
V.~Sitzmann, S.~Rezchikov, B.~Freeman, J.~Tenenbaum, and F.~Durand.
\newblock Light field networks: Neural scene representations with
  single-evaluation rendering.
\newblock \emph{NeurIPS}, 2021.

\bibitem[Barron et~al.(2021)Barron, Mildenhall, Tancik, Hedman, Martin-Brualla,
  and Srinivasan]{barron2021mip}
J.~T. Barron, B.~Mildenhall, M.~Tancik, P.~Hedman, R.~Martin-Brualla, and P.~P.
  Srinivasan.
\newblock Mip-nerf: A multiscale representation for anti-aliasing neural
  radiance fields.
\newblock In \emph{ICCV}, 2021.

\bibitem[Tangelder and Veltkamp(2008)]{tangelder2008survey}
J.~W. Tangelder and R.~C. Veltkamp.
\newblock A survey of content based 3d shape retrieval methods.
\newblock \emph{Multimedia tools and applications}, 2008.

\bibitem[Ahmed et~al.(2018)Ahmed, Saint, Shabayek, Cherenkova, Das, Gusev,
  Aouada, and Ottersten]{ahmed2018survey}
E.~Ahmed, A.~Saint, A.~E.~R. Shabayek, K.~Cherenkova, R.~Das, G.~Gusev,
  D.~Aouada, and B.~Ottersten.
\newblock A survey on deep learning advances on different 3d data
  representations.
\newblock \emph{arXiv}, 2018.

\bibitem[Xie et~al.(2021)Xie, Takikawa, Saito, Litany, Yan, Khan, Tombari,
  Tompkin, Sitzmann, and Sridhar]{xie2021neural}
Y.~Xie, T.~Takikawa, S.~Saito, O.~Litany, S.~Yan, N.~Khan, F.~Tombari,
  J.~Tompkin, V.~Sitzmann, and S.~Sridhar.
\newblock Neural fields in visual computing and beyond.
\newblock \emph{arXiv}, 2021.

\bibitem[Yifan et~al.(2021)Yifan, Rahmann, and
  Sorkine-Hornung]{yifan2021geometry}
W.~Yifan, L.~Rahmann, and O.~Sorkine-Hornung.
\newblock Geometry-consistent neural shape representation with implicit
  displacement fields.
\newblock \emph{arXiv}, 2021.

\bibitem[Takikawa et~al.(2021)Takikawa, Litalien, Yin, Kreis, Loop,
  Nowrouzezahrai, Jacobson, McGuire, and Fidler]{takikawa2021neural}
T.~Takikawa, J.~Litalien, K.~Yin, K.~Kreis, C.~Loop, D.~Nowrouzezahrai,
  A.~Jacobson, M.~McGuire, and S.~Fidler.
\newblock Neural geometric level of detail: Real-time rendering with implicit
  3d shapes.
\newblock In \emph{CVPR}, 2021.

\bibitem[M{\"u}ller et~al.(2022)M{\"u}ller, Evans, Schied, and
  Keller]{muller2022instant}
T.~M{\"u}ller, A.~Evans, C.~Schied, and A.~Keller.
\newblock Instant neural graphics primitives with a multiresolution hash
  encoding.
\newblock \emph{arXiv}, 2022.

\bibitem[Tang et~al.(2021)Tang, Chen, Yang, Wang, Liu, Yang, and
  Gao]{tang2021octfield}
J.-H. Tang, W.~Chen, J.~Yang, B.~Wang, S.~Liu, B.~Yang, and L.~Gao.
\newblock Octfield: Hierarchical implicit functions for 3d modeling.
\newblock \emph{arXiv}, 2021.

\bibitem[Williams et~al.(2021)Williams, Gojcic, Khamis, Zorin, Bruna, Fidler,
  and Litany]{williams2021neural}
F.~Williams, Z.~Gojcic, S.~Khamis, D.~Zorin, J.~Bruna, S.~Fidler, and
  O.~Litany.
\newblock Neural fields as learnable kernels for 3d reconstruction.
\newblock \emph{arXiv}, 2021.

\bibitem[Zakharov et~al.(2021)Zakharov, Ambrus, Guizilini, Park, Kehl, Durand,
  Tenenbaum, Sitzmann, Wu, and Gaidon]{zakharov2021single}
S.~Zakharov, R.~A. Ambrus, V.~C. Guizilini, D.~Park, W.~Kehl, F.~Durand, J.~B.
  Tenenbaum, V.~Sitzmann, J.~Wu, and A.~Gaidon.
\newblock Single-shot scene reconstruction.
\newblock In \emph{CoRL}, 2021.

\bibitem[Chen and Zhang(2019)]{chen2019learning}
Z.~Chen and H.~Zhang.
\newblock Learning implicit fields for generative shape modeling.
\newblock In \emph{CVPR}, 2019.

\bibitem[Lombardi et~al.(2019)Lombardi, Simon, Saragih, Schwartz, Lehrmann, and
  Sheikh]{lombardi2019neural}
S.~Lombardi, T.~Simon, J.~Saragih, G.~Schwartz, A.~Lehrmann, and Y.~Sheikh.
\newblock Neural volumes: Learning dynamic renderable volumes from images.
\newblock \emph{arXiv}, 2019.

\bibitem[Liu et~al.(2020)Liu, Zhang, Peng, Shi, Pollefeys, and
  Cui]{liu2020dist}
S.~Liu, Y.~Zhang, S.~Peng, B.~Shi, M.~Pollefeys, and Z.~Cui.
\newblock Dist: Rendering deep implicit signed distance function with
  differentiable sphere tracing.
\newblock In \emph{CVPR}, 2020.

\bibitem[Niemeyer et~al.(2020)Niemeyer, Mescheder, Oechsle, and
  Geiger]{niemeyer2020differentiable}
M.~Niemeyer, L.~Mescheder, M.~Oechsle, and A.~Geiger.
\newblock Differentiable volumetric rendering: Learning implicit 3d
  representations without 3d supervision.
\newblock In \emph{CVPR}, 2020.

\bibitem[Yu et~al.(2021)Yu, Li, Tancik, Li, Ng, and
  Kanazawa]{yu2021plenoctrees}
A.~Yu, R.~Li, M.~Tancik, H.~Li, R.~Ng, and A.~Kanazawa.
\newblock Plenoctrees for real-time rendering of neural radiance fields.
\newblock In \emph{ICCV}, 2021.

\bibitem[Yang et~al.(2021)Yang, Belongie, Hariharan, and
  Koltun]{yang2021geometry}
G.~Yang, S.~Belongie, B.~Hariharan, and V.~Koltun.
\newblock Geometry processing with neural fields.
\newblock \emph{NeurIPS}, 2021.

\bibitem[Kundu et~al.(2022)Kundu, Genova, Yin, Fathi, Pantofaru, Guibas,
  Tagliasacchi, Dellaert, and Funkhouser]{kundu2022panoptic}
A.~Kundu, K.~Genova, X.~Yin, A.~Fathi, C.~Pantofaru, L.~Guibas,
  A.~Tagliasacchi, F.~Dellaert, and T.~Funkhouser.
\newblock Panoptic neural fields: A semantic object-aware neural scene
  representation.
\newblock \emph{arXiv}, 2022.

\bibitem[Boss et~al.(2021)Boss, Braun, Jampani, Barron, Liu, and
  Lensch]{boss2021nerd}
M.~Boss, R.~Braun, V.~Jampani, J.~T. Barron, C.~Liu, and H.~Lensch.
\newblock Nerd: Neural reflectance decomposition from image collections.
\newblock In \emph{ICCV}, 2021.

\bibitem[Hadadan et~al.(2021)Hadadan, Chen, and Zwicker]{hadadan2021neural}
S.~Hadadan, S.~Chen, and M.~Zwicker.
\newblock Neural radiosity.
\newblock \emph{TOG}, 2021.

\bibitem[Jiang et~al.(2022)Jiang, Yi, Samei, Tuzel, and
  Ranjan]{jiang2022neuman}
W.~Jiang, K.~M. Yi, G.~Samei, O.~Tuzel, and A.~Ranjan.
\newblock Neuman: Neural human radiance field from a single video.
\newblock \emph{arXiv}, 2022.

\bibitem[Ortiz et~al.(2022)Ortiz, Clegg, Dong, Sucar, Novotny, Zollhoefer, and
  Mukadam]{ortiz2022isdf}
J.~Ortiz, A.~Clegg, J.~Dong, E.~Sucar, D.~Novotny, M.~Zollhoefer, and
  M.~Mukadam.
\newblock isdf: Real-time neural signed distance fields for robot perception.
\newblock \emph{arXiv}, 2022.

\bibitem[Adamkiewicz et~al.(2022)Adamkiewicz, Chen, Caccavale, Gardner,
  Culbertson, Bohg, and Schwager]{adamkiewicz2022vision}
M.~Adamkiewicz, T.~Chen, A.~Caccavale, R.~Gardner, P.~Culbertson, J.~Bohg, and
  M.~Schwager.
\newblock Vision-only robot navigation in a neural radiance world.
\newblock \emph{RA-L}, 2022.

\bibitem[Yen-Chen et~al.(2022)Yen-Chen, Florence, Barron, Lin, Rodriguez, and
  Isola]{yen2022nerf}
L.~Yen-Chen, P.~Florence, J.~T. Barron, T.-Y. Lin, A.~Rodriguez, and P.~Isola.
\newblock Nerf-supervision: Learning dense object descriptors from neural
  radiance fields.
\newblock \emph{arXiv}, 2022.

\bibitem[Yen-Chen et~al.(2021)Yen-Chen, Florence, Barron, Rodriguez, Isola, and
  Lin]{yen2021inerf}
L.~Yen-Chen, P.~Florence, J.~T. Barron, A.~Rodriguez, P.~Isola, and T.-Y. Lin.
\newblock inerf: Inverting neural radiance fields for pose estimation.
\newblock In \emph{IROS}, 2021.

\bibitem[Chitta et~al.(2021)Chitta, Prakash, and Geiger]{chitta2021neat}
K.~Chitta, A.~Prakash, and A.~Geiger.
\newblock Neat: Neural attention fields for end-to-end autonomous driving.
\newblock In \emph{ICCV}, 2021.

\bibitem[Simeonov et~al.(2022)Simeonov, Du, Tagliasacchi, Tenenbaum, Rodriguez,
  Agrawal, and Sitzmann]{simeonov2022neural}
A.~Simeonov, Y.~Du, A.~Tagliasacchi, J.~B. Tenenbaum, A.~Rodriguez, P.~Agrawal,
  and V.~Sitzmann.
\newblock Neural descriptor fields: Se (3)-equivariant object representations
  for manipulation.
\newblock In \emph{ICRA}, 2022.

\bibitem[Breyer et~al.(2020)Breyer, Chung, Ott, Roland, and
  Juan]{breyer2020volumetric}
M.~Breyer, J.~J. Chung, L.~Ott, S.~Roland, and N.~Juan.
\newblock Volumetric grasping network: Real-time 6 dof grasp detection in
  clutter.
\newblock In \emph{CoRL}, 2020.

\bibitem[Ichnowski et~al.(2021)Ichnowski, Avigal, Kerr, and
  Goldberg]{ichnowski2021dex}
J.~Ichnowski, Y.~Avigal, J.~Kerr, and K.~Goldberg.
\newblock Dex-nerf: Using a neural radiance field to grasp transparent objects.
\newblock In \emph{CoRL}, 2021.

\bibitem[Irshad et~al.(2022)Irshad, Zakharov, Ambrus, Kollar, Kira, and
  Gaidon]{irshad2022shapo}
M.~Z. Irshad, S.~Zakharov, R.~Ambrus, T.~Kollar, Z.~Kira, and A.~Gaidon.
\newblock Shapo: Implicit representations for multi-object shape, appearance,
  and pose optimization.
\newblock In \emph{ECCV}, 2022.

\bibitem[Huang et~al.(2022)Huang, Hodan, Ma, Zhang, Tran, Twigg, Wu, Yuan,
  Keskin, and Wang]{huang2022ncf}
L.~Huang, T.~Hodan, L.~Ma, L.~Zhang, L.~Tran, C.~Twigg, P.-C. Wu, J.~Yuan,
  C.~Keskin, and R.~Wang.
\newblock Neural correspondence field for object pose estimation.
\newblock \emph{ECCV}, 2022.

\bibitem[Williams et~al.(2022)Williams, Gojcic, Khamis, Zorin, Bruna, Fidler,
  and Litany]{williams2022neural}
F.~Williams, Z.~Gojcic, S.~Khamis, D.~Zorin, J.~Bruna, S.~Fidler, and
  O.~Litany.
\newblock Neural fields as learnable kernels for 3d reconstruction.
\newblock In \emph{CVPR}, 2022.

\bibitem[Gao et~al.(2021)Gao, Chang, Mall, Fei-Fei, and
  Wu]{gao2021objectfolder}
R.~Gao, Y.-Y. Chang, S.~Mall, L.~Fei-Fei, and J.~Wu.
\newblock Objectfolder: A dataset of objects with implicit visual, auditory,
  and tactile representations.
\newblock In \emph{CoRL}, 2021.

\bibitem[Moreau et~al.(2022)Moreau, Piasco, Tsishkou, Stanciulescu, and
  de~La~Fortelle]{moreau2022lens}
A.~Moreau, N.~Piasco, D.~Tsishkou, B.~Stanciulescu, and A.~de~La~Fortelle.
\newblock Lens: Localization enhanced by nerf synthesis.
\newblock In \emph{CoRL}, 2022.

\bibitem[Sucar et~al.(2021)Sucar, Liu, Ortiz, and Davison]{sucar2021imap}
E.~Sucar, S.~Liu, J.~Ortiz, and A.~J. Davison.
\newblock imap: Implicit mapping and positioning in real-time.
\newblock In \emph{ICCV}, 2021.

\bibitem[Zhu et~al.(2022)Zhu, Peng, Larsson, Xu, Bao, Cui, Oswald, and
  Pollefeys]{zhu2022nice}
Z.~Zhu, S.~Peng, V.~Larsson, W.~Xu, H.~Bao, Z.~Cui, M.~R. Oswald, and
  M.~Pollefeys.
\newblock Nice-slam: Neural implicit scalable encoding for slam.
\newblock In \emph{CVPR}, 2022.

\bibitem[Williams et~al.(2021)Williams, Trager, Bruna, and
  Zorin]{williams2021neural_splines}
F.~Williams, M.~Trager, J.~Bruna, and D.~Zorin.
\newblock Neural splines: Fitting 3d surfaces with infinitely-wide neural
  networks.
\newblock In \emph{CVPR}, 2021.

\bibitem[Liu et~al.(2022)Liu, Williams, Jacobson, Fidler, and
  Litany]{liu2022learning}
H.-T.~D. Liu, F.~Williams, A.~Jacobson, S.~Fidler, and O.~Litany.
\newblock Learning smooth neural functions via lipschitz regularization.
\newblock \emph{arXiv}, 2022.

\bibitem[Jang and Agapito(2021)]{jang2021codenerf}
W.~Jang and L.~Agapito.
\newblock Codenerf: Disentangled neural radiance fields for object categories.
\newblock In \emph{ICCV}, 2021.

\bibitem[Sitzmann et~al.(2020)Sitzmann, Martel, Bergman, Lindell, and
  Wetzstein]{sitzmann2020implicit}
V.~Sitzmann, J.~Martel, A.~Bergman, D.~Lindell, and G.~Wetzstein.
\newblock Implicit neural representations with periodic activation functions.
\newblock \emph{NeurIPS}, 2020.

\bibitem[Tancik et~al.(2020)Tancik, Srinivasan, Mildenhall, Fridovich-Keil,
  Raghavan, Singhal, Ramamoorthi, Barron, and Ng]{tancik2020fourier}
M.~Tancik, P.~Srinivasan, B.~Mildenhall, S.~Fridovich-Keil, N.~Raghavan,
  U.~Singhal, R.~Ramamoorthi, J.~Barron, and R.~Ng.
\newblock Fourier features let networks learn high frequency functions in low
  dimensional domains.
\newblock \emph{NeurIPS}, 2020.

\bibitem[Chen et~al.(2022)Chen, Xu, Geiger, Yu, and Su]{chen2022tensorf}
A.~Chen, Z.~Xu, A.~Geiger, J.~Yu, and H.~Su.
\newblock Tensorf: Tensorial radiance fields.
\newblock \emph{arXiv}, 2022.

\bibitem[Reiser et~al.(2021)Reiser, Peng, Liao, and Geiger]{reiser2021kilonerf}
C.~Reiser, S.~Peng, Y.~Liao, and A.~Geiger.
\newblock Kilonerf: Speeding up neural radiance fields with thousands of tiny
  mlps.
\newblock In \emph{ICCV}, 2021.

\bibitem[Liu et~al.(2020)Liu, Gu, Lin, Chua, and Theobalt]{liu2020neural}
L.~Liu, J.~Gu, K.~Z. Lin, T.-S. Chua, and C.~Theobalt.
\newblock Neural sparse voxel fields.
\newblock \emph{NeurIPS}, 2020.

\bibitem[Yang et~al.(2021)Yang, Zhou, Peng, Liang, Mu, and
  Hu]{yang2021recursive}
G.-W. Yang, W.-Y. Zhou, H.-Y. Peng, D.~Liang, T.-J. Mu, and S.-M. Hu.
\newblock Recursive-nerf: An efficient and dynamically growing nerf.
\newblock \emph{arXiv}, 2021.

\bibitem[Kato et~al.(2020)Kato, Beker, Morariu, Ando, Matsuoka, Kehl, and
  Gaidon]{kato2020differentiable}
H.~Kato, D.~Beker, M.~Morariu, T.~Ando, T.~Matsuoka, W.~Kehl, and A.~Gaidon.
\newblock Differentiable rendering: A survey.
\newblock \emph{arXiv}, 2020.

\bibitem[Tewari et~al.(2021)Tewari, Thies, Mildenhall, Srinivasan, Tretschk,
  Wang, Lassner, Sitzmann, Martin-Brualla, Lombardi,
  et~al.]{tewari2021advances}
A.~Tewari, J.~Thies, B.~Mildenhall, P.~Srinivasan, E.~Tretschk, Y.~Wang,
  C.~Lassner, V.~Sitzmann, R.~Martin-Brualla, S.~Lombardi, et~al.
\newblock Advances in neural rendering.
\newblock \emph{arXiv}, 2021.

\bibitem[Sitzmann et~al.(2019)Sitzmann, Zollhoefer, and
  Wetzstein]{sitzmann2019scene}
V.~Sitzmann, M.~Zollhoefer, and G.~Wetzstein.
\newblock Scene representation networks: Continuous 3d-structure-aware neural
  scene representations.
\newblock \emph{NeurIPS}, 2019.

\bibitem[Mustikovela et~al.(2021)Mustikovela, De~Mello, Prakash, Iqbal, Liu,
  Nguyen-Phuoc, Rother, and Kautz]{mustikovela2021self}
S.~K. Mustikovela, S.~De~Mello, A.~Prakash, U.~Iqbal, S.~Liu, T.~Nguyen-Phuoc,
  C.~Rother, and J.~Kautz.
\newblock Self-supervised object detection via generative image synthesis.
\newblock In \emph{ICCV}, 2021.

\bibitem[Niemeyer and Geiger(2021)]{niemeyer2021giraffe}
M.~Niemeyer and A.~Geiger.
\newblock Giraffe: Representing scenes as compositional generative neural
  feature fields.
\newblock In \emph{CVPR}, 2021.

\bibitem[Ost et~al.(2021)Ost, Mannan, Thuerey, Knodt, and Heide]{ost2021neural}
J.~Ost, F.~Mannan, N.~Thuerey, J.~Knodt, and F.~Heide.
\newblock Neural scene graphs for dynamic scenes.
\newblock In \emph{CVPR}, 2021.

\bibitem[Yang et~al.(2021)Yang, Zhang, Xu, Li, Zhou, Bao, Zhang, and
  Cui]{yang2021learning}
B.~Yang, Y.~Zhang, Y.~Xu, Y.~Li, H.~Zhou, H.~Bao, G.~Zhang, and Z.~Cui.
\newblock Learning object-compositional neural radiance field for editable
  scene rendering.
\newblock In \emph{ICCV}, 2021.

\bibitem[Henzler et~al.(2021)Henzler, Reizenstein, Labatut, Shapovalov,
  Ritschel, Vedaldi, and Novotny]{henzler2021unsupervised}
P.~Henzler, J.~Reizenstein, P.~Labatut, R.~Shapovalov, T.~Ritschel, A.~Vedaldi,
  and D.~Novotny.
\newblock Unsupervised learning of 3d object categories from videos in the
  wild.
\newblock In \emph{CVPR}, 2021.

\bibitem[M{\"u}ller et~al.(2022)M{\"u}ller, Simonelli, Porzi, Bul{\`o},
  Nie{\ss}ner, and Kontschieder]{muller2022autorf}
N.~M{\"u}ller, A.~Simonelli, L.~Porzi, S.~R. Bul{\`o}, M.~Nie{\ss}ner, and
  P.~Kontschieder.
\newblock Autorf: Learning 3d object radiance fields from single view
  observations.
\newblock \emph{arXiv}, 2022.

\bibitem[Reizenstein et~al.(2021)Reizenstein, Shapovalov, Henzler, Sbordone,
  Labatut, and Novotny]{reizenstein2021common}
J.~Reizenstein, R.~Shapovalov, P.~Henzler, L.~Sbordone, P.~Labatut, and
  D.~Novotny.
\newblock Common objects in 3d: Large-scale learning and evaluation of
  real-life 3d category reconstruction.
\newblock In \emph{ICCV}, 2021.

\bibitem[Sitzmann et~al.(2020)Sitzmann, Martel, Bergman, Lindell, and
  Wetzstein]{sitzmann2019siren}
V.~Sitzmann, J.~N. Martel, A.~W. Bergman, D.~B. Lindell, and G.~Wetzstein.
\newblock Implicit neural representations with periodic activation functions.
\newblock In \emph{Proc. NeurIPS}, 2020.

\bibitem[Chang et~al.(2015)Chang, Funkhouser, Guibas, Hanrahan, Huang, Li,
  Savarese, Savva, Song, Su, et~al.]{chang2015shapenet}
A.~X. Chang, T.~Funkhouser, L.~Guibas, P.~Hanrahan, Q.~Huang, Z.~Li,
  S.~Savarese, M.~Savva, S.~Song, H.~Su, et~al.
\newblock Shape{N}et: An information-rich 3d model repository.
\newblock \emph{arXiv}, 2015.

\bibitem[Zhou and Jacobson(2016)]{zhou2016thingi10k}
Q.~Zhou and A.~Jacobson.
\newblock Thingi10k: A dataset of 10,000 3d-printing models.
\newblock \emph{arXiv}, 2016.

\bibitem[Downs et~al.(2022)Downs, Francis, Koenig, Kinman, Hickman, Reymann,
  McHugh, and Vanhoucke]{downs2022google}
L.~Downs, A.~Francis, N.~Koenig, B.~Kinman, R.~Hickman, K.~Reymann, T.~B.
  McHugh, and V.~Vanhoucke.
\newblock Google scanned objects: A high-quality dataset of 3d scanned
  household items.
\newblock \emph{arXiv}, 2022.

\bibitem[Acc()]{AccuCities}
3d model of london and city models.
\newblock \url{https://www.accucities.com/}.
\newblock Accessed: 2022-05-18.

\bibitem[Davies et~al.(2020)Davies, Nowrouzezahrai, and
  Jacobson]{davies2020overfit}
T.~Davies, D.~Nowrouzezahrai, and A.~Jacobson.
\newblock Overfit neural networks as a compact shape representation.
\newblock \emph{arXiv}, 2020.

\bibitem[Kazhdan et~al.(2006)Kazhdan, Bolitho, and Hoppe]{kazhdan2006poisson}
M.~Kazhdan, M.~Bolitho, and H.~Hoppe.
\newblock Poisson surface reconstruction.
\newblock In \emph{SGP}, 2006.

\bibitem[Zhou et~al.(2018)Zhou, Park, and Koltun]{zhou2018open3d}
Q.-Y. Zhou, J.~Park, and V.~Koltun.
\newblock Open3d: A modern library for 3d data processing.
\newblock \emph{arXiv}, 2018.

\bibitem[Ravi et~al.(2020)Ravi, Reizenstein, Novotny, Gordon, Lo, Johnson, and
  Gkioxari]{ravi2020accelerating}
N.~Ravi, J.~Reizenstein, D.~Novotny, T.~Gordon, W.-Y. Lo, J.~Johnson, and
  G.~Gkioxari.
\newblock Accelerating 3d deep learning with pytorch3d.
\newblock \emph{arXiv}, 2020.

\end{thebibliography}
